%% file: neurips_2026.tex
\PassOptionsToPackage{numbers,sort&compress}{natbib}
\documentclass{article}

\usepackage[main,final]{neurips_2026}

\usepackage[utf8]{inputenc}
\usepackage[T1]{fontenc}
\usepackage{placeins}
\usepackage{float}
\usepackage{hyperref}
\usepackage{url}
\usepackage{booktabs}
\usepackage{amsfonts}
\usepackage{nicefrac}
\usepackage{microtype}
\usepackage[dvipsnames,table]{xcolor}
\usepackage{amsmath}
\usepackage{amssymb}
\usepackage[capitalize,nameinlink]{cleveref}
\usepackage{pifont}
\usepackage{fontawesome5}
\usepackage{multirow}
\usepackage{graphicx}

\makeatletter
\def\scaffoldtabletype{table}
\AddToHook{cmd/@caption/before}{%
  \ifx\@captype\scaffoldtabletype
    \setlength{\abovecaptionskip}{0pt}%
    \setlength{\belowcaptionskip}{2pt}%
  \fi
}
\makeatother

\newcommand{\cmark}{\ding{51}}
\newcommand{\xmark}{\ding{55}}
\newcommand{\B}[1]{\textbf{#1}}
\newcommand{\U}[1]{\underline{#1}}
\newcommand{\na}{\mbox{N/A}}
\definecolor{colorid}{RGB}{255, 245, 180}
\definecolor{colorcd}{RGB}{180, 220, 255}
\definecolor{colorod}{RGB}{255, 200, 200}
\newcommand{\datasetbox}[2]{\begingroup\setlength{\fboxsep}{1pt}\colorbox{#1}{\textbf{#2}}\endgroup}
\newcommand{\datasetboxplain}[2]{\begingroup\setlength{\fboxsep}{1pt}\colorbox{#1}{#2}\endgroup}
\newcommand{\dsid}[1]{\datasetbox{colorid}{#1}}
\newcommand{\dscd}[1]{\datasetbox{colorcd}{#1}}
\newcommand{\dsod}[1]{\datasetbox{colorod}{#1}}

\title{Scaffold3D: SfM-Conditioned Pointmap\\Prediction for Multi-View 3D Reconstruction}

\author{%
  Frano Raji\v{c}$^{1}$, Yutong Chen$^{1}$, Haofei Xu$^{1}$, Zador Pataki$^{1}$, Marc Pollefeys$^{1,2}$, Siyu Tang$^{1}$ \\
  $^{1}$ETH Zurich \qquad $^{2}$Microsoft
}

\begin{document}

\maketitle

\begin{abstract}
Feed-forward models make multi-view 3D reconstruction fast and robust, but their predictions often remain locally inaccurate and geometrically inconsistent across views. Structure-from-motion (SfM) offers a complementary geometric signal through camera poses and 3D points estimated by correspondence filtering and optimization. To propagate this globally consistent \emph{SfM scaffold} into dense geometry, we introduce \textbf{Scaffold3D}, an SfM-conditioned reconstruction framework that injects image patch-aligned SfM point tokens into a pairwise feed-forward pointmap predictor. Our approach preserves image-token reasoning while using explicit SfM structure to condition dense prediction. Because Scaffold3D predicts pairwise pointmaps in local frames, we fuse them with global alignment and an SfM anchoring term. Scaffold3D achieves stronger overall performance than feed-forward reconstruction models, point-token propagation, depth-completion baselines, and recent geometry-conditioned models on standard benchmarks (ScanNet++, ETH3D, and Tanks\&Temples) and out-of-domain MV-dVRK, and is competitive on 4D-DRESS. On standard benchmarks, gains over point-token propagation are especially clear in low- and no-overlap regions. Our pipeline, built around a \emph{two-view} SfM-conditioned pointmap predictor, outperforms several off-the-shelf \emph{multi-view} geometry-conditioned baselines. Together, these results establish SfM-scaffold conditioning as a practical interface between learned pointmap prediction and classical geometric optimization. Project page: \url{https://ethz-vlg.github.io/scaffold3d}.
\end{abstract}
\begin{figure}[H]
\centering
\vspace{-0.35cm}
\includegraphics[width=0.8\linewidth]{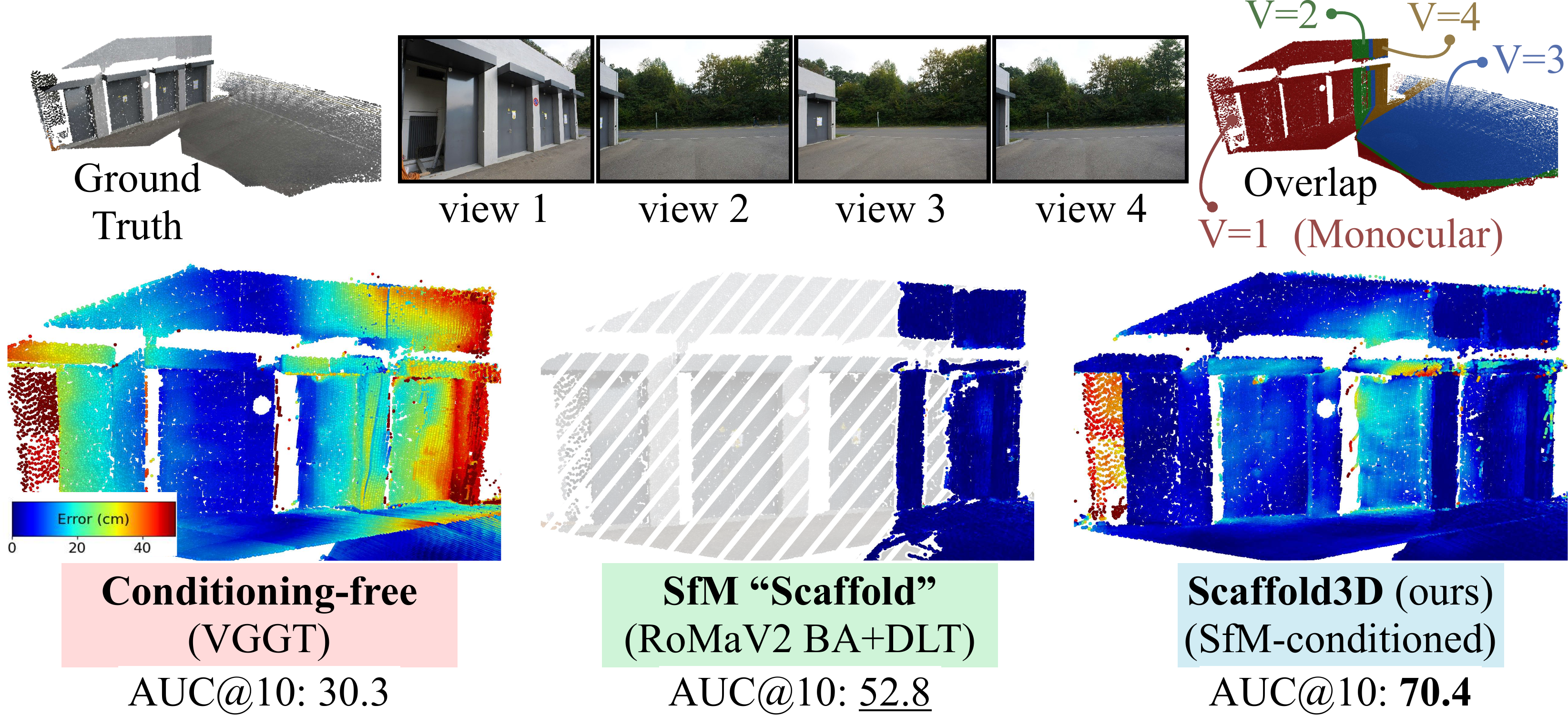}
\vspace{-0.25cm}
\caption{\textbf{Scaffold3D completes an SfM scaffold.} Top: ground-truth geometry, input views, and multi-view overlap. Bottom: a conditioning-free pointmap predictor is dense but inaccurate, while the SfM scaffold is accurate but incomplete. Scaffold3D conditions dense prediction on the SfM scaffold and improves AUC@10 cm on this ETH3D example from 30.3 and 52.8 to 70.4.}
\label{fig:teaser}
\vspace{-0.2cm}
\end{figure}

\section{Introduction}
Multi-view 3D reconstruction is a core component in robotics, AR/VR, digital content creation, and embodied perception. Modern feed-forward models have made dense reconstruction much faster and more robust, and now predict depth, pointmaps, and camera geometry directly from images~\cite{wang2024dust3r,mast3r_eccv24,wang2025vggt,wang2025pi3,keetha2025mapanything,depthanything3}. However, these models typically do not enforce explicit geometric constraints during prediction to improve multi-view consistency and precision. As a result, despite strong robustness and coverage, their dense outputs often remain geometrically misaligned across views and locally inaccurate, especially on fine structures and out-of-domain scenes.

Structure-from-motion (SfM) pipelines remain strong at enforcing geometric consistency through correspondence filtering and optimization~\cite{revisitedsfm,lindenberger2021pixsfm,duisterhof2025mastrsfm,wang2024vggsfm,pataki2025mpsfm}. Modern dense matchers and learned SfM pipelines have improved robustness to low texture and sparse, wide-baseline views~\cite{edstedt2025romav2,pataki2025mpsfm,pataki2026vidmap}. SfM pipelines estimate camera poses and 3D points, either before dense prediction or as part of a reconstruction pipeline that also uses learned priors. The resulting \emph{SfM scaffold} (\textit{i.e.}, poses, intrinsics, and 3D points) covers only a subset of image regions, but it provides explicit multi-view constraints. We therefore frame dense reconstruction as a propagation problem: given a reliable SfM scaffold, how should a learned model propagate it into dense pointmaps without losing its global consistency?

A common approach refines feed-forward predictions with geometric optimization, such as bundle adjustment or global alignment~\cite{wang2024dust3r,mast3r_eccv24,wang2025vggt}. When SfM geometry is used only for post-processing, the network predicts dense geometry without receiving the SfM cameras and points as inputs during its forward pass. Another option is monocular depth completion: complete each view from SfM depth samples and then fuse the views~\cite{TTPromptDC,viola2024marigolddc,ke2026capa,zuo2025omni}. This uses the scaffold points, but it treats dense prediction mostly as a per-image problem. Multi-view reconstruction should instead exploit cross-view consistency, so that observations from one view constrain another through camera geometry and correspondence.

Recent geometry-conditioned models move in this direction, but use the scaffold through different interfaces. Pow3R~\cite{pow3r_cvpr25}, Pi3X~\cite{wang2025pi3}, and MapAnything~\cite{keetha2025mapanything} feed SfM-derived geometry as scaffold depth maps, together with intrinsics or pose conditioning, rather than explicit 3D point observations. GGPT~\cite{chen2026ggpt} takes the opposite route: it preserves the scaffold as point tokens, but propagates them without access to image features, completely discarding appearance and matching cues that are useful for dense pointmap prediction. This motivates combining explicit SfM point observations with image features so that geometry and appearance jointly guide dense prediction.

We introduce \textbf{Scaffold3D}, an SfM-scaffold-conditioned pointmap prediction framework for multi-view reconstruction. Scaffold3D contains a feed-forward image encoder-decoder and augments the decoder with image patch-aligned SfM point tokens. For each view, SfM 3D points whose projections fall into the same image patch are encoded and aggregated into a single point token. Point tokens across patches interact through self-attention and are then injected into the image-token decoder. This allows partial but globally consistent SfM geometry to guide dense pairwise pointmap prediction beyond directly triangulated regions while preserving dense visual correspondence cues.

Since pairwise pointmaps are predicted in local coordinate frames, we further introduce an SfM-anchored global alignment formulation to fuse them into a globally consistent reconstruction. As a result, the SfM scaffold serves two complementary roles: locally, it guides the pairwise pointmap predictor toward geometrically informed dense reconstruction, and globally, it acts as a reliable anchor for fusing pairwise predictions into a coherent 3D scene.

Experiments on standard datasets~\cite{yeshwanth2023scannet++,schoeps2017eth3d,Knapitsch2017tandt} and out-of-domain data~\cite{wang20244ddress,dvrk-smv} support four conclusions. First, conditioning on estimated SfM scaffolds improves over unconditioned feed-forward baselines~\cite{wang2024dust3r,mast3r_eccv24,wang2025vggt,wang2025pi3} and feed-forward baselines that use geometric optimization only after dense prediction~\cite{duisterhof2025mastrsfm}. Second, combining image and point tokens improves over point-only propagation~\cite{chen2026ggpt}, especially in low- and no-overlap regions. Third, our pipeline, built around a \emph{two-view} Scaffold3D predictor, outperforms several off-the-shelf \emph{multi-view} geometry-conditioned baselines~\cite{keetha2025mapanything,wang2025pi3,chen2026ggpt}. Fourth, conditioning improves mean accuracy with all eleven scaffold sources, including those with under 3\% pixel coverage. We release training and evaluation code, pre-trained checkpoints, and dataset preprocessing scripts to facilitate future research and reproducibility.

\section{Related Work}
\label{sec:related_work}

\textbf{SfM-Based 3D Reconstruction.}
COLMAP~\cite{revisitedsfm} remains a standard reconstruction approach for views with sufficient overlap. Modern SfM frameworks improve robustness to low overlap, large baselines, or low texture by incorporating learning-based matchers, learned initializations, or depth priors~\cite{wang2024vggsfm,duisterhof2025mastrsfm,pataki2025mpsfm,pataki2026vidmap}. GGPT~\cite{chen2026ggpt} initializes bundle adjustment (BA) with feed-forward camera and point predictions, then constructs the scaffold from refined cameras and dense correspondences by direct linear triangulation (DLT). Despite improved matching robustness, triangulation still requires reliable cross-view correspondences and cannot constrain regions visible in only one view. Whereas prior-assisted SfM uses learned geometry to guide geometric optimization, Scaffold3D also conditions the image-token decoder on the SfM scaffold to complete geometry beyond triangulated support.

\textbf{Feed-Forward 3D Reconstruction.}
Recent transformer-based models~\cite{wang2024dust3r,spann3r,wang2025pi3,wang2025vggt,keetha2025mapanything,cut3r,zhang2024monst3r,dens3r,zhang2025d4rt,zhang2025FLARE} predict dense pointmaps or depth maps from RGB images, with cameras predicted directly or recovered geometrically. Feed-forward prediction is much faster than iterative SfM optimization. Large-scale training gives strong sparse-view robustness, but predictions can remain geometrically inconsistent across views, especially out of distribution~\cite{chen2026ggpt}. Stereo and multi-view stereo (MVS) methods combine cross-view matching with learned priors, but require calibrated inputs~\cite{izquierdo2025mvsanywhere,stereoanywhere,wang2025lightweightmvs}. Our pipeline takes only RGB images and estimates an SfM scaffold to condition dense prediction, achieving stronger overall reconstruction accuracy than the evaluated feed-forward baselines on standard benchmarks and MV-dVRK.

\textbf{Depth Completion (DC).}
Depth completion predicts dense depth from depth values available at only a subset of image pixels, typically through \textit{injection-based} methods~\cite{lin2024promptda,wang2025priorda,guo2025murre} trained with depth inputs or \textit{optimization-based} methods~\cite{viola2024marigolddc,TTPromptDC} that fit predictions to depth samples at test time, including multi-view VGGT adaptation in CAPA~\cite{ke2026capa}. Hybrid approaches~\cite{zuo2025omni,ognidc} predict depth gradients and integrate them into dense depth maps using a differentiable solver. However, independent per-view completion can produce inconsistent predictions in overlapping regions without depth samples, even with shared SfM guidance. Scaffold3D instead learns SfM-conditioned pairwise completion, jointly using both images without per-scene adaptation of the predictor.

\textbf{Multi-View Geometry Conditioning.}
Multi-view models incorporate geometry through different interfaces. Pow3R~\cite{pow3r_cvpr25} and Pi3X~\cite{wang2025pi3} encode depth or intrinsics as patch tokens, while MapAnything~\cite{keetha2025mapanything} fuses ray-geometry features with image embeddings. Align3R~\cite{lu2024align3r} encodes monocular depth as pointmaps and injects their features into a DUSt3R decoder for dynamic-video reconstruction. GGPT~\cite{chen2026ggpt} propagates partial 3D structures via point tokens alone. Scaffold3D instead injects partial, triangulated SfM points into an image-token decoder, retaining appearance-based reasoning during completion and using the same SfM geometry to anchor the merging of pairwise predictions.

\section{Method}
\label{sec:method}
\begin{figure}[t]
  \centering
  \includegraphics[width=\linewidth]{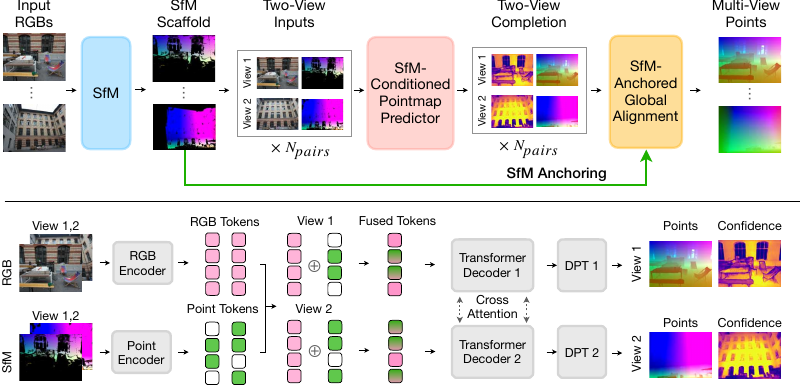}
  \vspace{-0.6cm}
  \caption{\textbf{Scaffold3D overview.} \textit{Top}: Pipeline. First, we run SfM to triangulate a partial geometry scaffold. Second, we use an SfM-conditioned feed-forward pointmap predictor to predict complete pointmaps for view pairs, conditioning on both image tokens and point tokens. Finally, we fuse pairwise pointmaps with SfM-anchored global alignment. \textit{Bottom}: Architecture of the predictor.}
  \vspace{-0.3cm}
  \label{fig:methodology}
\end{figure}

Given $N$ RGB input views $\{I_n \in \mathbb{R}^{H\times W \times 3}\}_{n=1}^N$, our goal is to predict dense 3D point maps $\{X_n \in \mathbb{R}^{H \times W \times 3}\}_{n=1}^{N}$ in a unified global coordinate system. As illustrated in Fig.~\ref{fig:methodology}, \emph{Scaffold3D} operates in three stages. First, we leverage an SfM pipeline to estimate a geometric scaffold $\{ S_n \in \mathbb{R}^{H\times W \times 3} \}_{n=1}^N$, which provides geometrically consistent 3D points wherever reliable cross-view correspondences are available, but leaves regions without such support incomplete. Second, an SfM-conditioned pointmap predictor predicts complete two-view pointmaps by conditioning on the RGB images and their respective SfM scaffolds. Finally, we introduce an SfM-anchored global alignment to aggregate these pairwise predictions into a globally consistent and dense 3D reconstruction.

\subsection{SfM Scaffold}
\label{sec:method_scaffold}
We adopt the SfM pipeline from GGPT~\cite{chen2026ggpt} to generate our geometric scaffold. It establishes dense correspondences using RoMaV2~\cite{edstedt2025romav2} and initializes the reconstruction with cameras and points predicted by a feed-forward reconstruction model, such as VGGT~\cite{wang2025vggt}. Global bundle adjustment (BA) refines these initial cameras and 3D points against the correspondences, yielding camera-to-world poses $\{T_n\}_{n=1}^{N}$ and intrinsics $\{K_n\}_{n=1}^{N}$. Matches are filtered by matching score and cycle consistency, then checked for epipolar consistency with the estimated cameras. Using the refined cameras, we triangulate tracks observed in at least two views using DLT and filter the resulting points by reprojection error and a minimum triangulation angle.

Using the resulting 3D-2D correspondences, we derive a multi-view SfM pointmap $\{S_n \in \mathbb{R}^{H\times W\times 3}\}_{n=1}^{N}$. Entries without valid triangulated support are left empty in $S_n$. We define the set of pixels with valid SfM support as $\Omega_{\mathcal{S}}=\{(n,p) \mid \text{pixel } p \text{ in view } n \text{ has a valid SfM estimate}\}$. These geometrically consistent yet incomplete points serve as a prior, which we propagate to regions lacking geometry support to achieve an accurate and complete reconstruction.

\subsection{SfM-Conditioned Pointmap Predictor}
To densify the partial reconstruction from $\Omega_{\mathcal{S}}$ across the entire image, Scaffold3D uses a two-view conditioned pointmap predictor. For a given image pair $(a,b)$, the network takes the RGB images and normalized SfM points as input to predict complete 3D pointmaps $(\hat{X}_{a,a}, \hat{X}_{b,a})$ in the coordinate system of view $a$. We describe the architecture, normalization, and training objectives below.

\subsubsection{Network Architecture}
The architecture of our SfM-conditioned pointmap predictor (Fig.~\ref{fig:methodology}) comprises an RGB encoder and a point encoder to produce view-specific tokens. Spatially aligned RGB and point tokens are fused via element-wise addition before the first decoder layer at each view. These fused tokens are then processed by two cross-attended decoders to predict points and confidence maps. We instantiate the RGB encoder and decoders using the MASt3R architecture~\cite{mast3r_eccv24} and introduce a novel point encoder for aggregating SfM observations into patch-aligned conditioning tokens.

\noindent\textbf{RGB encoder.} Following~\cite{mast3r_eccv24,wang2024dust3r}, for each view $v\in\{a,b\}$, the input image $I_v$ is patchified and encoded by a ViT encoder~\cite{vit}. We project the encoder features to the decoder width $d$ and denote the resulting RGB tokens as $f_{v}^0 \in \mathbb{R}^{L\times d}$, where $L=HW/{P_{\text{patch}}}^2$ and $P_{\text{patch}}$ is the patch size.

\noindent\textbf{Point encoder.} We map the SfM inputs to a sequence of point tokens spatially aligned with the RGB tokens. Given the normalized SfM input $\hat{S}_v \in \mathbb{R}^{H\times W \times 3}$, we partition the map into patches of size $P_{\text{patch}}$. For each pixel $(v,p)\in \Omega_{\mathcal S}$ with a valid SfM estimate, an MLP $\phi_{\text{pt}}$ encodes its 3D coordinates $\hat{S}_v(p)$ and its relative 2D location within the patch $\delta(p)$:
\begin{equation}
    h_{v,p} = \phi_{\mathrm{pt}}\!\left(\hat S_v(p),\delta(p)\right).
\end{equation}
For each nonempty patch $q$, we average the features $h_{v,p}$ within the patch to compute a patch-level feature $g_{v,q}$:
\begin{equation}
    g_{v,q} = \frac{1}{|\mathcal{K}_{v,q}|} \sum_{p\in\mathcal{K}_{v,q}} h_{v,p}, \qquad \mathcal{K}_{v,q}=\{p \mid (v, p) \in\Omega_{\mathcal{S}},\ p \text{ is in patch } q\}.
\end{equation}
We collect features $g_{v,q} \in \mathbb{R}^d$ for all patches with valid SfM support to form a sequence $g_v \in \mathbb{R}^{L'\times d}$, where $L' \leq L$. A residual MLP and layer normalization refine each aggregated token, followed by lightweight self-attention blocks that exchange information among point tokens:
\begin{equation}
    Z_v^{\ell+1}=\mathrm{PointBlock}^{\ell}\!\left(Z_v^{\ell}\right), \qquad Z_v^{0}=\{\rho(g_{v,q}) \mid |\mathcal{K}_{v,q}|>0\}.
\end{equation}
Here, $\rho$ denotes the per-token refinement. After processing, we scatter refined features onto the image-token grid as $\bar g_{v} \in \mathbb{R}^{L\times d}$, setting $\bar g_{v,q}=0$ for empty patches. This ensures that point tokens $\bar g_{v}$ are spatially aligned with image tokens $f_{v}^0$.

\noindent\textbf{RGB and point token fusion.} We fuse the RGB tokens $(f_a^0, f_b^0)$ and point tokens $(\bar g_a, \bar g_b)$ for each view via element-wise addition:
\begin{equation}
    \tilde f_{a}^{0}=f_{a}^{0}+\bar g_{a}, \quad \tilde f_{b}^{0}=f_{b}^{0}+\bar g_{b}.
\label{eq:point_add}
\end{equation}

\noindent\textbf{Decoders.} The fused tokens $\tilde f_{a}^{0}$ and $\tilde f_{b}^{0}$ are fed into decoders $\mathrm{Dec}_a$ and $\mathrm{Dec}_b$, which employ mutual cross-attention:
\begin{equation}
    \tilde f_a^{\ell+1}=\mathrm{Dec}_a^{\ell}(\tilde f_a^\ell,\tilde f_b^\ell), \qquad \tilde f_b^{\ell+1}=\mathrm{Dec}_b^{\ell}(\tilde f_b^\ell,\tilde f_a^\ell).
\end{equation}
The features from the final attention layers are processed by two point heads to predict the dense geometry and associated confidence:
\begin{equation}
    \left(\hat X_{a,a},\hat C_{a,a}\right)=\psi_a(\{\tilde f_a^\ell\}_{\ell}), \qquad \left(\hat X_{b,a},\hat C_{b,a}\right)=\psi_b(\{\tilde f_b^\ell\}_{\ell}).
\label{eq:output}
\end{equation}
Here, $\hat X_{a,a}$ and $\hat X_{b,a}$ represent 3D coordinates in the camera coordinate system of view $a$, following the convention in MAST3R~\cite{mast3r_eccv24}.

\noindent\textbf{Input normalization.} We transform the original SfM points $(S_a,S_b)$, initially in a global coordinate system with an arbitrary scale, into normalized local coordinates $(\hat S_a, \hat S_b)$ for view $a$. We first perform a conditioning-free pass (by setting $\bar g_a=\bar g_b=0$) to obtain unconditional point predictions $(\hat X_{a,a}^{\text{uncond}}, \hat X_{b,a}^{\text{uncond}})$. Additionally, we transform $S_a$ and $S_b$ into camera $a$'s coordinate system as $S_{a\rightarrow a}$ and $S_{b\rightarrow a}$ using the inverse SfM pose $T_a^{-1}$. We estimate the scale factor $s$ from depth ratios across SfM-supported pixels in the pair:
\begin{equation}
    s = \text{median}\left\{\frac{\left[\hat X_{v,a}^{\text{uncond}}(p)\right]_z}{\left[S_{v\rightarrow a}(p)\right]_z} \mid (v,p)\in\mathcal{V}_{a,b}\right\}.
\end{equation}
Here, $\mathcal{V}_{a,b}\subseteq\Omega_{\mathcal S}$ contains the pixels in views $a$ and $b$ where the predicted and scaffold depths are finite, nonzero, and have the same sign, ensuring positive scale ratios. We scale both pointmaps as $\hat S_a=s\,S_{a\rightarrow a}$ and $\hat S_b=s\,S_{b\rightarrow a}$. We then filter out points with nonfinite coordinates, nonpositive or near-zero depths, or extreme depths in camera $a$'s frame. If points are removed, we recompute the scale once from the retained points when possible and reapply the filters (see \cref{sec:app_impl_details}).

\subsubsection{Training Loss}
\label{sec:method_training}
We supervise both the unconditional and SfM-conditioned forward passes with equal weight. Following MASt3R~\cite{mast3r_eccv24}, each pass uses a confidence-weighted point map regression loss $\mathcal{L}_{\mathrm{pm}}$ and an auxiliary InfoNCE matching loss $\mathcal{L}_{\mathrm{match}}$:
\begin{equation}
    \mathcal{L}_{\mathrm{train}}
    = \frac{1}{2}\sum_{t\in\{\mathrm{uncond},\mathrm{cond}\}}
    \left(\mathcal{L}_{\mathrm{pm}}^{t} + 0.075\,\mathcal{L}_{\mathrm{match}}^{t}\right).
\end{equation}
Training uses a staged mixture of ground-truth scaffold points sampled from overlapping regions and estimated SfM points, with inference-time scale alignment for the latter (\cref{tab:training_curriculum}). We randomly drop point tokens during training to ensure the decoder remains robust to varying scaffold densities.

\subsection{SfM-Anchored Global Alignment}
\label{sec:method_alignment}
To aggregate pairwise predictions into a global 3D scene, we extend the global alignment framework from DUSt3R~\cite{wang2024dust3r} by introducing an SfM-anchoring term. Let $\mathcal{E}$ be the image-pair graph used for both pointmap prediction and global alignment. We use the complete graph for up to 16 views. For larger inputs, we use a simple sparse graph that connects neighboring views in input order and uses two hubs, the first view and the middle view, each connected to every other view. We fix the camera parameters $\{K_n, T_n\}_{n=1}^N$ provided by the SfM scaffold and optimize a dense depth map $D_v$ for each view along with an edge-local similarity transform $(P_e, \sigma_e)$ for each pair. Here, $P_e\in SE(3)$ maps the pairwise frame to the SfM world frame, and $\sigma_e > 0$ scales the transformed point. The scene-frame point for pixel $p$ is defined as:
\begin{equation}
    \chi_v(p;D_v)=T_v\,\Pi^{-1}(p,D_v(p);K_v),
    \label{eq:fixed_camera_point}
\end{equation}
where $\Pi^{-1}$ denotes pinhole back-projection. The pairwise loss $\mathcal{L}_{\mathrm{pair}}$ enforces consistency between optimized scene points and pairwise predictions, using log-confidence weights and averaging over edges and pixels:
\begin{equation}
    \begin{aligned}
    \mathcal{L}_{\mathrm{pair}} = \frac{1}{|\mathcal{E}|HW}\sum_{e=(a,b)\in\mathcal{E}}\sum_p \Bigl[&\log \hat C_{a,a}^e(p) \left\| \chi_a(p;D_a) - \sigma_e P_e\hat X_{a,a}^e(p) \right\|_2 \\
    + &\log \hat C_{b,a}^e(p) \left\| \chi_b(p;D_b) - \sigma_e P_e\hat X_{b,a}^e(p) \right\|_2\Bigr].
    \end{aligned}
    \label{eq:pair_alignment}
\end{equation}
To ensure the dense maps remain anchored to the high-precision scaffold, we add a weighted anchoring term for SfM-supported pixels:
\begin{equation}
    \mathcal{L}_{\mathrm{anc}} = \frac{1}{\sum_{(v,p)\in\Omega_{\mathcal{S}}} w_{v,p}} \sum_{(v,p)\in\Omega_{\mathcal{S}}} w_{v,p}\, \left\| \chi_v(p;D_v)-S_v(p) \right\|_2.
\end{equation}
The weights are DLT confidence scores $w_{v,p}=1/(1+e^{\mathrm{reproj}}_{v,p})$, where $e^{\mathrm{reproj}}_{v,p}$ is the reprojection error averaged over the visible 2D observations of the DLT track rasterized at pixel $p$. The final alignment is the result of:
\begin{equation}
    \min_{\{D_v\},\{P_e\},\{\sigma_e>0\}} \mathcal{L}_{\mathrm{pair}} + \lambda_{\mathrm{anc}}\mathcal{L}_{\mathrm{anc}}.
    \label{eq:anchored_ga}
\end{equation}
We initialize $D_v$ from the prediction with the highest mean confidence among edges with view $v$ as the reference, rescaled by that edge's initial alignment scale $\sigma_e$. To initialize $(P_e,\sigma_e)$, we estimate camera $b$'s pose in the predicted pair's reference frame using PnP, then align the predicted camera pair to the two fixed SfM cameras. At inference, the output is a set of dense point maps in the SfM world frame, with camera parameters preserved from the upstream SfM pipeline.

\section{Experiments}
\label{sec:experiments}
\noindent\textbf{Architecture and training.}
We train Scaffold3D on the 11 datasets as summarized in \cref{tab:training_mixture}; full architecture, optimizer, schedule, and compute details are provided in \cref{sec:app_impl_details}.

\noindent\textbf{Evaluation metrics.}
We align each predicted reconstruction to the metric ground truth with robust Umeyama alignment~\cite{revisitedsfm}, compute Euclidean point errors on ground-truth-covered pixels, and report the area under the accuracy curve (AUC) at the specified centimeter error thresholds. Our main tables use AUC@1/5/10, and AUC@M, where M is AUC@10 on monocular pixels visible in exactly one input view. All methods are evaluated on the same 512$\times$384 image canvas. For MV-dVRK, the same numeric thresholds are measured in millimeters.

\noindent\textbf{Evaluation datasets.}
We evaluate multi-view 3D reconstruction on five benchmarks. ScanNet++~\cite{yeshwanth2023scannet++} is an \dsid{in-domain} indoor benchmark, as it is included in the training data of Scaffold3D and all evaluated baselines (except MVSAnywhere, OMNI-DC, and Marigold-DC). ETH3D~\cite{schoeps2017eth3d} and Tanks\&Temples~\cite{Knapitsch2017tandt} test \dscd{cross-domain} generalization to real captures with different scene statistics. We also evaluate on two \dsod{out-of-domain} datasets: 4D-DRESS~\cite{wang20244ddress}, which contains clothed humans with masked white background, and MV-dVRK~\cite{dvrk-smv}, which contains synthetic robotic surgical scenes. \Cref{tab:dataset_support_stats} summarizes view count, GT coverage, view-overlap, and SfM-scaffold support statistics.

\noindent\textbf{Baselines.}
We compare four groups of methods. (1) Feed-forward predictors include DUSt3R~\cite{wang2024dust3r}, MASt3R~\cite{mast3r_eccv24}, MASt3R-SfM~\cite{duisterhof2025mastrsfm}, VGGT~\cite{wang2025vggt}, Pi3/Pi3X~\cite{wang2025pi3}, Pow3R~\cite{pow3r_cvpr25}, and MapAnything~\cite{keetha2025mapanything}. Our conditioning-free reference, MASt3R$^\dagger$, uses the same learned weights as Scaffold3D but sets $\bar g_a=\bar g_b=0$. (2) Monocular depth-completion baselines, OMNI-DC~\cite{zuo2025omni} and Marigold-DC~\cite{viola2024marigolddc}, complete each view independently from SfM scaffold depth samples. We evaluate both alone and with \emph{SfM overwrite}, which restores scaffold depths at supported pixels to preserve high-precision SfM geometry that completion may alter. Unconditioned two-view predictors merge pairwise outputs through global alignment with estimated cameras. Scaffold3D uses SfM-anchored alignment, as does conditioned Pow3R unless labeled \emph{DGA}. (3) For point-token propagation, we evaluate GGPT~\cite{chen2026ggpt} with different feed-forward initializers. (4) Geometry-conditioned baselines MVSAnywhere~\cite{izquierdo2025mvsanywhere}, MapAnything, Pi3X, and Pow3R are conditioned on the same dataset-specific SfM scaffold as Scaffold3D. Camera diagnostics are in \cref{tab:camera_diagnostics}. For scaffold construction, RoMaV2~\cite{edstedt2025romav2} uses the same 512$\times$384 images as the reconstruction models (see \cref{tab:fullres_romav2} for full-resolution matching).

\subsection{Main Results}

\input{tables/1_main_results.tex}

\noindent\textbf{Comparison to feed-forward and conditioned baselines.}
Scaffold3D achieves the highest all-pixel AUC@10 on the three standard benchmarks in \cref{tab:main_h16_results}: 69.6 on ScanNet++$_4$, 66.4 on ETH3D$_4$, and 60.9 on Tanks\&Temples$_4$. Its gains over point-token propagation are especially clear on monocular pixels, where geometry must be inferred beyond direct multi-view triangulation. Compared with the strongest GGPT variant on each dataset, Scaffold3D improves AUC@M from 29.0 to 40.3 on ETH3D and from 23.4 to 26.3 on Tanks\&Temples. On Tanks\&Temples, it also exceeds the best feed-forward predictor (24.0) and conditioned Pi3X (23.6). Pi3 and Pi3X lead on in-domain ScanNet++, while conditioned Pi3X reaches 40.9 AUC@M on ETH3D and CAPA reaches 27.9 on Tanks\&Temples.

On out-of-domain data, behavior depends on the quality and coverage of the SfM scaffold. Scaffold3D is best on most MV-dVRK metrics, including AUC@M, AUC@10, and AUC@5. On 4D-DRESS, only 0.9\% of pixels are monocular and 92.6\% have scaffold support (\cref{tab:dataset_support_stats}). Its clothed humans are absent from our training data, and conditioning lifts our backbone from 26.5 to 92.9 AUC@10, close to the best result (93.2).

\input{tables/2_view_count_scaling.tex}

\noindent\textbf{Effect of the number of views.}
\Cref{tab:view_count_scaling} evaluates the same scenes with 4, 8, and 16 views. More views reduce the monocular fraction, e.g., from 26.5\% to 3.9\% on ETH3D. Scaffold3D is best on all-pixel AUC@10 in eight of nine dataset/view-count combinations (second on ETH3D with 16 views) and best or second-best on monocular pixels in seven of nine. On ETH3D, its all-pixel AUC@10 lead over the best GGPT variant narrows from 6.4 to 0.1 points between 4 and 16 views. At 16 views, Scaffold3D exceeds every GGPT variant on monocular pixels across all datasets.

\begin{figure*}[!htbp]
\centering
\includegraphics[width=\linewidth]{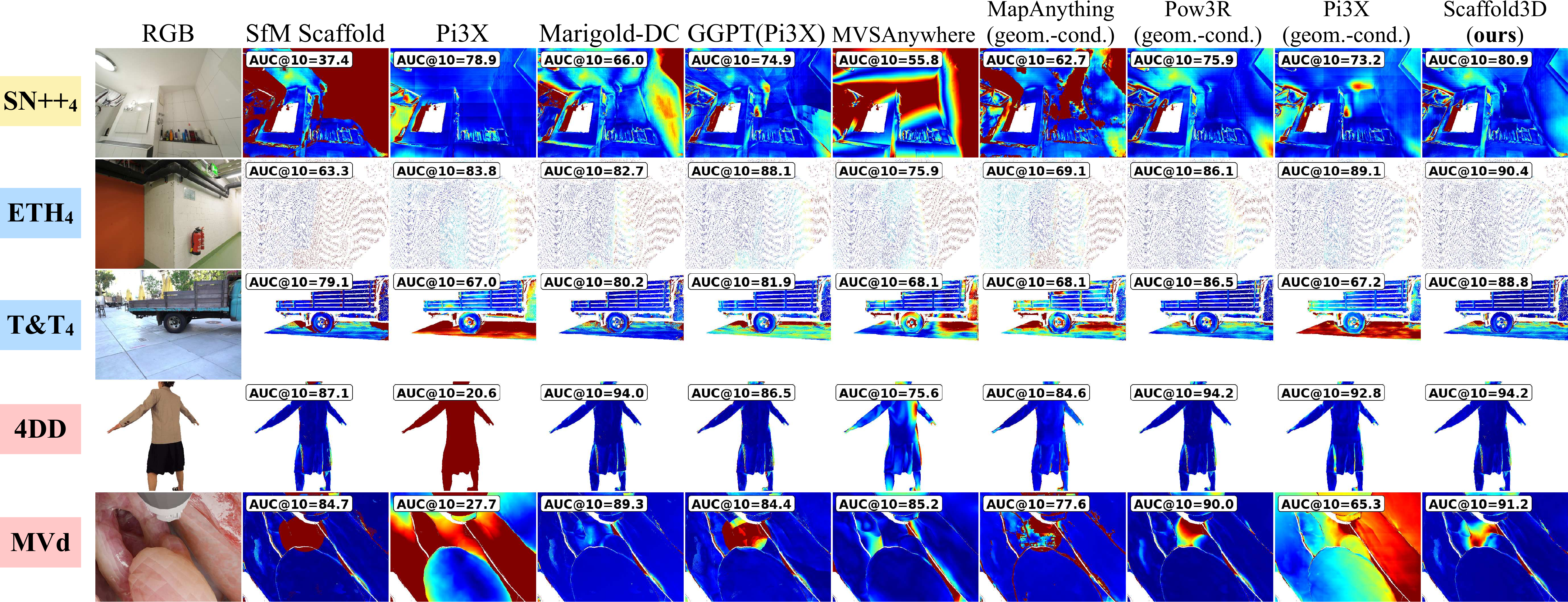}
\vspace{-0.7cm}
\caption{\textbf{Qualitative reconstruction comparison.} Rows show individual scenes and their first input view. Scaffold-conditioned methods share the scaffold. Reconstruction tiles report AUC@10.}
\label{fig:qualitative_comparison}
\end{figure*}

\noindent\textbf{Qualitative comparison.}
\Cref{fig:qualitative_comparison} compares representative reconstructions across in-domain, cross-domain, and out-of-domain scenes. These examples show that Scaffold3D preserves accurate scaffold-supported geometry while improving dense prediction beyond directly triangulated regions. Our project page provides interactive comparisons with selected baselines for all scenes in \cref{tab:main_h16_results}.

\subsection{Ablations and Analysis}

\input{tables/3_sfm_conditioning_ablation.tex}

\noindent\textbf{How should the scaffold be used?}
\Cref{tab:pipeline_fusion_ablation} compares ways to use the same SfM scaffold with our MASt3R$^\dagger$ backbone. The scaffold-only row evaluates the triangulated points themselves and therefore has 0.0 AUC@NS outside scaffold support. Naive depth completion uses scaffold depth samples to complete each view independently, reaching 39.7 AUC@NS on ETH3D$_4$. SfM-anchored DGA uses the scaffold only after dense prediction, fitting dense depth maps to fixed scaffold cameras, and reaches 39.8. GGPT propagates scaffold point tokens without refining image-token features and reaches 33.2. Scaffold3D reaches 49.1, improving over all three alternatives. It also improves over the strongest of these alternatives by 0.9 AUC@NS points on ScanNet++$_4$ and 4.5 on Tanks\&Temples$_4$. On this shared backbone, conditioning the image-token decoder on SfM points is more effective for completing geometry beyond scaffold support than these alternatives.

Additional cross-backbone comparisons in \cref{tab:posthoc_alignment} show that simple scaffold fusion with stronger predictors trained on broader data mixtures, such as Pi3, can approach Scaffold3D's accuracy and even exceed it on 4D-DRESS, where our unconditioned backbone is substantially weaker, highlighting that scaffold conditioning and the backbone's learned prior play complementary roles.

\noindent\textbf{Role of SfM anchoring.}
The unanchored-DGA ablation isolates the SfM anchoring term's contribution in \cref{eq:anchored_ga}. Anchoring improves all-pixel AUC@10 on every dataset, supporting its role in combining pairwise predictions into a reconstruction consistent with the scaffold. On MV-dVRK, removing the anchor improves accuracy outside scaffold support, but reduces overall accuracy.

\noindent\textbf{Architecture and training choices.}
The architecture ablations (\cref{tab:model_development_ablations}) support fusing point tokens with image tokens by addition at the decoder input and letting them exchange information through self-attention. Pretraining with ground-truth scaffolds also improves reconstruction accuracy. To compare our conditioning interface with existing alternatives, we fine-tune Pow3R on SfM scaffolds and use our SfM-anchored DGA, obtaining performance comparable to Scaffold3D (\cref{tab:pow3r_conditioning}), suggesting that several conditioning interfaces can learn to use the scaffold effectively.

\subsection{Scaffold Quality, Reliability, and Scale}
\label{sec:scaffold_quality}
\input{tables/16_scaffold_source_and_scaling.tex}

\noindent\textbf{How sensitive is Scaffold3D to the scaffold source?}
\Cref{tab:scaffold_quality} (left) rebuilds the scaffold from eleven sources, rerunning the full construction from each matcher's correspondences. In mean AUC@10, every source improves over the conditioning-free pass (49.4), and seven of the eleven sources improve over the per-dataset best ranked feed-forward baseline (64.6). Very sparse scaffolds already help: SuperPoint+LightGlue~\cite{superpoint,lindenberger2023lightglue} covers 0.3--2.6\% of pixels and lifts the mean to 58.3, and LoMa reaches 67.0 at similar coverage. RoMa v2 at full resolution reaches the highest mean accuracy, 76.8.

\noindent\textbf{When does conditioning help, and when is a feed-forward model preferable?}
Conditioning improves the conditioning-free pass on all ten evaluation splits, by 3.3 to 66.4 AUC@10 (\cref{tab:main_h16_results,tab:supplemental_ood_results}). On the hand-object and human datasets DexYCB and EgoBody, however, the scaffold covers only 5.5--13.4\% and 31.0\% of pixels (30.8--92.6\% on the main benchmarks), and Scaffold3D stays below Pi3, whose prior there is 12.0 to 19.6 AUC@10 stronger than our backbone's on DexYCB. With ground-truth cameras and point coordinates at the scaffold support (\cref{tab:ood_gt_scaffold}), Scaffold3D overtakes Pi3 on EgoBody-Exo (45.3 vs.\ 41.8) but not on DexYCB, supporting the complementary role of a strong backbone prior when scaffold coverage is sparse. A plain feed-forward model is thus preferable when its domain prior is much stronger and the scaffold covers little of the image.

\noindent\textbf{Scaling to more views.}
On ETH3D with 8 to 50 views (\cref{tab:scaffold_quality}, right), the sparse graph cuts total runtime by 33\% to 81\%. It matches the complete graph at 8 views ($-$0.5 AUC@10) and exceeds it at 16, 32, and 50 views ($+$2.9, $+$16.5, $+$36.3), consistent with instability in global alignment when many low-overlap pairs are included. With the sparse graph, Scaffold3D outperforms the feed-forward baselines at every view count from 4 to 50, and we use it beyond 16 views.

\subsection{Discussion and Limitations}
Scaffold3D assumes that the SfM scaffold is reliable but incomplete. We consider this realistic with modern dense matchers and SfM pipelines~\cite{edstedt2025romav2,pataki2025mpsfm,pataki2026vidmap}: when matching becomes hard, the scaffold mostly loses coverage rather than accuracy on our main benchmarks, and conditioning still helps (\cref{tab:scaffold_quality}). However, on textureless surfaces under extreme perspective change, we observed that matches can drift along the epipolar line, survive filtering, and triangulate to incorrect depths. Scaffold3D has no explicit mechanism for such points, and they influence the prediction. Conditioning still improves the same backbone there, but a feed-forward model with a stronger prior for the domain (such as Pi3) starts higher. These limitations motivate conditioning on per-point uncertainty to reduce reliance on unreliable observations and extending scaffold conditioning to stronger multi-view backbones. The method is also not real time (\cref{tab:runtime_breakdown}).

\section{Conclusion}
Scaffold3D uses SfM cameras and triangulated points as explicit conditioning signals, injects their observations as patch-aligned point tokens into image-token pointmap prediction, and fuses pairwise outputs with SfM-anchored global alignment. It achieves leading reconstruction accuracy on standard sparse-view benchmarks and MV-dVRK, with particularly clear gains over point-token propagation in low- and no-overlap regions. Together, these results show how explicit SfM geometry and learned image priors can work together to complete dense geometry beyond triangulated support.

\section*{Acknowledgements}
We thank Luigi Piccinelli, Yiming Wang, Sergey Prokudin, Nikita Karaev, and Xudong Jiang for valuable discussions and support. This study was conducted within the national ``Proficiency'' research project (No. PFFS-21-19), funded by the Swiss Innovation Agency Innosuisse in 2021 as one of 15 flagship initiatives. This work was supported by the Swiss AI Initiative under project IDs A43, A44, and A129, with computational resources provided by the Swiss National Supercomputing Centre on the Alps infrastructure. This work was also supported by the ETH Future Computing Laboratory. Yutong Chen is supported by a Google PhD Fellowship. The authors declare no competing interests.

{
\small

\input{references.tex}
}

\clearpage
\appendix

\makeatletter
\@addtoreset{figure}{section}
\@addtoreset{table}{section}
\makeatother
\renewcommand{\thefigure}{\thesection\arabic{figure}}
\renewcommand{\theHfigure}{\thesection.\arabic{figure}}
\renewcommand{\thetable}{\thesection\arabic{table}}
\renewcommand{\theHtable}{\thesection.\arabic{table}}

\label{sec:appendix}

\section{Additional Ablations}
\label{sec:app_ablations}
\Cref{tab:model_development_ablations} reports development runs that motivated the final design. Each block is a controlled comparison that isolates one design decision under comparable training conditions.

\input{tables/7_model_development_ablations.tex}

\Cref{tab:pi3x_finetune_ablation} studies Pi3X fine-tuning under our setup, comparing training duration and recipe choices before and after conditioning on the same SfM scaffold.

\input{tables/15_pi3x_finetune_ablation.tex}

\Cref{tab:pow3r_conditioning} compares scaffold-conditioning interfaces using the same SfM inputs and evaluation protocol. The Pow3R-pretrained rows show that our SfM-anchored DGA improves mean AUC@10 over plain DGA. The MASt3R-initialized controls share scaffold construction and SfM-anchored final alignment. Their training recipes differ across model families in encoder freezing and losses; the final Scaffold3D checkpoint additionally uses the staged curriculum.

\input{tables/13_pow3r_controls.tex}

\Cref{tab:point_budget_sensitivity} evaluates how many SfM points are needed for conditioning. The results saturate with a modest number of points injected into the predictor on most datasets, while the full filtered scaffold remains available for final alignment. The zero-point row supplies no scaffold points to the predictor, while retaining SfM-anchored alignment, isolating the contribution of point conditioning.

\input{tables/9_point_budget_sensitivity.tex}
\FloatBarrier

\Cref{tab:token_coverage} reports the percentage of image patches that receive a point token under the default evaluation protocol. Across evaluated scenes, this ranges from 8.9\% to 93.1\%.

\input{tables/21_token_coverage.tex}

\Cref{tab:fullres_romav2} checks whether using full-resolution 4:3 center crops for RoMaV2 matching changes the scaffold-conditioned comparison under the same evaluation protocol. The main effect is a stronger ETH3D scaffold, while the relative trends remain similar across methods.

\input{tables/11_fullres_romav2.tex}

\begin{samepage}
\Cref{tab:gt_scaffold_points} replaces scaffold points, camera poses, and intrinsics with ground truth at the original scaffold-supported pixels where ground truth is valid. This provides internally consistent geometry without adding scaffold support. Scaffold3D gains 10.0, 11.0, and 19.9 AUC@10 points on ScanNet++, ETH3D, and Tanks\&Temples, respectively, indicating room for improvement through better scaffold construction. OMNI-DC also benefits substantially, with scaffold-depth overwrite raising its Tanks\&Temples AUC@10 from 77.5 to 82.5.
\par
\end{samepage}

\input{tables/12_gt_scaffold_points.tex}

\Cref{tab:posthoc_alignment} fuses multiple feed-forward backbones with the same scaffold after prediction. Scaffold3D exceeds every post-hoc variant of its own backbone on all five datasets in AUC@10, with the smallest margin of 0.5 over GGPT on ScanNet++. Variant C on Pi3 is the strongest post-hoc baseline overall and comes within 0.2 to 1.8 AUC@10 of Scaffold3D, but Scaffold3D leads it outside scaffold support on Tanks\&Temples, 4D-DRESS, and MV-dVRK. Variant B on Pi3 and VGGT slightly exceeds Scaffold3D on 4D-DRESS (93.5 and 93.2 vs.\ 92.9), a domain where both backbones are much stronger than ours (71.4 and 73.3 vs.\ 26.5 AUC@10 before fusion). Native conditioning in Pi3X adds less to overall accuracy than per-view scale alignment and scaffold completion on the three standard benchmarks, although it is more accurate outside scaffold support on ETH3D and Tanks\&Temples.

All reported VGGT-Omega results use the original \texttt{vggt\_omega\_1b\_512.pt} checkpoint. The authors recommend a replacement for benchmarking following a potential, unconfirmed training-data contamination concern.\footnote{\def\UrlBreaks{\do\/\do\.\do\-\do\a\do\b\do\c\do\d\do\e\do\f\do\0\do\1\do\2\do\3\do\4\do\5\do\6\do\7\do\8\do\9}\url{https://github.com/facebookresearch/vggt-omega/blob/48b23c8ce72c9a8fcf987f0d7f43d8e4870759f6/reproduction.md}}

\input{tables/18_posthoc_alignment.tex}

\Cref{tab:camera_refinement} tests whether keeping the SfM cameras fixed during alignment limits reconstruction accuracy. Initializing the SfM-anchored alignment from them and refining intrinsics and extrinsics changes AUC@10 by at most 0.5.

\input{tables/20_camera_refinement.tex}

\FloatBarrier
\section{Dataset Details}
\label{sec:app_dataset_details}
\Cref{tab:dataset_support_stats} reports how much of each dataset is covered by ground truth, how often pixels are visible in one or multiple input views, and how much of the evaluation region is supported by the SfM scaffold. These statistics explain several trends in the main paper. ScanNet++, ETH3D, and Tanks\&Temples contain substantial monocular or no-scaffold regions in the four-view setting, leaving room for learned geometry propagation. In contrast, 4D-DRESS and MV-dVRK have high scaffold coverage and few monocular pixels, while DexYCB and EgoBody expose failure modes of our scaffold construction.

\input{tables/6_dataset_support_stats.tex}
\input{tables/10_training_mixture.tex}

\section{Implementation Details}
\label{sec:app_impl_details}
\noindent\textbf{Image resolution and cropping.} The training preprocessing uses landscape sizes of 512$\times$384, 512$\times$336, 512$\times$288, 512$\times$240, and 512$\times$160 pixels (width $\times$ height). For evaluation, we resize and center-crop input images to a shared 512$\times$384 canvas, applying the same transformation to depth maps and updating camera intrinsics accordingly. This can remove peripheral content from images with other aspect ratios. All reported reconstruction comparisons use this common evaluation protocol. The full-resolution matching experiment in \cref{tab:fullres_romav2} changes the images used to construct the scaffold, not the pointmap predictor's input resolution. Because development experiments showed reduced accuracy at other input resolutions, we recommend evaluating our released checkpoint at 512$\times$384 pixels, with broader resolution and aspect-ratio training, as used in Pi3 and MapAnything~\cite{wang2025pi3,keetha2025mapanything}, remaining a possible route to better generalization. Here, C denotes the shared 512$\times$384 input, while F retains the source-image pixel scale after a 4:3 center crop. For reference, retained sequences from ScanNet++, ETH3D, Tanks\&Temples, 4D-DRESS, and MV-dVRK yield F crops of 1453$\times$1090, 1695$\times$1271, 1233$\times$925, 1280$\times$960, and 1920$\times$1440 pixels, respectively. These are examples, not fixed dataset-wide resolutions, and individual matchers may resize their inputs internally.

\noindent\textbf{Scaffold filtering.} The default filtering thresholds are as follows. For bundle adjustment, the matching-score threshold is 0.6 and the cycle-error threshold is 2 pixels. For DLT, these thresholds are 0.1 and 4 pixels. Tracks must be observed in at least two views. The epipolar-error and mean reprojection-error thresholds are both 4 pixels, and at least one observing-camera pair must have a triangulation angle above $3^\circ$. We then retain points whose reprojection score $1/(1+e^{\mathrm{reproj}})$ exceeds 0.6, cycle-consistency score $1/(1+e^{\mathrm{cycle}})$ exceeds 0.7, and mean RoMaV2 matching score exceeds 0.7, where the errors are averaged over visible track observations. On 4D-DRESS and MV-dVRK, we omit the additional score-based filtering because development experiments showed that conditioning on these scaffolds improved reconstruction accuracy. We initialize the scaffold from MASt3R$^\dagger$, except on 4D-DRESS, where we use Pi3 because camera estimation by our conditioning-free predictor largely fails on this dataset (pose AUC@5$^\circ$ of 0.3, compared with 36.8 for Pi3 in \cref{tab:camera_diagnostics}). Bundle adjustment and DLT otherwise follow the GGPT pipeline~\cite{chen2026ggpt}.

\noindent\textbf{Input-depth filtering.} After scale alignment, we discard scaffold points with nonfinite coordinates or depths at most $10^{-8}$ in the reference camera's frame. We also discard depths above 20 times the 99th percentile of finite positive unconditional depths pooled across the pair. For each view separately, we then discard depths above 50 times the median of its remaining scaffold depths. If any points are removed, we recompute the median depth ratio from the retained points and update the scale if the result is finite and positive. We then apply the same filters again.

\noindent\textbf{Global alignment.} We set $\lambda_{\mathrm{anc}}=3$ and optimize for 300 iterations, using Adam with a cosine learning-rate schedule from $10^{-2}$ to $10^{-6}$.

\noindent\textbf{Architecture and training.} The SfM-conditioned pointmap predictor follows the MASt3R-Large image model configuration~\cite{mast3r_eccv24}: patch size 16, a 24-layer encoder with width 1024 and 16 heads, two 12-layer decoders with width 768 and 12 heads, and DPT-style dense pointmap prediction heads. In point MLP $\phi_{\text{pt}}$, we use a four-layer ReLU MLP and a three-layer ReLU MLP with width 512 to encode 3D coordinates and within-patch 2D offset separately, then a four-layer ReLU MLP to project both features into a single 768-dimensional feature. The patch point feature sequence is processed by four $\text{PointBlock}$ blocks with 12-head self-attention. We freeze the MASt3R image encoder throughout training. We train the remaining parameters using the four-stage curriculum in \cref{tab:training_curriculum}, with MP-SfM scaffolds in the earlier mixed stages and GGPT-pipeline scaffolds in the final stage. For computational reasons, we precompute the SfM scaffolds for all training samples before training starts. During the final fine-tuning stage, we apply point-token dropout, randomly dropping 5--72\% of the point tokens of each view.
The final training stage uses batch size 2 per GPU on 8 GPUs. We train and evaluate on NVIDIA GH200 nodes with four 96GB GPUs per node; the training runs that produced the final Scaffold3D checkpoint used less than 1K GPU-hours, while the full model-development and evaluation effort, including scaffold precomputation, baseline evaluations, ablations, and preliminary or failed runs, used on the order of tens of thousands of GPU-hours. We optimize trainable parameters with AdamW ($\beta_1=0.9$, $\beta_2=0.95$, weight decay 0.05, no weight decay for biases and normalization layers). The learning rate is linearly warmed up to $2\cdot10^{-5}$ over 1.2 epochs and then follows a half-cycle cosine decay to $5\cdot10^{-7}$.

\input{tables/23_training_curriculum.tex}

\section{Runtime}
\label{sec:app_runtime}
\Cref{tab:runtime_breakdown} breaks the inference time into scaffold construction, the two network forwards, and SfM-anchored global alignment, from 2 to 50 views and for both pair graphs. Scaffold construction is unchanged between the two graphs, retaining the same initialization and all-pairs matching. The network passes are relatively small compared with the final alignment at higher view counts. With the complete graph, the pairwise stages grow quadratically with the number of views; the sparse graph keeps 17, 41, 89, and 143 of 28, 120, 496, and 1225 pairs at 8, 16, 32, and 50 views and cuts total runtime by 33\% to 81\%. \Cref{tab:scaling_full} reports AUC@1/5/10 for both graphs. Beyond 16 views, the complete graph's many low-overlap pairs destabilize the global alignment (on 7 of 16 scenes at 32 views), while the sparse graph stays stable and outperforms the feed-forward baselines at every view count from 4 to 50 views and every metric.

\input{tables/8_runtime_breakdown.tex}
\input{tables/22_scaling_full.tex}

\FloatBarrier
\section{Additional Evaluation Results}
\label{sec:app_additional_results}
\begin{figure*}[!htbp]
    \centering
    \includegraphics[width=\linewidth]{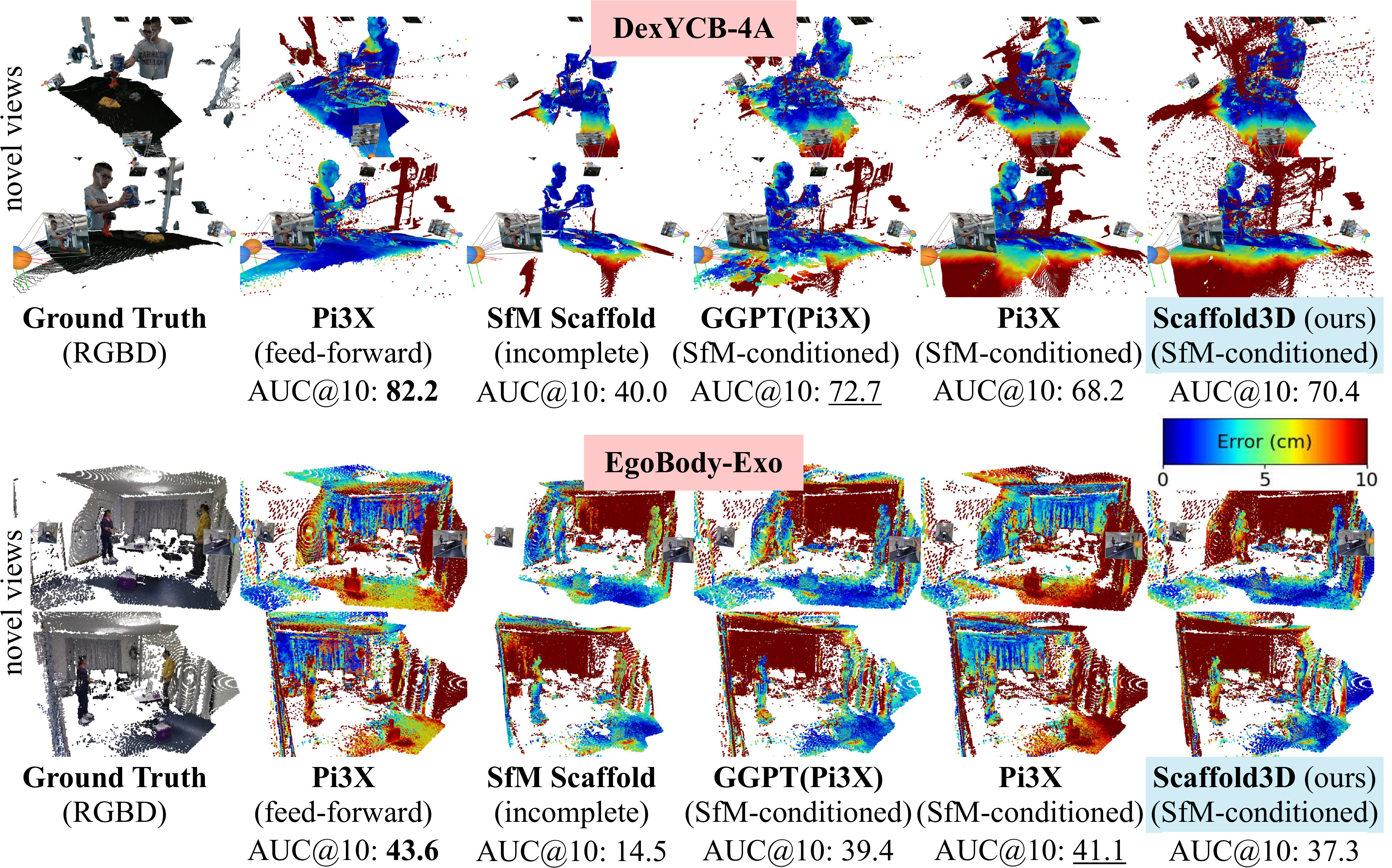}
    \caption{\textbf{Failure mode under inaccurate SfM scaffolds.} We visualize DexYCB-4A and EgoBody-Exo examples from two viewpoints each. Columns show ground truth RGB-D geometry, feed-forward Pi3X, the incomplete SfM scaffold, GGPT(Pi3X), Pi3X conditioned on the scaffold, and Scaffold3D. Predicted points are colored by metric 3D error, clipped at 10\,cm, where dark blue indicates low error and red indicates high error. These examples illustrate limited scaffold coverage and geometric errors that can affect conditioned predictions. Aggregate reconstruction results, camera diagnostics, and ground-truth-scaffold tests are reported in \cref{tab:supplemental_ood_results,tab:supplemental_camera_diagnostics,tab:ood_gt_scaffold}.}
    \label{fig:supp_ood_scaffold_failure}
\end{figure*}

\Cref{tab:supplemental_ood_results} evaluates Scaffold3D on DexYCB and EgoBody, which contain hand-object interactions and human-centric scenes. Despite limited and partly inaccurate scaffold support, conditioning improves AUC@10 over our conditioning-free backbone by 3.3--8.3 points on all five splits, although Pi3 remains more accurate. Pi3X benefits from conditioning on three splits and loses accuracy on two. \Cref{tab:supplemental_camera_diagnostics} reports camera accuracy, and \cref{fig:supp_ood_scaffold_failure} shows representative reconstructions.

\input{tables/4_supplemental_ood_results.tex}

\input{tables/14_supplemental_camera_diagnostics.tex}

Examining the retained DexYCB matches revealed an egregious failure case involving textureless surfaces and extreme viewpoint changes: incorrect correspondences remained close to the epipolar lines and therefore survived epipolar filtering. Our diagnostics found median displacements of 2 to 6\,px from the correct correspondences, predominantly along the epipolar line, leading to incorrect triangulated depths.

To separate camera errors from errors in the triangulated points, we substitute ground-truth cameras and points, separately and together (\cref{tab:ood_gt_scaffold}). On DexYCB, holding the cameras at their ground-truth values and replacing the triangulated points with ground-truth points improves AUC@10 by up to 7.5 points, confirming that triangulation errors limit reconstruction accuracy. On EgoBody-Exo, correcting the cameras alone or the points alone is enough for Scaffold3D to outperform both Pi3 and Pi3X. On DexYCB, however, correcting both still leaves a gap to Pi3. This supports the importance of a strong backbone prior when scaffold coverage remains sparse, even if the available geometry is accurate.

\input{tables/19_ood_gt_scaffold.tex}

\FloatBarrier
\section{Scaffold Quality and Reliability}
\label{sec:app_scaffold_quality}
\noindent\textbf{Scaffold sources.} \Cref{tab:scaffold_source_full} gives the per-dataset results of the scaffold-source study in \cref{tab:scaffold_quality} together with the quality of each scaffold. For every matcher, the full scaffold construction (track building, bundle adjustment, DLT triangulation, and the same filtering) is rerun from that matcher's correspondences. In this study, accuracy tracks scaffold camera and point accuracy more closely than coverage: the sparse LoMa scaffolds beat the much denser UFM scaffolds. Sparse but accurate scaffolds still help: the LoMa scaffolds cover at most 3.5\% of pixels but reach 67.0 mean AUC@10 at full resolution. The mean hides some per-dataset losses: on ScanNet++, the SuperPoint+LightGlue, UFM, RoMa, and MASt3R scaffolds fall below the conditioning-free pass (65.6), and on MV-dVRK, the SuperPoint+LightGlue and UFM (canvas) scaffolds fall below it (70.1).

\input{tables/17_scaffold_source_full.tex}

The main reconstruction tables compare methods under a shared scaffold whenever geometry conditioning is used (except GGPT variants, which construct scaffolds with their named initializers). \Cref{tab:camera_diagnostics} separates camera quality from dense reconstruction quality: many conditioned rows inherit the same cameras, so differences in point accuracy are primarily due to how methods use the scaffold rather than due to different camera estimates.

\input{tables/5_camera_diagnostics.tex}

\noindent\textbf{Evaluation variance.}
\label{sec:app_variance}
We report a single run per configuration and measured how much the numbers vary. Rerunning the same code and configuration in fresh caches changes AUC@10 on ETH3D$_4$ by 0.003, due to non-deterministic attention kernels. With the network predictions cached, five different random seeds for the latent initialization of the global alignment give identical AUC@10 on ETH3D$_4$ and MV-dVRK and a standard deviation of 0.05 on Tanks\&Temples$_4$. An update of the evaluation code that fixes the random seeds per scene for point-token sampling and for the alignment initialization moves a few scenes to a different solution. With the current code, Scaffold3D reads 23.5/52.9/66.1/40.2 (AUC@1/5/10/M) on ETH3D$_4$ and 71.8/86.1/90.7/44.1 on MV-dVRK, against 25.6/53.6/66.4/40.3 and 68.4/85.4/90.4/44.0 in \cref{tab:main_h16_results}, while all baselines we re-ran reproduce to within 0.02. In these checks, AUC@10 and AUC@M vary by at most about 0.3, but AUC@1 can move by up to 3.4 points, and margins below 0.5 AUC@10 or below a few AUC@1 points should not be read as reliable advantages. The scaffold-source study and the sparse-graph, 32-view, and 50-view results in \cref{tab:scaffold_quality} were evaluated with the current code.

\FloatBarrier
\section{Broader Impacts}
\label{sec:app_impact}
More accurate multi-view 3D reconstruction can benefit robotics, AR/VR, digital content creation, and surgical and embodied perception. Like other reconstruction methods, it could also be used to reconstruct people or private spaces without their consent, and improved reconstruction accuracy could make such privacy-invasive uses more effective.

\end{document}

%% file: tables/1_main_results.tex
\begin{table*}[t]
\setlength{\tabcolsep}{2.1pt}
\centering
\caption{\textbf{Multi-view 3D reconstruction.} Scaffold3D achieves the highest all-pixel AUC@10 on the \dsid{in-domain} and \dscd{cross-domain} benchmarks and on \dsod{out-of-domain} MV-dVRK, while remaining competitive on 4D-DRESS, where the SfM scaffold already covers most evaluated pixels. On monocular pixels, it exceeds the strongest GGPT variant by 11.3 AUC points on ETH3D and 2.9 on Tanks\&Temples. Each dataset block reports AUC@1/5/10/M. IT, PT, and D indicate image tokens, point tokens, and depth-map conditioning. Geometry-conditioned baselines use the same dataset-specific SfM scaffold. \textsuperscript{\ddag}Comparisons added after submission, shown in grey and excluded from the primary ranking marks. MASt3R\textsuperscript{\textdagger} is the conditioning-free pass of our predictor (same network and weights). Rows labeled ``+ SfM overwrite'' replace completed depths with scaffold depths wherever available, preserving the scaffold's high-precision geometry. \textsuperscript{\S}MVSAnywhere uses the scaffold cameras and depth ranges derived from the scaffold points, but does not consume point tokens. Best and second-best results per metric and dataset are \B{bold} and \U{underlined}.}
\resizebox{\linewidth}{!}{%
\begin{tabular}{lccc crrrr crrrr crrrr crrrr crrrr}
\toprule
\multirow[b]{3}{*}{Method} & \multirow[b]{3}{*}{IT} & \multirow[b]{3}{*}{PT} & \multirow[b]{3}{*}{D} & & \multicolumn{4}{c}{\dsid{ScanNet++$_4$}} & & \multicolumn{4}{c}{\dscd{ETH3D$_4$}} & & \multicolumn{4}{c}{\dscd{Tanks\&Temples$_4$}} & & \multicolumn{4}{c}{\dsod{4D-DRESS}} & & \multicolumn{4}{c}{\dsod{MV-dVRK}} \\
\cmidrule(lr){6-9} \cmidrule(lr){11-14} \cmidrule(lr){16-19} \cmidrule(lr){21-24} \cmidrule(lr){26-29}
 & & & & & \multicolumn{4}{c}{\textbf{AUC@1/5/10/M}} & & \multicolumn{4}{c}{\textbf{AUC@1/5/10/M}} & & \multicolumn{4}{c}{\textbf{AUC@1/5/10/M}} & & \multicolumn{4}{c}{\textbf{AUC@1/5/10/M}} & & \multicolumn{4}{c}{\textbf{AUC@1/5/10/M}} \\
 & & & & & \multicolumn{4}{c}{\emph{M pixels: 24.0\%}} & & \multicolumn{4}{c}{\emph{M pixels: 26.5\%}} & & \multicolumn{4}{c}{\emph{M pixels: 10.2\%}} & & \multicolumn{4}{c}{\emph{M pixels: 0.9\%}} & & \multicolumn{4}{c}{\emph{M pixels: 0.4\%}} \\
\midrule
\addlinespace[0.15em]
\multicolumn{29}{l}{\emph{Feed-forward predictors}} \\
\midrule
\textcolor{black!55}{VGGT-Omega$^\ddagger$~\cite{wang2025vggt}} & \textcolor{black!55}{\cmark} & \textcolor{black!55}{\xmark} & \textcolor{black!55}{\xmark} & & \textcolor{black!55}{3.2} & \textcolor{black!55}{21.9} & \textcolor{black!55}{38.8} & \textcolor{black!55}{26.6} & & \textcolor{black!55}{7.0} & \textcolor{black!55}{35.4} & \textcolor{black!55}{51.8} & \textcolor{black!55}{43.1} & & \textcolor{black!55}{7.5} & \textcolor{black!55}{30.9} & \textcolor{black!55}{45.7} & \textcolor{black!55}{23.1} & & \textcolor{black!55}{34.0} & \textcolor{black!55}{77.5} & \textcolor{black!55}{88.2} & \textcolor{black!55}{71.1} & & \textcolor{black!55}{16.1} & \textcolor{black!55}{51.6} & \textcolor{black!55}{68.1} & \textcolor{black!55}{29.1} \\
MASt3R-SfM~\cite{duisterhof2025mastrsfm} & \cmark & \xmark & \xmark &  & 12.3 & 32.8 & 43.9 & 12.1 &  & 13.9 & 36.8 & 49.3 & 24.4 &  & 19.6 & 41.0 & 52.9 & 13.4 &  & 56.6 & 72.0 & 76.9 & 27.7 &  & 51.4 & 68.7 & 75.8 & 24.7 \\
MASt3R~\cite{mast3r_eccv24} & \cmark & \xmark & \xmark &  & 8.0 & 33.6 & 49.8 & 37.9 &  & 6.7 & 24.5 & 36.7 & 23.5 &  & 8.5 & 31.0 & 45.0 & 18.3 &  & 22.2 & 52.4 & 64.7 & 32.1 &  & 10.0 & 33.1 & 47.4 & 13.4 \\
DUSt3R~\cite{wang2024dust3r} & \cmark & \xmark & \xmark &  & 9.4 & 34.9 & 50.4 & 39.6 &  & 5.1 & 21.1 & 33.3 & 21.7 &  & 5.8 & 24.3 & 37.5 & 14.2 &  & 7.8 & 31.6 & 46.1 & 19.1 &  & 14.0 & 41.6 & 56.4 & 19.1 \\
Pow3R~\cite{pow3r_cvpr25} & \cmark & \xmark & \xmark &  & 10.2 & 36.7 & 52.4 & 42.9 &  & 6.3 & 20.4 & 31.8 & 18.4 &  & 5.2 & 22.1 & 34.8 & 12.1 &  & 7.6 & 32.2 & 45.9 & 20.6 &  & 9.8 & 32.4 & 47.1 & 15.5 \\
VGGT~\cite{wang2025vggt} & \cmark & \xmark & \xmark &  & 8.2 & 38.5 & 55.6 & 44.7 &  & 6.4 & 26.3 & 41.4 & 28.5 &  & 6.6 & 29.8 & 43.3 & 20.3 &  & 13.1 & 54.2 & 73.3 & 45.9 &  & 9.7 & 33.7 & 49.1 & 15.0 \\
MapAnything~\cite{keetha2025mapanything} & \cmark & \xmark & \xmark &  & 11.7 & 45.2 & 61.0 & 53.2 &  & 1.7 & 11.2 & 21.2 & 17.1 &  & 2.9 & 17.2 & 30.2 & 11.8 &  & 3.7 & 22.3 & 40.2 & 25.8 &  & 2.6 & 16.5 & 30.8 & 5.1 \\
MASt3R$^\dagger$ & \cmark & \xmark & \xmark &  & 17.1 & 50.2 & 65.6 & 54.9 &  & 6.2 & 26.4 & 40.6 & 28.1 &  & 9.2 & 30.8 & 44.4 & 19.5 &  & 3.0 & 15.0 & 26.5 & 7.8 &  & 26.0 & 56.5 & 70.1 & 28.7 \\
Pi3X~\cite{wang2025pi3} & \cmark & \xmark & \xmark &  & 15.0 & 50.8 & 66.5 & \U{56.6} &  & 9.5 & 33.3 & 47.3 & 35.6 &  & 5.4 & 28.9 & 44.8 & 22.3 &  & 3.7 & 15.4 & 26.5 & 17.5 &  & 4.4 & 21.0 & 35.3 & 8.3 \\
Pi3~\cite{wang2025pi3} & \cmark & \xmark & \xmark &  & 16.7 & 52.0 & 68.0 & \B{57.8} &  & 6.2 & 29.2 & 44.1 & 31.5 &  & 5.6 & 31.0 & 46.6 & \U{24.0} &  & 10.5 & 50.4 & 71.4 & 49.4 &  & 18.3 & 51.9 & 67.0 & 27.2 \\
\midrule
\addlinespace[0.15em]
\multicolumn{29}{l}{\emph{Monocular depth completion}} \\
\midrule
OMNI-DC~\cite{zuo2025omni} & \cmark & \xmark & \cmark &  & 19.3 & 41.6 & 54.2 & 20.1 &  & 17.0 & 36.8 & 48.6 & 12.6 &  & 16.3 & 37.3 & 49.4 & 10.9 &  & 50.0 & 78.8 & 87.7 & 44.7 &  & 50.2 & 75.0 & 83.5 & 32.7 \\
OMNI-DC~\cite{zuo2025omni} + SfM overwrite & \cmark & \xmark & \cmark &  & 22.5 & 44.5 & 56.1 & 20.0 &  & \U{25.6} & 47.0 & 57.1 & 12.7 &  & \U{26.1} & \U{48.8} & 59.5 & 11.0 &  & 74.8 & 88.4 & 92.7 & 45.5 &  & 66.1 & 82.9 & 88.0 & 32.9 \\
Marigold-DC~\cite{viola2024marigolddc} & \cmark & \xmark & \cmark &  & 19.7 & 42.9 & 55.3 & 27.8 &  & 17.4 & 39.1 & 52.2 & 20.9 &  & 8.7 & 28.1 & 43.4 & 15.7 &  & 69.8 & 87.8 & 92.6 & 51.5 &  & 64.3 & 83.2 & 88.9 & 40.7 \\
Marigold-DC~\cite{viola2024marigolddc} + SfM overwrite & \cmark & \xmark & \cmark &  & 21.8 & 44.4 & 56.3 & 27.7 &  & \U{25.6} & 48.5 & 59.6 & 20.8 &  & 25.2 & 47.1 & 58.1 & 15.3 &  & \U{76.2} & \U{89.1} & \U{93.1} & 50.4 &  & \U{68.6} & \U{84.7} & \U{89.7} & 40.4 \\
\textcolor{black!55}{CAPA (VGGT)$^\ddagger$~\cite{ke2026capa}} & \textcolor{black!55}{\cmark} & \textcolor{black!55}{\xmark} & \textcolor{black!55}{\cmark} & & \textcolor{black!55}{22.6} & \textcolor{black!55}{52.1} & \textcolor{black!55}{66.4} & \textcolor{black!55}{55.0} & & \textcolor{black!55}{20.0} & \textcolor{black!55}{46.3} & \textcolor{black!55}{60.0} & \textcolor{black!55}{41.1} & & \textcolor{black!55}{13.2} & \textcolor{black!55}{38.0} & \textcolor{black!55}{52.7} & \textcolor{black!55}{27.9} & & \textcolor{black!55}{38.7} & \textcolor{black!55}{68.1} & \textcolor{black!55}{79.8} & \textcolor{black!55}{55.9} & & \textcolor{black!55}{45.5} & \textcolor{black!55}{72.6} & \textcolor{black!55}{82.9} & \textcolor{black!55}{39.8} \\
\textcolor{black!55}{CAPA (VGGT)$^\ddagger$~\cite{ke2026capa} + SfM overwrite} & \textcolor{black!55}{\cmark} & \textcolor{black!55}{\xmark} & \textcolor{black!55}{\cmark} & & \textcolor{black!55}{25.1} & \textcolor{black!55}{54.2} & \textcolor{black!55}{67.9} & \textcolor{black!55}{56.3} & & \textcolor{black!55}{27.3} & \textcolor{black!55}{54.0} & \textcolor{black!55}{66.0} & \textcolor{black!55}{41.3} & & \textcolor{black!55}{27.6} & \textcolor{black!55}{49.3} & \textcolor{black!55}{60.9} & \textcolor{black!55}{27.6} & & \textcolor{black!55}{75.3} & \textcolor{black!55}{88.0} & \textcolor{black!55}{92.4} & \textcolor{black!55}{55.4} & & \textcolor{black!55}{68.8} & \textcolor{black!55}{84.5} & \textcolor{black!55}{89.8} & \textcolor{black!55}{41.8} \\
\midrule
\addlinespace[0.15em]
\multicolumn{29}{l}{\emph{Point-token propagation}} \\
\midrule
GGPT (VGGT)~\cite{chen2026ggpt} & \xmark & \cmark & \xmark &  & 16.2 & 48.0 & 63.2 & 44.4 &  & 14.0 & 42.2 & 56.6 & 26.7 &  & 14.8 & 42.6 & 55.1 & 17.2 &  & 69.3 & 86.4 & 91.9 & 49.7 &  & 56.6 & 76.3 & 83.0 & 27.5 \\
GGPT (Pi3X)~\cite{chen2026ggpt} & \xmark & \cmark & \xmark &  & 16.9 & 50.0 & 65.3 & 51.0 &  & 15.5 & 45.4 & 60.0 & 29.0 &  & 17.0 & 43.7 & 56.1 & 20.8 &  & 63.7 & 82.6 & 87.6 & 21.8 &  & 55.7 & 74.3 & 81.0 & 23.0 \\
GGPT (Pi3)~\cite{chen2026ggpt} & \xmark & \cmark & \xmark &  & 14.0 & 50.2 & 66.1 & 54.9 &  & 7.4 & 33.3 & 50.0 & 27.3 &  & 10.4 & 36.6 & 51.9 & 23.4 &  & 72.4 & 87.8 & 92.7 & 53.4 &  & 61.7 & 81.4 & 87.9 & 35.8 \\
GGPT (MASt3R$^\dagger$)~\cite{chen2026ggpt} & \xmark & \cmark & \xmark &  & \U{25.0} & \U{55.8} & \U{69.4} & 53.5 &  & 15.6 & 44.6 & 58.6 & 21.6 &  & 21.6 & 47.9 & \U{60.1} & 17.3 &  & 49.0 & 59.9 & 64.4 & 11.2 &  & \B{68.8} & 84.1 & 89.1 & 38.8 \\
\midrule
\addlinespace[0.15em]
\multicolumn{29}{l}{\emph{Geometry-conditioned predictors}} \\
\midrule
MVSAnywhere~\cite{izquierdo2025mvsanywhere}\textsuperscript{\S} & \cmark & \xmark & \xmark &  & 10.2 & 27.5 & 40.1 & 23.6 &  & 11.3 & 30.3 & 41.8 & 21.5 &  & 9.9 & 29.8 & 42.9 & 16.4 &  & 30.2 & 63.1 & 77.4 & \B{57.9} &  & 48.9 & 71.0 & 79.8 & 31.2 \\
MapAnything~\cite{keetha2025mapanything} & \cmark & \xmark & \cmark &  & 12.4 & 37.9 & 50.4 & 30.9 &  & 6.7 & 26.3 & 39.5 & 21.4 &  & 9.3 & 31.1 & 44.8 & 15.5 &  & 39.0 & 74.5 & 82.9 & 37.3 &  & 49.8 & 72.9 & 79.2 & 17.6 \\
Pow3R~\cite{pow3r_cvpr25} (SfM-anchored DGA) & \cmark & \xmark & \cmark &  & 23.4 & 49.4 & 61.7 & 42.1 &  & \B{25.7} & \U{50.9} & \U{62.3} & 27.2 &  & \U{26.1} & 48.1 & 59.3 & 21.0 &  & \B{76.9} & \B{89.2} & \B{93.2} & 41.2 &  & 65.2 & 83.5 & 88.6 & 36.5 \\
Pow3R~\cite{pow3r_cvpr25} (DGA) & \cmark & \xmark & \cmark &  & 20.0 & 49.1 & 62.2 & 42.5 &  & 18.7 & 45.3 & 58.4 & 25.7 &  & 18.9 & 43.0 & 56.0 & 20.8 &  & 59.2 & 84.6 & 90.9 & 39.5 &  & 67.6 & 83.4 & 88.3 & 36.7 \\
Pi3X~\cite{wang2025pi3} & \cmark & \xmark & \cmark &  & 20.4 & 54.5 & 68.5 & 55.9 &  & 12.4 & 42.1 & 57.8 & \B{40.9} &  & 11.6 & 37.7 & 52.7 & 23.6 &  & 64.7 & 86.4 & 91.9 & \U{57.6} &  & 66.7 & 84.2 & 89.5 & \U{42.9} \\
\rowcolor{cyan!16} Scaffold3D (ours) & \cmark & \cmark & \xmark &  & \B{25.7} & \B{56.5} & \B{69.6} & 56.3 &  & \U{25.6} & \B{53.6} & \B{66.4} & \U{40.3} &  & \B{26.4} & \B{49.2} & \B{60.9} & \B{26.3} &  & 74.4 & 88.6 & 92.9 & 51.2 &  & 68.4 & \B{85.4} & \B{90.4} & \B{44.0} \\
\bottomrule
\vspace{-0.5cm}
\end{tabular}
}
\label{tab:main_h16_results}
\end{table*}

%% file: tables/2_view_count_scaling.tex
\begin{table*}[t]
\setlength{\tabcolsep}{2.1pt}
\centering
\caption{\textbf{View-count scaling.} We evaluate the same scenes with 4, 8, and 16 input views. Each dataset block reports AUC@10/M. Additional views reduce the fraction of no-overlap pixels. Scaffold3D remains best or comparable as the number of views varies.}
\resizebox{\linewidth}{!}{%
\begin{tabular}{l cccc cccc cccc cccc cccc cccc}
\toprule
\multirow[b]{3}{*}{Method} & & \multicolumn{7}{c}{\dsid{ScanNet++}} & & \multicolumn{7}{c}{\dscd{ETH3D}} & & \multicolumn{7}{c}{\dscd{Tanks\&Temples}} \\
\cmidrule(lr){3-9} \cmidrule(lr){11-17} \cmidrule(lr){19-25}
& & \multicolumn{3}{c}{\textbf{All points}} & & \multicolumn{3}{c}{\textbf{Monocular}} & & \multicolumn{3}{c}{\textbf{All points}} & & \multicolumn{3}{c}{\textbf{Monocular}} & & \multicolumn{3}{c}{\textbf{All points}} & & \multicolumn{3}{c}{\textbf{Monocular}} \\
 &  & 4 & 8 & 16 &  & 4 & 8 & 16 &  & 4 & 8 & 16 &  & 4 & 8 & 16 &  & 4 & 8 & 16 &  & 4 & 8 & 16 \\
\midrule
\emph{Monocular pixels (\%)} &  & \emph{--} & \emph{--} & \emph{--} &  & \emph{24.0} & \emph{9.1} & \emph{2.9} &  & \emph{--} & \emph{--} & \emph{--} &  & \emph{26.5} & \emph{12.9} & \emph{3.9} &  & \emph{--} & \emph{--} & \emph{--} &  & \emph{10.2} & \emph{4.7} & \emph{2.7} \\
\midrule
\addlinespace[0.15em]
\multicolumn{25}{l}{\emph{Feed-forward predictors}} \\
\midrule
DUSt3R~\cite{wang2024dust3r} &  & 50.4 & 49.6 & 51.0 &  & 39.6 & 36.7 & 38.9 &  & 33.3 & 27.6 & 25.9 &  & 21.7 & 14.5 & 9.2 &  & 37.5 & 34.6 & 30.9 &  & 14.2 & 6.1 & 5.5 \\
MASt3R~\cite{mast3r_eccv24} &  & 49.8 & 49.8 & 50.0 &  & 37.9 & 40.7 & 39.3 &  & 36.7 & 31.1 & 28.9 &  & 23.5 & 15.3 & 19.0 &  & 45.0 & 43.4 & 40.8 &  & 18.3 & 8.9 & 9.8 \\
Pow3R~\cite{pow3r_cvpr25} &  & 52.4 & 52.3 & 53.6 &  & 42.9 & 39.6 & 42.2 &  & 31.8 & 27.4 & 26.6 &  & 18.4 & 13.3 & 12.8 &  & 34.8 & 32.5 & 28.6 &  & 12.1 & 7.0 & 7.8 \\
MapAnything~\cite{keetha2025mapanything} &  & 61.0 & 60.0 & 61.3 &  & 53.2 & 53.1 & 52.1 &  & 21.2 & 17.0 & 13.5 &  & 17.1 & 12.4 & 9.9 &  & 30.2 & 26.9 & 24.9 &  & 11.8 & 8.0 & 5.2 \\
VGGT~\cite{wang2025vggt} &  & 55.6 & 56.6 & 57.9 &  & 44.7 & 48.1 & 44.4 &  & 41.4 & 37.1 & 31.2 &  & 28.5 & 19.9 & 19.8 &  & 43.3 & 44.5 & 41.1 &  & 20.3 & 14.7 & 12.9 \\
\textcolor{black!55}{VGGT-Omega$^\ddagger$~\cite{wang2025vggt}} & & \textcolor{black!55}{38.8} & \textcolor{black!55}{38.2} & \textcolor{black!55}{37.6} & & \textcolor{black!55}{26.6} & \textcolor{black!55}{25.3} & \textcolor{black!55}{28.1} & & \textcolor{black!55}{51.8} & \textcolor{black!55}{48.3} & \textcolor{black!55}{42.9} & & \textcolor{black!55}{43.1} & \textcolor{black!55}{35.3} & \textcolor{black!55}{29.7} & & \textcolor{black!55}{45.7} & \textcolor{black!55}{45.6} & \textcolor{black!55}{41.3} & & \textcolor{black!55}{23.1} & \textcolor{black!55}{18.6} & \textcolor{black!55}{13.9} \\
Pi3~\cite{wang2025pi3} &  & 68.0 & 70.2 & 72.4 &  & \B{57.8} & \U{61.7} & \U{61.8} &  & 44.1 & 35.7 & 34.9 &  & 31.5 & 21.2 & 20.3 &  & 46.6 & 47.0 & 44.0 &  & \U{24.0} & 18.1 & 15.8 \\
Pi3X~\cite{wang2025pi3} &  & 66.5 & 70.1 & 72.0 &  & \U{56.6} & \B{62.1} & 61.7 &  & 47.3 & 42.3 & 38.9 &  & 35.6 & 27.5 & 23.0 &  & 44.8 & 44.6 & 42.6 &  & 22.3 & 18.6 & 16.4 \\
MASt3R$^\dagger$ &  & 65.6 & 66.4 & 66.8 &  & 54.9 & 57.0 & 52.8 &  & 40.6 & 36.4 & 34.8 &  & 28.1 & 19.8 & 15.9 &  & 44.4 & 44.7 & 39.7 &  & 19.5 & 10.2 & 13.2 \\
\midrule
\addlinespace[0.15em]
\multicolumn{25}{l}{\emph{Point-token propagation}} \\
\midrule
GGPT (VGGT)~\cite{chen2026ggpt} &  & 63.2 & 70.2 & 74.8 &  & 44.4 & 51.6 & 48.6 &  & 56.6 & 58.0 & 62.8 &  & 26.7 & 20.0 & 22.8 &  & 55.1 & 61.2 & 59.7 &  & 17.2 & 17.2 & 17.7 \\
GGPT (Pi3)~\cite{chen2026ggpt} &  & 66.1 & 71.0 & 74.6 &  & 54.9 & 58.7 & 58.6 &  & 50.0 & 49.5 & 52.4 &  & 27.3 & 22.6 & 28.4 &  & 51.9 & 57.4 & 57.9 &  & 23.4 & 19.6 & 19.7 \\
GGPT (Pi3X)~\cite{chen2026ggpt} &  & 65.3 & 71.2 & 75.4 &  & 51.0 & 53.3 & 52.0 &  & 60.0 & 60.6 & 63.8 &  & 29.0 & 24.0 & 25.3 &  & 56.1 & 61.7 & 61.1 &  & 20.8 & 20.6 & 21.7 \\
GGPT (MASt3R$^\dagger$)~\cite{chen2026ggpt} &  & \U{69.4} & \U{74.9} & \U{78.8} &  & 53.5 & 55.7 & 54.6 &  & 58.6 & 60.2 & 66.9 &  & 21.6 & 19.9 & 25.9 &  & \U{60.1} & \U{67.6} & \U{67.0} &  & 17.3 & 17.0 & 19.2 \\
\midrule
\addlinespace[0.15em]
\multicolumn{25}{l}{\emph{Geometry-conditioned predictors}} \\
\midrule
MVSAnywhere~\cite{izquierdo2025mvsanywhere} &  & 40.1 & 51.5 & 58.1 &  & 23.6 & 31.5 & 35.7 &  & 41.8 & 48.4 & 52.5 &  & 21.5 & 21.5 & 27.5 &  & 42.9 & 47.2 & 47.7 &  & 16.4 & 17.7 & 20.2 \\
MapAnything~\cite{keetha2025mapanything} &  & 50.4 & 63.5 & 69.3 &  & 30.9 & 35.6 & 40.8 &  & 39.5 & 43.5 & 40.0 &  & 21.4 & 16.0 & 15.3 &  & 44.8 & 53.5 & 51.0 &  & 15.5 & 15.1 & 14.4 \\
Pi3X~\cite{wang2025pi3} &  & 68.5 & 73.8 & 77.4 &  & 55.9 & \U{61.7} & 61.3 &  & 57.8 & 56.1 & 59.2 &  & \B{40.9} & \U{34.3} & \B{37.3} &  & 52.7 & 56.7 & 55.3 &  & 23.6 & \B{23.0} & \U{21.8} \\
Pow3R~\cite{pow3r_cvpr25} &  & 61.7 & 72.3 & 77.1 &  & 42.1 & 49.8 & 50.4 &  & \U{62.3} & \U{64.9} & \B{67.6} &  & 27.2 & 24.3 & 28.0 &  & 59.3 & 66.9 & 66.4 &  & 21.0 & 15.2 & 20.0 \\
\rowcolor{cyan!16} Scaffold3D (ours) &  & \B{69.6} & \B{75.7} & \B{79.0} &  & 56.3 & 60.6 & \B{62.7} &  & \B{66.4} & \B{67.4} & \U{67.0} &  & \U{40.3} & \B{36.1} & \U{36.6} &  & \B{60.9} & \B{69.5} & \B{68.2} &  & \B{26.3} & \U{21.3} & \B{25.1} \\
\bottomrule
\vspace{-0.8cm}
\end{tabular}
}
\label{tab:view_count_scaling}
\end{table*}

%% file: tables/3_sfm_conditioning_ablation.tex
\begin{table*}[t]
\setlength{\tabcolsep}{2.1pt}
\centering
\caption{\textbf{SfM conditioning ablation.} We compare simple, training-free ways to use the SfM scaffold with MASt3R$^\dagger$ (the conditioning-free pass of Scaffold3D) against our learned Scaffold3D predictor. Each dataset block reports AUC@10 overall, outside SfM support (NS), on monocular pixels (M), and on non-monocular pixels (NM). M/NM labels are approximated from GT reprojection. \na{} is used for methods not using SfM. DGA denotes DUSt3R global alignment. Unanchored DGA removes the final SfM anchoring term. See \cref{tab:dataset_support_stats} for the overlap and SfM-support ratios.}
\resizebox{\linewidth}{!}{%
\begin{tabular}{l crrrr crrrr crrrr crrrr crrrr}
\toprule
\multirow[b]{2}{*}{Method} & & \multicolumn{4}{c}{\dsid{ScanNet++$_4$}} & & \multicolumn{4}{c}{\dscd{ETH3D$_4$}} & & \multicolumn{4}{c}{\dscd{Tanks\&Temples$_4$}} & & \multicolumn{4}{c}{\dsod{4D-DRESS}} & & \multicolumn{4}{c}{\dsod{MV-dVRK}} \\
\cmidrule(lr){3-6} \cmidrule(lr){8-11} \cmidrule(lr){13-16} \cmidrule(lr){18-21} \cmidrule(lr){23-26}
& & \multicolumn{4}{c}{\textbf{AUC@10/NS/M/NM}} & & \multicolumn{4}{c}{\textbf{AUC@10/NS/M/NM}} & & \multicolumn{4}{c}{\textbf{AUC@10/NS/M/NM}} & & \multicolumn{4}{c}{\textbf{AUC@10/NS/M/NM}} & & \multicolumn{4}{c}{\textbf{AUC@10/NS/M/NM}} \\
\midrule
MASt3R$^\dagger$ feed-forward &  & 65.6 & \na & 54.9 & 68.9 &  & 40.6 & \na & 28.1 & 44.1 &  & 44.4 & \na & 19.5 & 46.2 &  & 26.5 & \na & 7.8 & 26.9 &  & 70.1 & \na & 28.7 & 70.3 \\
SfM scaffold &  & 25.2 & 0.0 & 0.2 & 33.1 &  & 42.0 & 0.0 & 0.3 & 56.7 &  & 52.0 & 0.0 & 0.4 & 56.2 &  & 87.6 & 0.0 & 4.8 & 88.7 &  & 83.1 & 0.0 & 11.3 & 83.4 \\
GGPT (MASt3R$^\dagger$) &  & \U{69.1} & 64.9 & 53.7 & \B{74.0} &  & 56.3 & 33.2 & 21.7 & 67.1 &  & 58.5 & 29.4 & 18.4 & 62.0 &  & 83.5 & 23.2 & 13.6 & 84.4 &  & \U{89.1} & 56.5 & 38.8 & \U{89.4} \\
Naive depth completion (MASt3R$^\dagger$) &  & 68.4 & 64.1 & \B{57.1} & 72.1 &  & 61.4 & 39.7 & 33.6 & 69.9 &  & 58.5 & 27.2 & 20.9 & 61.8 &  & \U{90.4} & 34.0 & 27.0 & \U{91.1} &  & 88.1 & 45.9 & 33.6 & 88.3 \\
MASt3R$^\dagger$ + SfM-anchored DGA &  & 68.0 & 63.8 & \U{56.4} & 71.7 &  & \U{61.9} & 39.8 & 32.5 & \U{70.5} &  & \U{59.8} & 31.0 & 24.0 & \U{62.8} &  & 86.7 & 45.8 & 34.2 & 87.3 &  & 85.7 & 49.0 & 34.5 & 86.0 \\
Scaffold3D (Unanchored DGA) &  & \U{69.1} & \U{65.3} & 54.5 & \U{73.7} &  & 61.2 & \U{47.8} & \U{39.2} & 68.0 &  & 55.2 & \U{34.9} & \U{26.2} & 57.9 &  & 61.5 & \U{55.3} & \U{39.3} & 61.8 &  & 88.4 & \B{81.0} & \B{44.3} & 88.6 \\
\rowcolor{cyan!16} Scaffold3D (ours) &  & \B{69.6} & \B{65.8} & 56.3 & \B{74.0} &  & \B{66.4} & \B{49.1} & \B{40.3} & \B{74.7} &  & \B{60.9} & \B{35.5} & \B{26.3} & \B{64.1} &  & \B{92.9} & \B{72.7} & \B{51.2} & \B{93.5} &  & \B{90.4} & \U{66.4} & \U{44.0} & \B{90.6} \\
\bottomrule
\vspace{-0.8cm}
\end{tabular}
}
\label{tab:pipeline_fusion_ablation}
\end{table*}

%% file: tables/16_scaffold_source_and_scaling.tex
\begin{table*}[t]
\centering
\caption{\textbf{Scaffold source and view count.} \emph{Left:} AUC@10 of Scaffold3D with scaffolds from eleven sources, using resized and center-cropped 512$\times$384 images (C) or full-resolution 4:3 center crops (F) for matching. Cov.: scaffold pixel coverage range; best FF: per-dataset best feed-forward baseline ranked in \cref{tab:main_h16_results}; cyan: default configurations. \emph{Right:} AUC@10, pair counts, and total runtime (s) on ETH3D for both pair graphs. Details in \cref{tab:scaffold_source_full,tab:scaling_full,tab:runtime_breakdown}.}
\nointerlineskip
\begin{minipage}[t]{0.585\linewidth}
\vspace{0pt}
\centering
\resizebox{\linewidth}{!}{%
\setlength{\tabcolsep}{2.4pt}
\begin{tabular}{llc c ccccc}
\toprule
Scaffold source & Res. & Cov.\ (\%) & \textbf{Mean} & SN++$_4$ & ETH3D$_4$ & T\&T$_4$ & 4D-DR. & MV-dVRK \\
\midrule
None (MASt3R$^\dagger$) & -- & -- & 49.4 & 65.6 & 40.6 & 44.4 & 26.5 & 70.1 \\
None (best FF) & -- & -- & 64.6 & 68.0 & 49.3 & 52.9 & 76.9 & 75.8 \\
\midrule
SP+LG~\cite{superpoint,lindenberger2023lightglue} & -- & 0.3--2.6 & 58.3 & 64.6 & 42.7 & 49.4 & 68.2 & 66.5 \\
UFM~\cite{zhang2025ufm} & C & 19.6--83.5 & 61.2 & 58.9 & 49.7 & 49.5 & 77.9 & 70.0 \\
UFM~\cite{zhang2025ufm} & F & 20.7--83.1 & 62.1 & 59.9 & 52.1 & 49.8 & 77.5 & 71.4 \\
LoMa~\cite{nordstrom2026loma} & C & 0.2--2.7 & 64.3 & 65.8 & 47.9 & 50.8 & 81.1 & 75.8 \\
RoMa~\cite{edstedt2024roma} & F & 4.9--70.2 & 65.4 & 58.1 & 55.7 & 55.9 & 83.7 & 73.4 \\
MASt3R~\cite{mast3r_eccv24} & F & 13.0--36.7 & 66.5 & 64.5 & 50.5 & 52.4 & 83.7 & 81.4 \\
RoMa~\cite{edstedt2024roma} & C & 6.8--72.1 & 66.5 & 60.8 & 54.7 & 56.9 & 86.2 & 74.0 \\
MASt3R~\cite{mast3r_eccv24} & C & 13.1--36.9 & 66.8 & 64.9 & 50.8 & 52.8 & 82.0 & 83.4 \\
LoMa~\cite{nordstrom2026loma} & F & 0.3--3.5 & 67.0 & 66.1 & 53.1 & 51.2 & 86.2 & 78.3 \\
\rowcolor{cyan!16} RoMa v2~\cite{edstedt2025romav2} & C & 30.8--92.6 & \U{76.0} & \B{69.3} & \U{66.1} & \B{60.5} & \U{93.2} & \B{90.7} \\
RoMa v2~\cite{edstedt2025romav2} & F & 32.4--92.4 & \B{76.8} & \U{69.2} & \B{71.0} & \U{60.0} & \B{93.5} & \U{90.4} \\
\bottomrule
\end{tabular}}
\end{minipage}\hfill
\begin{minipage}[t]{0.395\linewidth}
\vspace{0pt}
\centering
\resizebox{\linewidth}{!}{%
\setlength{\tabcolsep}{2.4pt}
\renewcommand{\arraystretch}{1.0305}\begin{tabular}{@{}llrrrrr@{}}
\toprule
\multicolumn{2}{l}{Views} & 4 & 8 & 16 & 32 & 50 \\
\midrule
\multicolumn{7}{l}{\emph{AUC@10 on ETH3D}} \\
\rowcolor{cyan!16}\multicolumn{2}{l}{Scaffold3D (sparse)} & \B{66.4} & \U{66.9} & \B{69.9} & \B{67.8} & \B{60.3} \\
\multicolumn{2}{l}{Scaffold3D (complete)} & \B{66.4} & \B{67.4} & \U{67.0} & \U{51.3} & 24.0 \\
\multicolumn{2}{l}{Pi3X~\cite{wang2025pi3}} & \U{47.3} & 42.3 & 38.9 & 32.7 & \U{32.9} \\
\multicolumn{2}{l}{Pi3~\cite{wang2025pi3}} & 44.1 & 35.7 & 34.9 & 26.5 & 25.3 \\
\multicolumn{2}{l}{VGGT~\cite{wang2025vggt}} & 41.4 & 37.1 & 31.2 & 26.0 & 26.1 \\
\multicolumn{2}{l}{MapAnything~\cite{keetha2025mapanything}} & 21.2 & 17.0 & 13.5 & 11.6 & 11.2 \\
\midrule
\multicolumn{7}{l}{\emph{Pairs and total runtime}} \\
\multirow{2}{*}{Sparse graph} & Pairs & 6 & 17 & 41 & 89 & 143 \\
 & Time (s) & 35.9 & 57.6 & 152.2 & 220.5 & 361.6 \\
\addlinespace[1pt]
\multirow{2}{*}{Complete graph} & Pairs & 6 & 28 & 120 & 496 & 1225 \\
 & Time (s) & 35.9 & 86.0 & 312.9 & 786.0 & 1876.5 \\
\bottomrule
\end{tabular}}
\end{minipage}
\label{tab:scaffold_quality}
\end{table*}

%% file: tables/7_model_development_ablations.tex
\begin{table*}[!htbp]
\setlength{\tabcolsep}{2.1pt}
\centering
\caption{\textbf{Model-development ablations.} We report development runs that isolate key choices in Scaffold3D. Each dataset block reports AUC at 1, 5, and 10 cm. The controlled blocks motivate adding point tokens to image tokens, aggregating point tokens so the model can use all available SfM points, and pretraining with GT-overlap points before real SfM. These ablations use earlier training configurations than the final model (e.g., smaller training and data scale, different source of SfM scaffold), but are comparable within each block. Cyan rows indicate the selected design choices.}
\begin{tabular}{l crrr crrr crrr}
\toprule
\multirow[b]{2}{*}{Variant} & & \multicolumn{3}{c}{\dsid{ScanNet++$_4$}} & & \multicolumn{3}{c}{\dscd{ETH3D$_4$}} & & \multicolumn{3}{c}{\dscd{T\&T$_4$}} \\
\cmidrule(lr){3-5} \cmidrule(lr){7-9} \cmidrule(lr){11-13}
& & \multicolumn{3}{c}{\textbf{AUC@1/5/10}} & & \multicolumn{3}{c}{\textbf{AUC@1/5/10}} & & \multicolumn{3}{c}{\textbf{AUC@1/5/10}} \\
\midrule
\multicolumn{13}{l}{\emph{Token fusion (first sweep)}} \\
\midrule
Append point tokens as extra tokens &  & \U{20.3} & \U{49.5} & 64.6 &  & 5.6 & 21.3 & 34.0 &  & 4.8 & 20.3 & 33.7 \\
Mask image tokens at point locations &  & 15.5 & 43.7 & 59.4 &  & 6.5 & 21.9 & 35.3 &  & 4.1 & 17.5 & 30.3 \\
Replace image tokens with point tokens &  & 17.6 & 49.3 & \U{65.5} &  & \U{7.4} & \U{26.5} & \U{40.6} &  & \U{5.9} & \U{24.1} & \U{38.9} \\
\rowcolor{cyan!16} Add point tokens to image tokens &  & \B{25.7} & \B{59.4} & \B{72.8} &  & \B{11.6} & \B{37.1} & \B{53.6} &  & \B{11.7} & \B{35.2} & \B{50.6} \\
\midrule
\multicolumn{13}{l}{\emph{Token fusion (second sweep)}} \\
\midrule
Replace image tokens with point tokens &  & \U{27.1} & \U{59.4} & 71.9 &  & 12.5 & 37.0 & 52.6 &  & 14.6 & 38.9 & 54.2 \\
Replace image tokens + point-token dropout &  & 25.0 & 58.8 & \U{72.3} &  & 12.6 & 36.8 & 52.7 &  & 13.4 & 38.3 & 53.7 \\
Gate image/point tokens &  & 20.8 & 56.4 & 71.4 &  & \U{18.1} & \U{47.1} & \U{62.3} &  & \U{16.4} & \U{43.2} & \U{58.6} \\
\rowcolor{cyan!16} Add point tokens to image tokens &  & \B{29.8} & \B{61.3} & \B{74.0} &  & \B{20.3} & \B{50.1} & \B{64.6} &  & \B{18.5} & \B{45.9} & \B{60.5} \\
\midrule
\multicolumn{13}{l}{\emph{Point-token capacity}} \\
\midrule
Unaggregated point tokens (768 points) &  & 5.2 & 31.3 & 52.3 &  & \U{4.2} & 30.3 & 52.4 &  & 9.8 & 46.9 & \U{66.8} \\
Patch-aggregated point tokens (all points) &  & \B{5.6} & \U{32.0} & \U{52.9} &  & 4.0 & \U{30.9} & \U{52.8} &  & \U{10.7} & \U{47.0} & 66.7 \\
\rowcolor{cyan!16} Patch aggregation + point self-attention &  & \U{5.3} & \B{32.6} & \B{53.6} &  & \B{4.9} & \B{32.8} & \B{54.1} &  & \B{11.5} & \B{48.8} & \B{67.9} \\
\midrule
\multicolumn{13}{l}{\emph{Sparse-input curriculum}} \\
\midrule
Real SfM from start &  & \U{17.4} & \U{48.1} & \U{64.2} &  & \U{8.3} & \U{28.2} & \U{41.8} &  & \U{6.1} & \U{23.2} & \U{36.6} \\
\rowcolor{cyan!16} GT-overlap pretrain $\rightarrow$ real SfM &  & \B{20.3} & \B{51.9} & \B{67.2} &  & \B{11.5} & \B{33.7} & \B{48.3} &  & \B{11.7} & \B{34.0} & \B{48.9} \\
\bottomrule
\end{tabular}
\label{tab:model_development_ablations}
\end{table*}

%% file: tables/15_pi3x_finetune_ablation.tex
\begin{table*}[!htbp]
\setlength{\tabcolsep}{1.8pt}
\centering
\caption{\textbf{Pi3X fine-tuning ablations.} We compare released Pi3X~\cite{wang2025pi3} with variants fine-tuned under our SfM-conditioned training setup. The first numeric block reports mean AUC@10 over the five datasets. Each dataset block reports all-pixel AUC@10 for the conditioning-free feed-forward output (FF), conditioning on the same dataset-specific SfM scaffold with predicted cameras (Pred.), and conditioning with the scaffold cameras fixed (Fixed). Thresholds are in centimeters except MV-dVRK, which uses millimeters. Unless stated otherwise, variants use 67K steps, learning rate $10^{-5}$, bfloat16, and a frozen image encoder.}
\resizebox{\linewidth}{!}{%
\begin{tabular}{l crrr crrr crrr crrr crrr crrr}
\toprule
\multirow[b]{2}{*}{Variant} & & \multicolumn{3}{c}{\textbf{Overall}} & & \multicolumn{3}{c}{\dsid{ScanNet++$_4$}} & & \multicolumn{3}{c}{\dscd{ETH3D$_4$}} & & \multicolumn{3}{c}{\dscd{T\&T$_4$}} & & \multicolumn{3}{c}{\dsod{4D-DRESS}} & & \multicolumn{3}{c}{\dsod{MV-dVRK}} \\
\cmidrule(lr){3-5} \cmidrule(lr){7-9} \cmidrule(lr){11-13} \cmidrule(lr){15-17} \cmidrule(lr){19-21} \cmidrule(lr){23-25}
& & \textbf{FF} & \textbf{Pred.} & \textbf{Fixed} & & \textbf{FF} & \textbf{Pred.} & \textbf{Fixed} & & \textbf{FF} & \textbf{Pred.} & \textbf{Fixed} & & \textbf{FF} & \textbf{Pred.} & \textbf{Fixed} & & \textbf{FF} & \textbf{Pred.} & \textbf{Fixed} & & \textbf{FF} & \textbf{Pred.} & \textbf{Fixed} \\
\midrule
\multicolumn{25}{l}{\emph{Training length}} \\
\midrule
Released Pi3X (no fine-tuning) &  & 44.1 & 67.4 & 72.1 &  & 66.5 & 67.1 & 68.5 &  & \B{47.3} & \B{54.5} & 57.8 &  & \B{44.8} & \B{51.2} & \B{52.7} &  & 26.5 & \U{90.0} & \U{91.9} &  & 35.3 & 74.1 & 89.5 \\
34K steps (1 day) &  & 47.4 & \U{68.1} & \B{72.4} &  & 69.2 & 69.7 & 71.6 &  & 38.8 & \U{53.0} & \B{58.6} &  & 40.2 & \U{46.2} & \U{49.6} &  & 35.9 & 88.8 & \U{91.9} &  & \B{53.0} & \U{83.0} & 90.2 \\
67K steps (2 days) &  & \U{51.6} & \U{68.1} & \U{72.3} &  & \U{70.5} & \U{70.5} & \U{72.0} &  & \U{39.4} & 52.5 & \U{57.9} &  & 39.8 & 45.2 & 49.3 &  & \U{55.5} & 89.6 & 91.8 &  & \U{52.8} & 82.5 & \B{90.6} \\
123K steps (4 days) &  & \B{52.8} & \B{68.7} & \U{72.3} &  & \B{71.4} & \B{72.0} & \B{72.7} &  & 39.2 & 52.3 & 57.4 &  & \U{41.9} & 45.4 & 48.6 &  & \B{59.3} & \B{90.3} & \B{92.3} &  & 52.3 & \B{83.5} & \U{90.5} \\
\midrule
\multicolumn{25}{l}{\emph{Training choices (67K steps)}} \\
\midrule
Base recipe &  & 51.6 & 68.1 & \U{72.3} &  & \U{70.5} & 70.5 & 72.0 &  & 39.4 & 52.5 & \B{57.9} &  & 39.8 & 45.2 & \U{49.3} &  & 55.5 & \U{89.6} & 91.8 &  & 52.8 & 82.5 & \B{90.6} \\
Learning rate $10^{-4}$ &  & 45.5 & 63.5 & 70.3 &  & 61.8 & 63.5 & 67.9 &  & 32.1 & 46.1 & 55.4 &  & 33.4 & 42.3 & 46.7 &  & 47.0 & 82.1 & 91.6 &  & 52.9 & 83.7 & 89.7 \\
Scaffold-quality gating &  & \U{53.9} & 68.6 & \B{72.4} &  & 70.3 & 69.3 & 71.6 &  & \B{40.7} & \B{53.9} & \U{57.7} &  & \U{40.7} & \B{46.2} & \B{50.3} &  & 59.6 & 89.1 & 92.0 &  & \U{58.1} & 84.3 & 90.3 \\
GT-scaffold probability 0.1 &  & 49.6 & 67.9 & 71.9 &  & \B{70.9} & \B{71.2} & \U{72.1} &  & \B{40.7} & 50.9 & 56.8 &  & 40.4 & 45.3 & 48.7 &  & 44.7 & 89.3 & 91.7 &  & 51.2 & 82.9 & 90.2 \\
No confidence loss &  & 52.8 & 67.8 & 72.2 &  & \B{70.9} & 70.4 & \U{72.1} &  & \U{40.1} & 52.7 & \U{57.7} &  & \B{40.8} & \U{45.4} & 49.1 &  & 56.1 & 89.5 & \U{92.1} &  & 56.1 & 81.2 & 90.3 \\
Unfrozen image encoder &  & \B{56.1} & \U{68.7} & 72.2 &  & 67.3 & \U{70.9} & 71.7 &  & 38.1 & 52.9 & \B{57.9} &  & 40.1 & 45.2 & 48.9 &  & \B{69.2} & 88.9 & 92.0 &  & \B{66.0} & \U{85.6} & \U{90.4} \\
Conditioning probability $2.5\times$ &  & 53.7 & \B{69.3} & 72.2 &  & 68.9 & 70.7 & \B{72.4} &  & 39.5 & \U{53.7} & 57.4 &  & 40.5 & 44.9 & 48.7 &  & \U{61.8} & \B{90.5} & \B{92.3} &  & 57.7 & \B{86.8} & 90.3 \\
\bottomrule
\end{tabular}
}
\label{tab:pi3x_finetune_ablation}
\end{table*}

%% file: tables/13_pow3r_controls.tex
\begin{table*}[!htbp]
\setlength{\tabcolsep}{1.8pt}
\centering
\caption{\textbf{Comparison of scaffold-conditioning interfaces.} We compare scaffold-conditioning interfaces under the same SfM inputs and evaluation protocol. The first numeric column reports mean AUC@10 over the five dataset blocks. Each dataset block reports AUC at 1, 5, and 10 in the dataset unit: centimeters except MV-dVRK, which uses millimeters. Conditioning input icons denote \faIcon{layer-group}: scaffold as patchified depth map, \faIcon{project-diagram}: scaffold as 3D point tokens or patchified pointmaps, \faIcon{location-arrow}: scaffold cameras or relative pose, and \faIcon{camera}: scaffold intrinsics/rays. Cyan marks our final Scaffold3D checkpoint trained with the staged curriculum. The pretrained Pow3R using our SfM-anchored alignment improves over plain DGA by 3.9 AUC@10 on ETH3D and 3.3 on Tanks\&Temples. Variants trained from the same MASt3R checkpoint suggest three takeaways: (1) in the simplified depth/pointmap-only interface, decoder fusion improves over encoder fusion by 0.5 mean AUC@10 for \faIcon{layer-group} depth inputs and 1.7 for \faIcon{project-diagram} pointmap inputs, (2) \faIcon{project-diagram} pointmap decoder fusion improves over \faIcon{layer-group} depth decoder fusion by 0.9 mean AUC@10, and (3) different conditioning mechanisms, including Pow3R-style conditioning and pointmap decoder fusion, achieve comparable performance. These results support SfM-conditioned training and alignment without establishing a uniquely superior conditioning interface. These conclusions are tied to the dataset mixture, training length, hyperparameters, and SfM construction used here.}
\resizebox{\linewidth}{!}{%
\begin{tabular}{llc crrr crrr crrr crrr crrr}
\toprule
\multirow[b]{2}{*}{Variant} & \multirow[b]{2}{*}{Input} & \textbf{Overall} & & \multicolumn{3}{c}{\dsid{ScanNet++$_4$}} & & \multicolumn{3}{c}{\dscd{ETH3D$_4$}} & & \multicolumn{3}{c}{\dscd{T\&T$_4$}} & & \multicolumn{3}{c}{\dsod{4D-DRESS}} & & \multicolumn{3}{c}{\dsod{MV-dVRK}} \\
\cmidrule(lr){3-3} \cmidrule(lr){5-7} \cmidrule(lr){9-11} \cmidrule(lr){13-15} \cmidrule(lr){17-19} \cmidrule(lr){21-23}
& & \textbf{AUC@10} & & \multicolumn{3}{c}{\textbf{AUC@1/5/10}} & & \multicolumn{3}{c}{\textbf{AUC@1/5/10}} & & \multicolumn{3}{c}{\textbf{AUC@1/5/10}} & & \multicolumn{3}{c}{\textbf{AUC@1/5/10}} & & \multicolumn{3}{c}{\textbf{AUC@1/5/10}} \\
\midrule
\multicolumn{23}{l}{\emph{Pow3R-pretrained references}} \\
\midrule
Pow3R~\cite{pow3r_cvpr25} (DGA) & \faIcon{layer-group}\,\faIcon{location-arrow}\,\faIcon{camera} & 71.2 &  & 20.0 & 49.1 & 62.2 &  & 18.7 & 45.3 & 58.4 &  & 18.9 & 43.0 & 56.0 &  & 59.2 & 84.6 & 90.9 &  & 67.6 & 83.4 & 88.3 \\
Pow3R~\cite{pow3r_cvpr25} & \faIcon{layer-group}\,\faIcon{location-arrow}\,\faIcon{camera} & 73.0 &  & 23.4 & 49.4 & 61.7 &  & \U{25.7} & 50.9 & 62.3 &  & 26.1 & 48.1 & 59.3 &  & \B{76.9} & \U{89.2} & 93.2 &  & 65.2 & 83.5 & 88.6 \\
Pow3R-pretrained + SfM fine-tune & \faIcon{layer-group}\,\faIcon{location-arrow}\,\faIcon{camera} & 75.6 &  & 25.9 & 56.2 & 69.1 &  & 25.6 & 52.0 & 64.5 &  & \B{27.0} & 48.9 & 59.9 &  & 75.3 & \U{89.2} & \B{93.4} &  & \U{69.9} & \B{86.1} & \B{91.0} \\
\midrule
\multicolumn{23}{l}{\emph{Variants trained from the MASt3R checkpoint}} \\
\midrule
Pointmap encoder fusion & \faIcon{project-diagram} & 74.1 &  & 24.7 & 53.9 & 67.5 &  & 23.0 & 49.7 & 62.3 &  & 25.3 & 48.1 & 59.4 &  & 69.8 & 86.8 & 91.8 &  & 65.4 & 83.9 & 89.5 \\
Depth encoder fusion & \faIcon{layer-group} & 74.4 &  & 24.5 & 54.2 & 67.3 &  & 25.3 & 51.1 & 63.3 &  & 26.4 & 48.8 & 60.3 &  & 71.0 & 87.3 & 92.1 &  & 62.7 & 83.1 & 89.1 \\
Depth decoder fusion & \faIcon{layer-group} & 74.9 &  & \U{26.5} & \U{57.0} & \U{69.9} &  & \B{25.9} & 52.8 & 65.2 &  & \U{26.9} & \B{49.3} & \B{60.9} &  & 61.3 & 84.0 & 90.3 &  & 60.1 & 82.1 & 88.4 \\
Pow3R with pointmap decoder fusion & \faIcon{project-diagram}\,\faIcon{location-arrow}\,\faIcon{camera} & 75.4 &  & 26.3 & 55.6 & 68.9 &  & 25.3 & 52.7 & 65.5 &  & 24.1 & 47.4 & 59.2 &  & 76.0 & 88.9 & 93.1 &  & 69.8 & 85.3 & 90.3 \\
Pow3R (native: depthmap encoder fusion) & \faIcon{layer-group}\,\faIcon{location-arrow}\,\faIcon{camera} & 75.7 &  & \B{27.3} & 56.7 & 69.4 &  & 24.4 & 52.7 & 66.0 &  & 25.2 & 48.0 & 59.5 &  & \U{76.3} & \B{89.4} & \B{93.4} &  & 68.8 & 85.2 & 90.3 \\
Pow3R with depthmap decoder fusion & \faIcon{layer-group}\,\faIcon{location-arrow}\,\faIcon{camera} & \U{75.8} &  & 25.8 & 55.1 & 68.9 &  & 25.1 & \B{53.7} & \B{66.6} &  & 26.3 & 48.3 & 59.7 &  & 75.6 & \U{89.2} & \U{93.3} &  & \B{71.6} & \U{85.9} & \U{90.6} \\
Pointmap decoder fusion & \faIcon{project-diagram} & \U{75.8} &  & 25.9 & \B{57.3} & \B{70.1} &  & 24.7 & 53.2 & 65.9 &  & 26.6 & 49.0 & \U{60.6} &  & 72.2 & 87.9 & 92.5 &  & 64.9 & 84.2 & 89.9 \\
\midrule
\multicolumn{23}{l}{\emph{Final checkpoint trained with the staged Scaffold3D curriculum}} \\
\midrule
\rowcolor{cyan!16} Scaffold3D: pointmap decoder fusion & \faIcon{project-diagram} & \B{76.0} &  & 25.7 & 56.5 & 69.6 &  & 25.6 & \U{53.6} & \U{66.4} &  & 26.4 & \U{49.2} & \B{60.9} &  & 74.4 & 88.6 & 92.9 &  & 68.4 & 85.4 & 90.4 \\
\bottomrule
\end{tabular}
}
\label{tab:pow3r_conditioning}
\end{table*}

%% file: tables/9_point_budget_sensitivity.tex
\begin{table*}[!htbp]
\setlength{\tabcolsep}{2.1pt}
\centering
\caption{\textbf{Conditioning point-budget sensitivity.} The subsampled-budget rows vary the number of scaffold points injected into Scaffold3D, while final SfM-anchored alignment retains the full filtered scaffold. Each dataset block reports AUC@1, AUC@5, AUC@10, and AUC@M. M denotes AUC@10 on monocular pixels. Performance saturates after a moderate point budget in this subsampling experiment, suggesting diminishing returns from injecting every available scaffold point into the predictor. The separate BA-tracks row (4316 tracks on average) uses only bundle-adjusted tracks and omits the DLT-triangulated scaffold observations.}
\resizebox{\linewidth}{!}{%
\begin{tabular}{r crrrr crrrr crrrr crrrr crrrr}
\toprule
\multirow[b]{2}{*}{\#Points} & & \multicolumn{4}{c}{\dsid{ScanNet++$_4$}} & & \multicolumn{4}{c}{\dscd{ETH3D$_4$}} & & \multicolumn{4}{c}{\dscd{T\&T$_4$}} & & \multicolumn{4}{c}{\dsod{4D-DRESS}} & & \multicolumn{4}{c}{\dsod{MV-dVRK}} \\
\cmidrule(lr){3-6} \cmidrule(lr){8-11} \cmidrule(lr){13-16} \cmidrule(lr){18-21} \cmidrule(lr){23-26}
& & \multicolumn{4}{c}{\textbf{AUC@1/5/10/M}} & & \multicolumn{4}{c}{\textbf{AUC@1/5/10/M}} & & \multicolumn{4}{c}{\textbf{AUC@1/5/10/M}} & & \multicolumn{4}{c}{\textbf{AUC@1/5/10/M}} & & \multicolumn{4}{c}{\textbf{AUC@1/5/10/M}} \\
\midrule
\multicolumn{26}{l}{\emph{Subsampled SfM scaffold observations}} \\
\midrule
0 &  & 24.7 & 54.8 & 68.0 & 56.1 &  & 24.2 & 49.4 & 62.1 & 32.1 &  & 26.1 & 48.6 & 60.0 & 23.8 &  & 56.6 & 79.4 & 86.6 & 34.3 &  & 57.1 & 79.0 & 85.7 & 34.5 \\
512 &  & 25.7 & 56.6 & 69.6 & \B{58.7} &  & 23.2 & 51.2 & 64.8 & 40.3 &  & 25.6 & 48.6 & 60.2 & 22.7 &  & 74.0 & 88.4 & 92.8 & 51.6 &  & 64.6 & 83.6 & 89.2 & 40.9 \\
1024 &  & \U{26.1} & 56.6 & 69.5 & \U{58.2} &  & 23.0 & 51.3 & 65.0 & \U{40.7} &  & 26.4 & 48.8 & 60.1 & 23.5 &  & 74.2 & 88.5 & \U{92.9} & 52.8 &  & 65.3 & 83.7 & 89.3 & 41.7 \\
2048 &  & \B{26.3} & 56.7 & 69.6 & 57.7 &  & 23.3 & 51.6 & 65.1 & 39.8 &  & 24.1 & 48.2 & 59.9 & 24.7 &  & \B{75.1} & \B{88.7} & \B{93.0} & \U{53.7} &  & 67.1 & 84.4 & 89.7 & 42.2 \\
4096 &  & 25.8 & \U{56.8} & 69.7 & 57.8 &  & 22.7 & 52.0 & 65.7 & 40.2 &  & 25.5 & 48.7 & 60.2 & 24.8 &  & 72.4 & 88.1 & 92.7 & 53.5 &  & \U{69.0} & 84.9 & 90.0 & 42.8 \\
8192 &  & \U{26.1} & \B{56.9} & 69.8 & 57.9 &  & \U{24.5} & 52.3 & 65.7 & 40.6 &  & 25.9 & \U{49.3} & \U{60.9} & 25.2 &  & 73.6 & 88.4 & 92.8 & \B{53.9} &  & 68.0 & 84.8 & 90.0 & 43.1 \\
16384 &  & 25.5 & 56.6 & 69.7 & 57.1 &  & 24.2 & 52.3 & 65.9 & \B{40.8} &  & 25.2 & 48.9 & 60.4 & 25.2 &  & 73.8 & 88.5 & 92.8 & 52.6 &  & 63.9 & 84.1 & 89.7 & 42.8 \\
32768 &  & 26.0 & \B{56.9} & \B{70.0} & 57.1 &  & 24.1 & 52.0 & 65.6 & 39.6 &  & 25.9 & 49.2 & 60.6 & 25.7 &  & \U{74.8} & \B{88.7} & \B{93.0} & 50.8 &  & \B{69.4} & \U{85.3} & \U{90.2} & 43.6 \\
65536 &  & 25.3 & 56.2 & 69.3 & 56.4 &  & 23.8 & 52.6 & 65.9 & 40.5 &  & \B{26.9} & \B{49.9} & \B{61.3} & \B{26.6} &  & 73.8 & 88.5 & \U{92.9} & 51.1 &  & 68.3 & 85.2 & \U{90.2} & \B{44.0} \\
131072 &  & \U{26.1} & 56.5 & 69.6 & 56.1 &  & \B{24.6} & \B{53.3} & \B{66.3} & 40.4 &  & \U{26.6} & \U{49.3} & 60.7 & 25.9 &  & 74.2 & \U{88.6} & \U{92.9} & 51.1 &  & 65.8 & 84.7 & 90.0 & \B{44.0} \\
\rowcolor{cyan!16} all &  & 25.9 & \U{56.8} & \U{69.9} & 56.2 &  & 24.0 & \U{53.1} & \U{66.1} & 40.0 &  & 25.8 & 49.2 & \U{60.9} & \U{26.1} &  & 73.3 & 88.4 & 92.8 & 51.2 &  & \B{69.4} & \B{85.6} & \B{90.5} & \U{43.9} \\
\midrule
\multicolumn{26}{l}{\emph{BA tracks only, no DLT triangulation}} \\
\midrule
4316 &  & 26.0 & 56.3 & 69.3 & 57.0 &  & 24.0 & 52.6 & 65.5 & \B{40.8} &  & \U{26.6} & 48.9 & 60.5 & 24.6 &  & 66.7 & 85.1 & 90.2 & 37.1 &  & 64.8 & 83.0 & 88.5 & 35.8 \\
\bottomrule
\end{tabular}
}
\label{tab:point_budget_sensitivity}
\end{table*}

%% file: tables/21_token_coverage.tex
\begin{table}[!htbp]
\setlength{\tabcolsep}{4.0pt}
\centering
\caption{\textbf{Point-token coverage.} Percentage of the 768 image tokens per view (16$\times$16 patches on the 512$\times$384 canvas) that receive a point token, i.e., contain at least one scaffold observation. Statistics are over scenes; each scene value is the mean over its views.}
\begin{tabular}{lccccccc}
\toprule
Dataset & Scenes & Min & 25\% & 50\% & 75\% & Max & Mean \\
\midrule
ScanNet++$_4$ & 29 & 8.9 & 54.0 & 60.6 & 64.5 & 78.6 & 58.1 \\
ETH3D$_4$ & 26 & 37.5 & 56.8 & 69.9 & 76.9 & 88.9 & 67.1 \\
Tanks\&Temples$_4$ & 32 & 35.5 & 70.0 & 74.4 & 79.9 & 87.8 & 72.1 \\
4D-DRESS & 24 & 20.6 & 22.6 & 25.1 & 27.3 & 29.8 & 25.0 \\
MV-dVRK & 23 & 63.2 & 81.2 & 84.5 & 89.8 & 93.1 & 84.2 \\
\bottomrule
\end{tabular}
\label{tab:token_coverage}
\end{table}

%% file: tables/11_fullres_romav2.tex
\begin{table*}[!htbp]
\setlength{\tabcolsep}{3.0pt}
\centering
\caption{\textbf{Full-resolution RoMaV2 scaffold matching.} We rerun scaffold-conditioned methods using full-resolution 4:3 center crops as RoMaV2 matcher inputs. Matches are expressed on the 512$\times$384 grid used for BA/DLT, conditioning, and evaluation. After DLT, visible track observations are rasterized to this grid by rounding their 2D coordinates and averaging collisions. Each cell reports AUC@10 with the change relative to the corresponding 512$\times$384 RoMaV2 scaffold run. Depth-completion rows include SfM overwrite. Full-resolution matching strengthens the scaffold most clearly on ETH3D, but the relative trends remain similar, suggesting that the main conclusions are not an artifact of low-resolution matching.}
\resizebox{\linewidth}{!}{%
\begin{tabular}{lccccc}
\toprule
Method & \dsid{ScanNet++$_4$} & \dscd{ETH3D$_4$} & \dscd{Tanks\&Temples$_4$} & \dsod{4D-DRESS} & \dsod{MV-dVRK} \\
\midrule
\addlinespace[0.15em]
\multicolumn{6}{l}{\emph{Monocular depth completion}} \\
\midrule
OMNI-DC~\cite{zuo2025omni} & 55.6 (-0.5) & 63.6 (+6.4) & \U{59.1} (-0.4) & \U{93.6} (+1.0) & 87.3 (-0.7) \\
Marigold-DC~\cite{viola2024marigolddc} & 56.8 (+0.5) & 66.0 (+6.4) & 58.1 (+0.0) & \B{93.8} (+0.7) & \U{88.8} (-0.9) \\
\midrule
\addlinespace[0.15em]
\multicolumn{6}{l}{\emph{Point-token propagation}} \\
\midrule
GGPT (VGGT)~\cite{chen2026ggpt} & 63.5 (+0.3) & 58.7 (+2.1) & 53.8 (-1.2) & 92.7 (+0.8) & 83.6 (+0.6) \\
GGPT (Pi3)~\cite{chen2026ggpt} & 67.1 (+1.0) & 50.5 (+0.5) & 51.2 (-0.7) & 93.1 (+0.4) & 88.3 (+0.4) \\
GGPT (Pi3X)~\cite{chen2026ggpt} & 65.7 (+0.3) & 63.0 (+3.0) & 54.8 (-1.3) & 89.3 (+1.7) & 77.9 (-3.1) \\
GGPT (MASt3R$^\dagger$)~\cite{chen2026ggpt} & \U{69.1} (+0.0) & 60.9 (+4.7) & 57.9 (-0.5) & 82.7 (-0.7) & 88.7 (-0.4) \\
\midrule
\addlinespace[0.15em]
\multicolumn{6}{l}{\emph{Geometry-conditioned predictors}} \\
\midrule
MVSAnywhere~\cite{izquierdo2025mvsanywhere}\textsuperscript{\S} & 40.9 (+0.8) & 41.6 (-0.2) & 42.6 (-0.2) & 77.2 (-0.3) & 79.1 (-0.6) \\
MapAnything~\cite{keetha2025mapanything} & 51.2 (+0.8) & 42.7 (+3.2) & 44.4 (-0.5) & 82.5 (-0.4) & 78.9 (-0.2) \\
Pow3R~\cite{pow3r_cvpr25} & 61.9 (+0.2) & \U{68.0} (+5.7) & 58.5 (-0.8) & \B{93.8} (+0.6) & 87.9 (-0.7) \\
\rowcolor{cyan!16} Scaffold3D (ours) & \B{69.5} (-0.1) & \B{71.0} (+4.5) & \B{60.0} (-0.9) & 93.5 (+0.6) & \B{90.5} (+0.1) \\
\bottomrule
\end{tabular}
}
\label{tab:fullres_romav2}
\end{table*}

%% file: tables/12_gt_scaffold_points.tex
\begin{table*}[!htbp]
\setlength{\tabcolsep}{3.0pt}
\centering
\caption{\textbf{Ground-truth scaffold geometry.} We replace scaffold points, camera poses, and intrinsics with ground truth, retaining the original scaffold-support pixels where ground truth is valid. Each cell reports AUC@10 and its change from the corresponding estimated-scaffold evaluation. All methods use the same dataset-specific scaffold support, including GGPT. OMNI-DC is reported with and without scaffold-depth overwrite.}
\resizebox{\linewidth}{!}{%
\begin{tabular}{lccccc}
\toprule
Method & \dsid{ScanNet++$_4$} & \dscd{ETH3D$_4$} & \dscd{Tanks\&Temples$_4$} & \dsod{4D-DRESS} & \dsod{MV-dVRK} \\
\midrule
\addlinespace[0.15em]
\multicolumn{6}{l}{\emph{Monocular depth completion}} \\
\midrule
OMNI-DC~\cite{zuo2025omni} & 64.3 (+10.1) & 72.0 (+23.4) & 77.5 (+28.1) & 94.4 (+6.7) & 91.1 (+7.5) \\
\quad + SfM overwrite & 67.5 (+11.4) & 74.7 (+17.6) & \B{82.5} (+23.0) & \B{98.7} (+6.1) & \U{97.0} (+9.0) \\
\midrule
\addlinespace[0.15em]
\multicolumn{6}{l}{\emph{Point-token propagation}} \\
\midrule
GGPT (VGGT)~\cite{chen2026ggpt} & 68.8 (+5.2) & 56.3 (+1.9) & 62.7 (+8.8) & 95.4 (+4.8) & 89.4 (+5.0) \\
GGPT (Pi3)~\cite{chen2026ggpt} & 71.4 (+3.9) & 46.5 (+0.1) & 57.7 (+8.4) & 95.6 (+3.1) & 93.2 (+4.8) \\
GGPT (Pi3X)~\cite{chen2026ggpt} & 74.6 (+6.2) & 58.1 (+3.3) & 64.4 (+8.9) & 90.3 (+3.1) & 86.1 (+2.8) \\
GGPT (MASt3R$^\dagger$)~\cite{chen2026ggpt} & 76.6 (+7.5) & 66.0 (+9.8) & 73.6 (+15.1) & 86.3 (+2.8) & 94.2 (+5.0) \\
\midrule
\addlinespace[0.15em]
\multicolumn{6}{l}{\emph{Geometry-conditioned predictors}} \\
\midrule
MVSAnywhere~\cite{izquierdo2025mvsanywhere}\textsuperscript{\S} & 42.1 (+2.1) & 42.5 (+0.7) & 42.2 (-0.7) & 78.1 (+0.6) & 80.7 (+1.0) \\
MapAnything~\cite{keetha2025mapanything} & 56.2 (+5.7) & 26.6 (-12.9) & 42.1 (-2.7) & 85.2 (+2.3) & 82.0 (+2.8) \\
Pi3X~\cite{wang2025pi3} & \U{78.4} (+9.9) & 64.1 (+6.2) & 58.9 (+6.2) & 96.4 (+4.4) & 95.2 (+5.7) \\
Pow3R~\cite{pow3r_cvpr25} & 74.6 (+12.9) & \U{76.3} (+14.0) & \U{81.2} (+21.8) & \U{98.6} (+5.4) & 96.7 (+8.1) \\
\rowcolor{cyan!16} Scaffold3D (ours) & \B{79.6} (+10.0) & \B{77.4} (+11.0) & 80.7 (+19.9) & 98.0 (+5.1) & \B{97.3} (+7.0) \\
\bottomrule
\end{tabular}
}
\label{tab:gt_scaffold_points}
\end{table*}

%% file: tables/18_posthoc_alignment.tex
\begin{table*}[!htbp]
\setlength{\tabcolsep}{3.0pt}
\centering
\caption{\textbf{Post-hoc alignment of feed-forward predictions to the scaffold.} Each backbone is fused with the same dataset-specific SfM scaffold as Scaffold3D by training-free variants: (A) rigidly align the scaffold to the prediction with Umeyama and paste the scaffold over the prediction; (B) complete the scaffold depth maps with predicted depth aligned by scale and shift in inverse depth, after rigidly aligning the prediction to the scaffold; (C) as B, but with a single per-view median scale in depth, without a shift or rigid pre-alignment; (GGPT) point-token propagation~\cite{chen2026ggpt} from the backbone's prediction. ``Conditioned'' feeds the scaffold through the model's own conditioning interface (Pi3X and MapAnything rows of \cref{tab:main_h16_results}). Each dataset block reports AUC@10 on all pixels and outside SfM support (AUC@NS). For unconditioned rows, NS is evaluated with the same scaffold mask. Here GGPT uses the shared scaffold, whereas in \cref{tab:main_h16_results} each GGPT variant constructs its scaffold with the named initializer.}
\resizebox{\linewidth}{!}{%
\newcommand{\metricpair}[2]{\makebox[23pt][r]{#1}\hspace{5pt}\makebox[23pt][r]{#2}}
\begin{tabular}{@{}llccccc@{}}
\toprule
\multirow[b]{2}{*}{Backbone} & \multirow[b]{2}{*}{Variant} & \multicolumn{1}{c}{\dsid{ScanNet++$_4$}} & \multicolumn{1}{c}{\dscd{ETH3D$_4$}} & \multicolumn{1}{c}{\dscd{T\&T$_4$}} & \multicolumn{1}{c}{\dsod{4D-DRESS}} & \multicolumn{1}{c}{\dsod{MV-dVRK}} \\
\cmidrule(lr){3-3} \cmidrule(lr){4-4} \cmidrule(lr){5-5} \cmidrule(lr){6-6} \cmidrule(lr){7-7}
 & & \multicolumn{1}{c}{AUC@10/NS} & \multicolumn{1}{c}{AUC@10/NS} & \multicolumn{1}{c}{AUC@10/NS} & \multicolumn{1}{c}{AUC@10/NS} & \multicolumn{1}{c}{AUC@10/NS} \\
\midrule
SfM scaffold & -- & \metricpair{25.2}{0.0} & \metricpair{42.0}{0.0} & \metricpair{52.0}{0.0} & \metricpair{87.6}{0.2} & \metricpair{83.1}{0.0} \\
\midrule
Pi3~\cite{wang2025pi3} & -- & \metricpair{68.0}{\B{65.8}} & \metricpair{44.1}{37.1} & \metricpair{46.6}{31.0} & \metricpair{71.4}{57.2} & \metricpair{67.0}{47.4} \\
 & A & \metricpair{68.3}{65.1} & \metricpair{57.8}{33.1} & \metricpair{58.4}{25.6} & \metricpair{92.0}{58.4} & \metricpair{88.0}{47.2} \\
 & B & \metricpair{65.9}{60.8} & \metricpair{63.1}{41.8} & \metricpair{59.6}{31.1} & \metricpair{\B{93.5}}{\B{77.7}} & \metricpair{88.1}{48.9} \\
 & C & \metricpair{\U{69.4}}{\U{65.7}} & \metricpair{\U{65.7}}{\U{48.9}} & \metricpair{\U{60.0}}{31.7} & \metricpair{92.6}{66.6} & \metricpair{88.6}{53.0} \\
 & GGPT & \metricpair{67.6}{65.2} & \metricpair{46.4}{35.4} & \metricpair{49.2}{30.7} & \metricpair{92.4}{71.6} & \metricpair{88.4}{58.1} \\
\midrule
Pi3X~\cite{wang2025pi3} & -- & \metricpair{66.5}{64.5} & \metricpair{47.3}{39.8} & \metricpair{44.8}{29.1} & \metricpair{26.5}{17.2} & \metricpair{35.3}{17.8} \\
 & A & \metricpair{67.3}{63.5} & \metricpair{59.4}{36.7} & \metricpair{58.9}{28.7} & \metricpair{88.7}{12.0} & \metricpair{84.5}{15.7} \\
 & B & \metricpair{66.4}{61.7} & \metricpair{62.1}{40.2} & \metricpair{59.3}{30.3} & \metricpair{92.5}{67.5} & \metricpair{86.5}{34.3} \\
 & C & \metricpair{\U{69.4}}{\U{65.7}} & \metricpair{64.7}{46.7} & \metricpair{\U{60.0}}{31.6} & \metricpair{91.1}{49.1} & \metricpair{86.7}{38.2} \\
 & GGPT & \metricpair{68.3}{65.0} & \metricpair{54.8}{35.7} & \metricpair{55.5}{32.9} & \metricpair{87.2}{42.7} & \metricpair{83.3}{41.5} \\
 & conditioned & \metricpair{68.5}{65.0} & \metricpair{57.8}{47.5} & \metricpair{52.7}{\U{33.8}} & \metricpair{91.9}{\U{76.6}} & \metricpair{\U{89.5}}{\U{65.1}} \\
\midrule
VGGT~\cite{wang2025vggt} & -- & \metricpair{55.6}{52.7} & \metricpair{41.4}{33.9} & \metricpair{43.3}{28.4} & \metricpair{73.3}{52.4} & \metricpair{49.1}{26.8} \\
 & A & \metricpair{59.0}{51.9} & \metricpair{57.0}{29.9} & \metricpair{56.7}{20.3} & \metricpair{91.9}{53.9} & \metricpair{85.7}{26.4} \\
 & B & \metricpair{62.2}{55.4} & \metricpair{61.7}{39.9} & \metricpair{59.8}{32.3} & \metricpair{\U{93.2}}{71.8} & \metricpair{86.4}{33.9} \\
 & C & \metricpair{59.2}{51.2} & \metricpair{61.6}{39.8} & \metricpair{59.1}{30.7} & \metricpair{90.9}{41.1} & \metricpair{86.2}{33.5} \\
 & GGPT & \metricpair{63.6}{58.5} & \metricpair{54.4}{36.2} & \metricpair{53.8}{30.7} & \metricpair{90.6}{67.3} & \metricpair{84.4}{43.3} \\
\midrule
\textcolor{black!55}{VGGT-Omega~\cite{wang2025vggt}} & \textcolor{black!55}{--} & \metricpair{\textcolor{black!55}{39.3}}{\textcolor{black!55}{38.1}} & \metricpair{\textcolor{black!55}{52.0}}{\textcolor{black!55}{47.2}} & \metricpair{\textcolor{black!55}{45.7}}{\textcolor{black!55}{30.0}} & \metricpair{\textcolor{black!55}{88.2}}{\textcolor{black!55}{79.5}} & \metricpair{\textcolor{black!55}{67.8}}{\textcolor{black!55}{53.4}} \\
\textcolor{black!55}{} & \textcolor{black!55}{A} & \metricpair{\textcolor{black!55}{46.8}}{\textcolor{black!55}{35.2}} & \metricpair{\textcolor{black!55}{62.0}}{\textcolor{black!55}{41.0}} & \metricpair{\textcolor{black!55}{57.6}}{\textcolor{black!55}{22.9}} & \metricpair{\textcolor{black!55}{93.5}}{\textcolor{black!55}{79.1}} & \metricpair{\textcolor{black!55}{88.5}}{\textcolor{black!55}{52.0}} \\
\textcolor{black!55}{} & \textcolor{black!55}{B} & \metricpair{\textcolor{black!55}{57.5}}{\textcolor{black!55}{48.8}} & \metricpair{\textcolor{black!55}{66.2}}{\textcolor{black!55}{48.0}} & \metricpair{\textcolor{black!55}{60.3}}{\textcolor{black!55}{32.8}} & \metricpair{\textcolor{black!55}{93.6}}{\textcolor{black!55}{81.1}} & \metricpair{\textcolor{black!55}{88.3}}{\textcolor{black!55}{50.8}} \\
\textcolor{black!55}{} & \textcolor{black!55}{C} & \metricpair{\textcolor{black!55}{62.9}}{\textcolor{black!55}{56.3}} & \metricpair{\textcolor{black!55}{70.6}}{\textcolor{black!55}{57.4}} & \metricpair{\textcolor{black!55}{60.4}}{\textcolor{black!55}{33.7}} & \metricpair{\textcolor{black!55}{93.7}}{\textcolor{black!55}{80.1}} & \metricpair{\textcolor{black!55}{89.0}}{\textcolor{black!55}{57.3}} \\
\textcolor{black!55}{} & \textcolor{black!55}{GGPT} & \metricpair{\textcolor{black!55}{57.3}}{\textcolor{black!55}{49.3}} & \metricpair{\textcolor{black!55}{62.0}}{\textcolor{black!55}{45.6}} & \metricpair{\textcolor{black!55}{59.2}}{\textcolor{black!55}{31.8}} & \metricpair{\textcolor{black!55}{93.7}}{\textcolor{black!55}{82.1}} & \metricpair{\textcolor{black!55}{89.6}}{\textcolor{black!55}{64.1}} \\
\midrule
MapAnything~\cite{keetha2025mapanything} & -- & \metricpair{61.0}{58.4} & \metricpair{21.2}{18.3} & \metricpair{30.2}{16.1} & \metricpair{40.2}{25.0} & \metricpair{30.8}{15.8} \\
 & A & \metricpair{63.4}{57.7} & \metricpair{49.4}{14.2} & \metricpair{55.2}{14.9} & \metricpair{89.7}{27.0} & \metricpair{84.4}{14.5} \\
 & B & \metricpair{24.7}{4.6} & \metricpair{46.1}{7.3} & \metricpair{52.8}{2.7} & \metricpair{90.2}{31.6} & \metricpair{83.6}{6.4} \\
 & C & \metricpair{65.3}{59.5} & \metricpair{56.2}{28.1} & \metricpair{56.8}{20.3} & \metricpair{90.5}{37.3} & \metricpair{85.0}{21.0} \\
 & GGPT & \metricpair{63.5}{58.1} & \metricpair{51.8}{23.1} & \metricpair{56.6}{22.8} & \metricpair{89.2}{43.3} & \metricpair{83.4}{32.5} \\
 & conditioned & \metricpair{50.4}{42.0} & \metricpair{39.5}{25.8} & \metricpair{44.8}{22.0} & \metricpair{82.9}{51.3} & \metricpair{79.2}{28.0} \\
\midrule
MASt3R$^\dagger$ & -- & \metricpair{65.6}{62.3} & \metricpair{40.6}{32.5} & \metricpair{44.4}{27.0} & \metricpair{26.5}{12.6} & \metricpair{70.1}{43.2} \\
 & A & \metricpair{64.3}{61.3} & \metricpair{53.6}{23.6} & \metricpair{55.5}{13.7} & \metricpair{88.6}{13.1} & \metricpair{87.3}{43.2} \\
 & B & \metricpair{64.6}{58.8} & \metricpair{58.0}{32.1} & \metricpair{58.6}{26.2} & \metricpair{89.8}{27.5} & \metricpair{88.6}{50.8} \\
 & C & \metricpair{68.4}{64.1} & \metricpair{61.4}{39.7} & \metricpair{58.5}{27.2} & \metricpair{90.4}{34.0} & \metricpair{88.1}{45.9} \\
 & GGPT & \metricpair{69.1}{64.9} & \metricpair{56.3}{33.2} & \metricpair{58.5}{29.4} & \metricpair{83.5}{23.2} & \metricpair{89.1}{56.5} \\
\midrule
\rowcolor{cyan!16} Scaffold3D (ours) & conditioned & \metricpair{\B{69.6}}{\B{65.8}} & \metricpair{\B{66.4}}{\B{49.1}} & \metricpair{\B{60.9}}{\B{35.5}} & \metricpair{92.9}{72.7} & \metricpair{\B{90.4}}{\B{66.4}} \\
\bottomrule
\end{tabular}
}
\label{tab:posthoc_alignment}
\end{table*}

%% file: tables/20_camera_refinement.tex
\begin{table}[!htbp]
\setlength{\tabcolsep}{4.0pt}
\centering
\caption{\textbf{Fixed vs.\ refined scaffold cameras.} AUC@10 of Scaffold3D when the SfM-anchored global alignment keeps the SfM cameras fixed (default) or is initialized from them and refines intrinsics and extrinsics. The better value per dataset is \B{bold}.}
\begin{tabular}{lcccc}
\toprule
Cameras & \dsid{ScanNet++$_4$} & \dscd{ETH3D$_4$} & \dscd{T\&T$_4$} & \dsod{MV-dVRK} \\
\midrule
Fixed SfM cameras & \B{69.6} & \B{66.4} & \B{60.9} & 90.4 \\
Refined & 69.4 & 65.9 & \B{60.9} & \B{90.6} \\
\bottomrule
\end{tabular}
\label{tab:camera_refinement}
\end{table}

%% file: tables/6_dataset_support_stats.tex
\begin{table*}[!tbp]
\setlength{\tabcolsep}{2.1pt}
\centering
\caption{\textbf{Dataset support statistics.} Views denotes the number of input views. GT is the fraction of pixels covered by the evaluation ground truth. The view-count columns split GT-covered pixels by the estimated number of views they are visible in. S and NS denote pixels with and without SfM-scaffold support, respectively. 4D-DRESS and MV-dVRK have few monocular pixels and high SfM-scaffold coverage, so they leave less room for gains from geometry propagation beyond SfM support. DexYCB-4A/B/C use camera subsets 1--4, 3--6, and 5--8, respectively; DexYCB-8 uses all eight views. DexYCB and EgoBody have much lower scaffold support due to difficult matching and triangulation. \textcolor{black}{\textbf{Black}} marks the easiest extremes in overlap and SfM-scaffold coverage, whereas \textcolor{red}{\textbf{red}} marks the most difficult extremes.}
\begin{tabular}{lcc cccccc cc}
\toprule
\multirow[b]{2}{*}{Dataset} & \multirow[b]{2}{*}{Views} & \multirow[b]{2}{*}{GT} & \multicolumn{6}{c}{Visible in} & \multicolumn{2}{c}{Coverage} \\
\cmidrule(lr){4-9} \cmidrule(lr){10-11}
 & & & 1 & \mbox{>1} & 2 & 3 & 4 & \mbox{$\ge\!4$} & S & NS \\
\midrule
\datasetboxplain{colorid}{ScanNet++$_4$} & 4 & 96.6 & \textcolor{red}{\U{24.0}} & 76.0 & 39.7 & 27.1 & 9.2 & 9.2 & 30.8 & 69.2 \\
\datasetboxplain{colorid}{ScanNet++$_8$} & 8 & 96.5 & 9.1 & 90.9 & 19.8 & 22.6 & 21.7 & 48.5 & 78.8 & 21.2 \\
\datasetboxplain{colorid}{ScanNet++$_{16}$} & 16 & 96.3 & 2.9 & 97.1 & 6.0 & 8.4 & 10.4 & 82.7 & 90.1 & 9.9 \\
\midrule
\datasetboxplain{colorcd}{ETH3D$_4$} & 4 & 22.6 & \textcolor{red}{\B{26.5}} & 73.5 & 30.7 & 32.5 & 10.4 & 10.4 & 52.0 & 48.0 \\
\datasetboxplain{colorcd}{ETH3D$_8$} & 8 & 21.5 & 12.9 & 87.1 & 18.7 & 22.0 & 17.9 & 46.4 & 83.3 & 16.7 \\
\datasetboxplain{colorcd}{ETH3D$_{16}$} & 16 & 21.4 & 3.9 & 96.1 & 9.6 & 13.8 & 13.8 & 72.7 & 92.4 & 7.6 \\
\midrule
\datasetboxplain{colorcd}{Tanks\&Temples$_4$} & 4 & 34.6 & 10.2 & 89.8 & 22.6 & 28.0 & 39.2 & 39.2 & 71.0 & 29.0 \\
\datasetboxplain{colorcd}{Tanks\&Temples$_8$} & 8 & 36.4 & 4.7 & 95.3 & 7.1 & 10.6 & 12.6 & 77.6 & \U{93.5} & 6.5 \\
\datasetboxplain{colorcd}{Tanks\&Temples$_{16}$} & 16 & 34.6 & 2.7 & 97.3 & 4.6 & 5.3 & 5.3 & 87.4 & \B{95.7} & 4.3 \\
\midrule
\datasetboxplain{colorod}{4D-DRESS} & 4--12 & 17.4 & 0.9 & \U{99.1} & 4.4 & 9.9 & 16.8 & 84.8 & 92.6 & 7.4 \\
\datasetboxplain{colorod}{MV-dVRK} & 4--12 & 83.5 & 0.4 & \B{99.6} & 13.2 & 17.5 & 17.5 & 68.8 & 88.7 & 11.3 \\
\midrule
\datasetboxplain{colorod}{DexYCB-4A} & 4 & 56.2 & 19.6 & 80.4 & 32.4 & 32.3 & 15.7 & 15.7 & 5.6 & \textcolor{red}{\U{94.4}} \\
\datasetboxplain{colorod}{DexYCB-4B} & 4 & 69.2 & 21.6 & 78.4 & 31.9 & 38.3 & 8.2 & 8.2 & 5.5 & \textcolor{red}{\B{94.5}} \\
\datasetboxplain{colorod}{DexYCB-4C} & 4 & 75.5 & 13.2 & 86.8 & 40.2 & 31.5 & 15.1 & 15.1 & 13.4 & 86.6 \\
\datasetboxplain{colorod}{DexYCB-8} & 8 & 66.6 & 7.2 & 92.8 & 11.6 & 13.7 & 18.9 & 67.5 & 7.7 & 92.3 \\
\datasetboxplain{colorod}{EgoBody-Exo} & 3/5 & 11.7 & 11.4 & 88.6 & 19.5 & 30.0 & 16.6 & 39.1 & 31.0 & 69.0 \\
\bottomrule
\end{tabular}
\label{tab:dataset_support_stats}
\end{table*}

%% file: tables/10_training_mixture.tex
\begin{table}[!tbp]
\setlength{\tabcolsep}{2.1pt}
\centering
\caption{\textbf{Training data mixture for the final fine-tuning stage.} Dataset selection and approximate ratios are mainly inspired by MASt3R~\cite{mast3r_eccv24}. The table reports the number of preprocessed training samples per source dataset after filtering and the input-view range sampled during preprocessing.}
\begin{tabular}{lrrc}
\toprule
Dataset & Samples & Share (\%) & Views \\
\midrule
BlendedMVS~\cite{bmvs} & 27,324 & 15.2 & 3--6 \\
MegaDepth~\cite{li2018megadepth} & 27,267 & 15.2 & 3--6 \\
ARKitScenes~\cite{baruch2021arkitscenes} & 21,656 & 12.1 & 3--6 \\
OmniWorld-Game~\cite{zhou2025omniworld} & 20,480 & 11.4 & 3--6 \\
ScanNet++ v2~\cite{yeshwanth2023scannet++} & 18,122 & 10.1 & 3--6 \\
Waymo~\cite{Sun_2020_CVPR_waymo} & 17,719 & 9.9 & 3--6 \\
Habitat~\cite{szot2021habitat} & 16,959 & 9.4 & 3--5 \\
Co3D-v2~\cite{reizenstein21common} & 11,445 & 6.4 & 3--6 \\
WildRGBD~\cite{xia2024rgbd} & 9,091 & 5.1 & 3--6 \\
StaticThings3D~\cite{mayer2016large} & 9,089 & 5.1 & 3--6 \\
Virtual KITTI 2~\cite{cabon2020virtual} & 464 & 0.3 & 3--6 \\
\midrule
\textbf{Total} & \textbf{179,616} & \textbf{100.0} & 3--6 \\
\bottomrule
\end{tabular}
\label{tab:training_mixture}
\end{table}

%% file: tables/23_training_curriculum.tex
\begin{table}[!htbp]
\centering
\setlength{\tabcolsep}{4pt}
\caption{\textbf{Training curriculum.} GT denotes ground-truth scaffolds sampled from multi-view-overlap regions. Percentages give the probability of using GT rather than the listed SfM source. Sample and update counts for stages 1--3 are approximate and derived from the training configurations. Only stage 4 uses the exact training mixture in \cref{tab:training_mixture}.}
\small
\begin{tabular}{@{}clrrrr@{}}
\toprule
Stage & Scaffold sampling & \shortstack{Samples\\per epoch} & \shortstack{Global\\batch} & Epochs & \shortstack{Optimizer\\updates} \\
\midrule
1 & 95\% GT / MP-SfM & $\approx$200K & 16 & 3 & $\approx$37.5K \\
2 & GT only & $\approx$200K & 32 & 10 & $\approx$62.5K \\
3 & 50\% GT / MP-SfM & $\approx$200K & 16 & 3 & $\approx$37.5K \\
4 & 40\% GT / GGPT-SfM & 179,616 & 16 & 6 & $\approx$67K \\
\bottomrule
\end{tabular}
\label{tab:training_curriculum}
\end{table}

%% file: tables/8_runtime_breakdown.tex
\begin{table}[!tp]
\setlength{\tabcolsep}{3.0pt}
\centering
\caption{\textbf{Runtime breakdown.} Wall-clock time per stage for Scaffold3D on one ETH3D scene per view count, in seconds. Total is the sum of the four stages. The sparse graph connects each view to its neighbor in input order plus two hub views to all views. At 2 and 4 views it keeps every pair. These timings use Pi3 to initialize the scaffold, so no initial pairwise DGA is required. The scaffold stage includes this initialization and matches all pairs in both graphs. Parentheses give the change with respect to the complete graph. SfM-anchored DGA accounts for the largest share of runtime at higher view counts.}
\resizebox{\linewidth}{!}{%
\begin{tabular}{@{}clrrrrrr@{}}
\toprule
Views & Graph & Pairs & \shortstack{SfM\\scaffold} & \shortstack{Initial\\forward} & \shortstack{Conditioned\\forward} & \shortstack{SfM-anchored\\DGA} & Total \\
\midrule
2 & complete = sparse & 1 & 4.3 & 0.1 & 0.1 & 4.3 & 8.8 \\
4 & complete = sparse & 6 & 16.3 & 0.7 & 0.9 & 18.0 & 35.9 \\
\midrule
8 & complete & 28 & 26.1 & 3.2 & 4.4 & 52.3 & 86.0 \\
8 & sparse & 17 ($-$39\%) & 26.1 & 1.9 & 2.2 & 27.4 & 57.6 ($-$33\%) \\
\midrule
16 & complete & 120 & 61.2 & 10.5 & 22.2 & 219.0 & 312.9 \\
16 & sparse & 41 ($-$66\%) & 61.2 & 3.7 & 7.8 & 79.5 & 152.2 ($-$51\%) \\
\midrule
32 & complete & 496 & 71.3 & 43.6 & 114.1 & 557.0 & 786.0 \\
32 & sparse & 89 ($-$82\%) & 71.3 & 7.6 & 21.5 & 120.1 & 220.5 ($-$72\%) \\
\midrule
50 & complete & 1225 & 128.8 & 109.5 & 277.2 & 1361.0 & 1876.5 \\
50 & sparse & 143 ($-$88\%) & 128.8 & 12.8 & 33.2 & 186.8 & 361.6 ($-$81\%) \\
\bottomrule
\end{tabular}
}
\label{tab:runtime_breakdown}
\end{table}

%% file: tables/22_scaling_full.tex
\begin{table*}[!tbp]
\setlength{\tabcolsep}{3.0pt}
\centering
\caption{\textbf{Accuracy with more views.} AUC@1/5/10 on the full ETH3D set at each view count. At 4 views the sparse graph keeps every pair, so both graphs are identical.}
\resizebox{\linewidth}{!}{%
\begin{tabular}{l rrr c rrr c rrr c rrr c rrr}
\toprule
\multirow[b]{2}{*}{Method} & \multicolumn{3}{c}{4 views} & & \multicolumn{3}{c}{8 views} & & \multicolumn{3}{c}{16 views} & & \multicolumn{3}{c}{32 views} & & \multicolumn{3}{c}{50 views} \\
\cmidrule(lr){2-4} \cmidrule(lr){6-8} \cmidrule(lr){10-12} \cmidrule(lr){14-16} \cmidrule(lr){18-20}
 & @1 & @5 & @10 & & @1 & @5 & @10 & & @1 & @5 & @10 & & @1 & @5 & @10 & & @1 & @5 & @10 \\
\midrule
Scaffold3D (sparse graph) & \B{25.6} & \B{53.6} & \B{66.4} & & \B{29.2} & \U{55.4} & \U{66.9} & & \B{29.3} & \B{58.1} & \B{69.9} & & \B{28.8} & \B{57.1} & \B{67.8} & & \B{26.6} & \B{50.8} & \B{60.3} \\
Scaffold3D (complete graph) & \B{25.6} & \B{53.6} & \B{66.4} & & \U{29.0} & \B{55.7} & \B{67.4} & & \U{28.2} & \U{55.7} & \U{67.0} & & \U{21.0} & \U{43.0} & \U{51.3} & & \U{9.6} & \U{19.8} & 24.0 \\
Pi3X~\cite{wang2025pi3} & \U{9.5} & \U{33.3} & \U{47.3} & & 5.6 & 27.1 & 42.3 & & 3.7 & 23.5 & 38.9 & & 3.4 & 19.5 & 32.7 & & 3.3 & \U{19.8} & \U{32.9} \\
Pi3~\cite{wang2025pi3} & 6.2 & 29.2 & 44.1 & & 3.9 & 21.5 & 35.7 & & 2.4 & 19.7 & 34.9 & & 2.1 & 14.8 & 26.5 & & 2.3 & 14.4 & 25.3 \\
VGGT~\cite{wang2025vggt} & 6.4 & 26.3 & 41.4 & & 4.7 & 23.5 & 37.1 & & 3.8 & 18.6 & 31.2 & & 3.4 & 15.6 & 26.0 & & 3.5 & 16.2 & 26.1 \\
MapAnything~\cite{keetha2025mapanything} & 1.7 & 11.2 & 21.2 & & 1.2 & 8.7 & 17.0 & & 0.8 & 6.5 & 13.5 & & 0.8 & 5.6 & 11.6 & & 0.7 & 5.5 & 11.2 \\
\bottomrule
\end{tabular}
}
\label{tab:scaling_full}
\end{table*}

%% file: tables/4_supplemental_ood_results.tex
\begin{table*}[!htbp]
\setlength{\tabcolsep}{2.1pt}
\centering
\caption{\textbf{Additional out-of-domain datasets.} We report the same AUC@1/5/10/M format as \cref{tab:main_h16_results} on hand-object and human-scene datasets. These datasets test reconstruction with limited and partly inaccurate SfM support. Conditioning improves Scaffold3D over its conditioning-free pass (MASt3R$^\dagger$) in AUC@10 on every split, although Pi3 remains more accurate. Pi3X benefits from conditioning on three of the five splits. DexYCB-4A/B/C use camera subsets 1--4, 3--6, and 5--8, respectively; DexYCB-8 uses all eight views. EgoBody-Exo uses 3--5 exocentric cameras.}
\resizebox{\linewidth}{!}{%
\begin{tabular}{l crrrr crrrr crrrr crrrr crrrr}
\toprule
\multirow[b]{3}{*}{Method} & & \multicolumn{4}{c}{\dsod{DexYCB-4A}} & & \multicolumn{4}{c}{\dsod{DexYCB-4B}} & & \multicolumn{4}{c}{\dsod{DexYCB-4C}} & & \multicolumn{4}{c}{\dsod{DexYCB-8}} & & \multicolumn{4}{c}{\dsod{EgoBody-Exo}} \\
\cmidrule(lr){3-6} \cmidrule(lr){8-11} \cmidrule(lr){13-16} \cmidrule(lr){18-21} \cmidrule(lr){23-26}
& & \multicolumn{4}{c}{\textbf{AUC@1/5/10/M}} & & \multicolumn{4}{c}{\textbf{AUC@1/5/10/M}} & & \multicolumn{4}{c}{\textbf{AUC@1/5/10/M}} & & \multicolumn{4}{c}{\textbf{AUC@1/5/10/M}} & & \multicolumn{4}{c}{\textbf{AUC@1/5/10/M}} \\
& & \multicolumn{4}{c}{\emph{M pixels: 19.6\%}} & & \multicolumn{4}{c}{\emph{M pixels: 21.6\%}} & & \multicolumn{4}{c}{\emph{M pixels: 13.2\%}} & & \multicolumn{4}{c}{\emph{M pixels: 7.2\%}} & & \multicolumn{4}{c}{\emph{M pixels: 11.4\%}} \\
\midrule
\addlinespace[0.15em]
\multicolumn{26}{l}{\emph{Feed-forward predictors}} \\
\midrule
MASt3R-SfM~\cite{duisterhof2025mastrsfm} &  & 6.6 & 16.9 & 22.7 & 10.1 &  & 13.9 & 29.9 & 36.9 & 36.2 &  & 33.9 & 54.0 & 61.2 & 35.7 &  & 23.4 & 38.3 & 45.1 & 20.5 &  & 2.5 & 18.2 & 33.5 & 16.5 \\
Pow3R~\cite{pow3r_cvpr25} &  & 12.0 & 24.7 & 30.0 & 10.4 &  & 11.1 & 34.8 & 46.8 & 22.6 &  & 6.8 & 33.5 & 47.7 & 13.4 &  & 7.4 & 32.2 & 43.9 & 6.1 &  & 2.5 & 18.4 & 34.3 & 20.3 \\
DUSt3R~\cite{wang2024dust3r} &  & 14.1 & 28.4 & 35.8 & 14.6 &  & 12.8 & 37.2 & 47.5 & 25.8 &  & 9.5 & 39.8 & 52.0 & 17.0 &  & 9.6 & 38.0 & 50.7 & 9.6 &  & \B{3.2} & 19.5 & 34.8 & 17.8 \\
MapAnything~\cite{keetha2025mapanything} &  & 12.4 & 43.4 & 57.9 & 27.9 &  & 17.3 & 57.1 & 70.2 & 41.8 &  & 12.6 & 58.7 & 72.0 & 33.0 &  & 12.1 & 54.2 & 69.0 & 19.2 &  & 2.1 & 12.4 & 23.6 & 21.2 \\
MASt3R$^\dagger$ &  & 13.3 & 46.3 & 59.3 & 28.7 &  & 20.0 & 53.5 & 66.5 & 43.0 &  & 25.6 & 57.1 & 68.0 & 31.3 &  & 29.6 & 61.8 & 73.0 & 24.1 &  & 2.3 & 19.0 & 35.1 & 15.7 \\
MASt3R~\cite{mast3r_eccv24} &  & 17.4 & 50.2 & 64.2 & 33.3 &  & 21.9 & 53.8 & 66.3 & 45.3 &  & 31.8 & 64.8 & 75.2 & 37.1 &  & 30.8 & 63.6 & 75.0 & 32.6 &  & 2.7 & 19.1 & 34.9 & 18.0 \\
VGGT~\cite{wang2025vggt} &  & 11.1 & 48.1 & 65.4 & 34.9 &  & 28.6 & 68.0 & 79.8 & 58.0 &  & 17.8 & 57.8 & 73.3 & 37.2 &  & 15.6 & 59.3 & 75.0 & 31.1 &  & 2.8 & 22.2 & 40.3 & \U{29.1} \\
\textcolor{black!55}{VGGT-Omega$^\ddagger$~\cite{wang2025vggt}} & & \textcolor{black!55}{25.7} & \textcolor{black!55}{66.7} & \textcolor{black!55}{78.7} & \textcolor{black!55}{52.1} & & \textcolor{black!55}{46.7} & \textcolor{black!55}{77.4} & \textcolor{black!55}{84.9} & \textcolor{black!55}{66.2} & & \textcolor{black!55}{32.5} & \textcolor{black!55}{71.9} & \textcolor{black!55}{81.7} & \textcolor{black!55}{46.6} & & \textcolor{black!55}{38.5} & \textcolor{black!55}{76.2} & \textcolor{black!55}{84.9} & \textcolor{black!55}{45.6} & & \textcolor{black!55}{2.8} & \textcolor{black!55}{18.1} & \textcolor{black!55}{34.9} & \textcolor{black!55}{22.1} \\
Pi3X~\cite{wang2025pi3} &  & \U{27.5} & \B{68.5} & \U{78.8} & \U{44.1} &  & 15.6 & 63.9 & 76.9 & 58.8 &  & 34.5 & 73.1 & 82.9 & 48.1 &  & 35.4 & 75.5 & 84.8 & 41.9 &  & 3.0 & \U{23.4} & \U{42.9} & \B{29.3} \\
Pi3~\cite{wang2025pi3} &  & \B{27.6} & \U{68.2} & \B{78.9} & \B{46.3} &  & \B{34.7} & \B{73.1} & \B{82.4} & \B{63.8} &  & \B{42.3} & \U{75.9} & \U{84.5} & \U{51.6} &  & 40.9 & \U{75.8} & \U{85.0} & \U{43.0} &  & 2.8 & 23.1 & 41.8 & 23.5 \\
\midrule
\addlinespace[0.15em]
\multicolumn{26}{l}{\emph{Monocular depth completion}} \\
\midrule
OMNI-DC~\cite{zuo2025omni} &  & 22.8 & 50.6 & 62.8 & 27.2 &  & 23.8 & 52.8 & 64.8 & 25.5 &  & 30.0 & 66.3 & 77.6 & 38.3 &  & 33.7 & 68.0 & 78.2 & 23.3 &  & 2.3 & 16.4 & 31.4 & 6.8 \\
Marigold-DC~\cite{viola2024marigolddc} &  & 22.8 & 51.9 & 63.7 & 31.2 &  & 26.0 & 58.2 & 70.1 & 38.9 &  & 36.0 & 72.0 & 81.3 & 42.8 &  & \U{41.7} & 72.9 & 81.3 & 30.5 &  & 2.6 & 19.3 & 35.8 & 11.3 \\
\midrule
\addlinespace[0.15em]
\multicolumn{26}{l}{\emph{Point-token propagation}} \\
\midrule
GGPT (MASt3R$^\dagger$)~\cite{chen2026ggpt} &  & 17.7 & 50.7 & 62.8 & 28.2 &  & 25.1 & 57.8 & 69.9 & 39.0 &  & 32.0 & 67.4 & 76.6 & 35.3 &  & 38.3 & 70.8 & 79.3 & 25.4 &  & 2.7 & 21.2 & 39.0 & 16.6 \\
GGPT (VGGT)~\cite{chen2026ggpt} &  & 20.5 & 54.6 & 68.7 & 36.4 &  & 29.2 & 64.2 & 75.8 & 50.8 &  & 37.1 & 73.3 & 82.7 & 45.1 &  & 38.2 & 73.2 & 82.6 & 36.6 &  & 2.6 & \B{23.5} & \B{43.3} & 28.4 \\
GGPT (Pi3X)~\cite{chen2026ggpt} &  & 24.3 & 59.9 & 71.3 & 37.9 &  & 27.3 & 62.9 & 74.9 & 50.7 &  & \U{41.2} & \U{75.9} & 84.2 & 50.4 &  & 40.7 & 74.8 & 83.3 & 38.3 &  & 2.5 & 22.9 & 41.8 & 25.3 \\
GGPT (Pi3)~\cite{chen2026ggpt} &  & 26.3 & 61.5 & 73.1 & 41.9 &  & \U{32.5} & 66.7 & 77.0 & 53.4 &  & \B{42.3} & \B{76.3} & \B{84.7} & \B{52.7} &  & \B{42.2} & \U{75.8} & 84.3 & 42.8 &  & 2.0 & 21.2 & 40.6 & 21.8 \\
\midrule
\addlinespace[0.15em]
\multicolumn{26}{l}{\emph{Geometry-conditioned predictors}} \\
\midrule
MVSAnywhere~\cite{izquierdo2025mvsanywhere} &  & 6.5 & 20.0 & 29.6 & 9.1 &  & 14.0 & 37.6 & 50.6 & 30.2 &  & 18.9 & 50.6 & 64.9 & 32.1 &  & 16.7 & 43.9 & 58.2 & 16.9 &  & 2.5 & 16.0 & 30.3 & 11.2 \\
Pow3R~\cite{pow3r_cvpr25} &  & 14.6 & 39.4 & 51.6 & 20.8 &  & 19.4 & 44.5 & 56.5 & 34.0 &  & 29.1 & 64.1 & 75.3 & 40.0 &  & 27.0 & 57.4 & 68.8 & 25.7 &  & 2.9 & 20.5 & 36.6 & 14.2 \\
MapAnything~\cite{keetha2025mapanything} &  & 17.1 & 43.5 & 53.7 & 20.3 &  & 20.7 & 52.7 & 64.1 & 37.2 &  & 30.4 & 59.3 & 67.8 & 30.3 &  & 36.7 & 63.7 & 71.1 & 19.1 &  & 2.2 & 18.2 & 34.5 & 14.9 \\
Pi3X~\cite{wang2025pi3} &  & 23.2 & 61.2 & 74.3 & 43.3 &  & 29.1 & \U{69.3} & \U{80.2} & \U{62.4} &  & 40.1 & 75.5 & 83.9 & 50.4 &  & 39.8 & \B{77.2} & \B{86.1} & \B{49.3} &  & 2.8 & 21.2 & 38.8 & 21.6 \\
\rowcolor{cyan!16} Scaffold3D (ours) &  & 22.1 & 53.3 & 65.7 & 30.7 &  & 30.2 & 61.8 & 72.5 & 48.6 &  & 34.8 & 66.9 & 76.3 & 39.5 &  & 40.2 & 70.2 & 79.3 & 31.7 &  & \U{3.1} & 21.2 & 38.4 & 21.8 \\
\bottomrule
\end{tabular}
}
\label{tab:supplemental_ood_results}
\end{table*}

%% file: tables/14_supplemental_camera_diagnostics.tex
\begin{table*}[!htbp]
\setlength{\tabcolsep}{2.1pt}
\centering
\caption{\textbf{Supplemental camera diagnostics.} We report camera-pose AUC at 1, 5, and 10 degrees and F$^\circ$, the mean field-of-view error in degrees, for the datasets in \cref{tab:supplemental_ood_results}. These diagnostics separate camera accuracy from dense reconstruction quality on the hand-object and human-centric stress tests. Lower is better for F$^\circ$.}
\resizebox{\linewidth}{!}{%
\begin{tabular}{l crrrr crrrr crrrr crrrr crrrr}
\toprule
\multirow[b]{2}{*}{Method} & & \multicolumn{4}{c}{\dsod{DexYCB-4A}} & & \multicolumn{4}{c}{\dsod{DexYCB-4B}} & & \multicolumn{4}{c}{\dsod{DexYCB-4C}} & & \multicolumn{4}{c}{\dsod{DexYCB-8}} & & \multicolumn{4}{c}{\dsod{EgoBody-Exo}} \\
\cmidrule(lr){3-6} \cmidrule(lr){8-11} \cmidrule(lr){13-16} \cmidrule(lr){18-21} \cmidrule(lr){23-26}
& & \multicolumn{4}{c}{\textbf{AUC@1/5/10/F$^\circ$}} & & \multicolumn{4}{c}{\textbf{AUC@1/5/10/F$^\circ$}} & & \multicolumn{4}{c}{\textbf{AUC@1/5/10/F$^\circ$}} & & \multicolumn{4}{c}{\textbf{AUC@1/5/10/F$^\circ$}} & & \multicolumn{4}{c}{\textbf{AUC@1/5/10/F$^\circ$}} \\
\midrule
\addlinespace[0.15em]
\multicolumn{26}{l}{\emph{Feed-forward predictors}} \\
\midrule
DUSt3R~\cite{wang2024dust3r} &  & 0.0 & 11.0 & 20.6 & 1.9 &  & 0.0 & 12.5 & 34.2 & 2.6 &  & 0.8 & 21.5 & 44.4 & 2.6 &  & 0.0 & 17.3 & 42.5 & 2.1 &  & 27.5 & 68.0 & 82.7 & 1.1 \\
MASt3R~\cite{mast3r_eccv24} &  & 3.3 & 54.7 & 77.2 & 1.3 &  & 1.7 & 41.5 & 62.8 & 1.8 &  & 7.5 & 56.7 & 76.5 & 1.5 &  & 13.4 & 64.3 & 80.5 & 1.3 &  & 38.9 & 83.4 & 91.7 & 0.9 \\
MASt3R-SfM~\cite{duisterhof2025mastrsfm} &  & 16.7 & 22.3 & 27.5 & \B{0.8} &  & \B{23.4} & 31.3 & 32.4 & \U{1.1} &  & 16.7 & 82.6 & 91.4 & \B{0.5} &  & 48.6 & 58.4 & 60.6 & \B{0.5} &  & 36.4 & 83.2 & 91.6 & \B{0.6} \\
Pow3R~\cite{pow3r_cvpr25} &  & 0.0 & 5.3 & 10.3 & 3.6 &  & 0.0 & 7.0 & 27.8 & 2.8 &  & 0.0 & 10.2 & 36.7 & 2.2 &  & 0.0 & 10.9 & 27.9 & 2.5 &  & 26.0 & 74.9 & 87.2 & 1.7 \\
MapAnything~\cite{keetha2025mapanything} &  & 5.8 & 52.5 & 74.5 & 1.3 &  & 1.7 & 41.8 & 69.3 & 1.4 &  & 4.2 & 58.7 & 78.5 & 1.6 &  & 2.7 & 54.3 & 76.2 & 1.5 &  & 19.6 & 66.8 & 83.1 & 4.2 \\
VGGT~\cite{wang2025vggt} &  & 8.3 & 59.5 & 78.9 & 1.1 &  & 9.2 & 66.7 & 82.9 & \B{1.0} &  & 2.5 & 38.0 & 61.9 & 2.0 &  & 13.4 & 65.5 & 82.2 & \U{0.7} &  & 49.7 & 83.1 & 91.5 & 0.9 \\
Pi3~\cite{wang2025pi3} &  & \B{26.7} & \B{77.8} & \B{88.9} & 1.4 &  & 4.2 & 58.0 & 78.5 & \B{1.0} &  & 18.3 & 67.8 & 83.8 & \U{0.7} &  & 15.2 & 72.0 & 85.8 & 0.9 &  & 48.9 & 83.8 & 91.9 & \U{0.8} \\
Pi3X~\cite{wang2025pi3} &  & \U{17.5} & \U{73.0} & \U{86.5} & 1.0 &  & 6.7 & 57.8 & 75.8 & 1.7 &  & 15.8 & 74.3 & 87.2 & 1.8 &  & 16.4 & 76.8 & 88.4 & \U{0.7} &  & 42.9 & 85.1 & 92.5 & \B{0.6} \\
MASt3R$^\dagger$ &  & 5.0 & 45.3 & 69.0 & 1.3 &  & 4.2 & 36.7 & 63.5 & \U{1.1} &  & 5.0 & 48.5 & 72.1 & 1.3 &  & 6.6 & 57.7 & 77.2 & 1.4 &  & 37.0 & 79.5 & 89.8 & 1.5 \\
\midrule
\addlinespace[0.15em]
\multicolumn{26}{l}{\emph{Monocular depth completion}} \\
\midrule
OMNI-DC~\cite{zuo2025omni} &  & 12.5 & 62.7 & 80.5 & \U{0.9} &  & 14.2 & 73.8 & 86.9 & 2.1 &  & 46.7 & 83.2 & 91.6 & 1.5 &  & \B{66.8} & \B{92.6} & \B{96.3} & 1.0 &  & 55.7 & 86.6 & 93.3 & \U{0.8} \\
Marigold-DC~\cite{viola2024marigolddc} &  & 12.5 & 62.7 & 80.5 & \U{0.9} &  & 14.2 & 73.8 & 86.9 & 2.1 &  & 46.7 & 83.2 & 91.6 & 1.5 &  & \B{66.8} & \B{92.6} & \B{96.3} & 1.0 &  & 55.7 & 86.6 & 93.3 & \U{0.8} \\
\midrule
\addlinespace[0.15em]
\multicolumn{26}{l}{\emph{Point-token propagation}} \\
\midrule
GGPT (VGGT)~\cite{chen2026ggpt} &  & 10.0 & 63.3 & 81.2 & 1.1 &  & 14.2 & 70.7 & 85.3 & 2.2 &  & \B{52.5} & \U{85.3} & \U{92.7} & 1.5 &  & 48.2 & 88.5 & 94.2 & 1.0 &  & 54.2 & 86.7 & 93.3 & 0.9 \\
GGPT (Pi3)~\cite{chen2026ggpt} &  & 14.2 & 60.7 & 79.0 & 1.0 &  & 14.2 & \U{75.0} & \U{87.5} & 2.1 &  & 42.5 & 83.3 & 91.7 & 1.7 &  & 52.0 & 89.5 & 94.8 & 1.1 &  & \U{56.0} & \U{86.8} & \U{93.4} & \U{0.8} \\
GGPT (Pi3X)~\cite{chen2026ggpt} &  & 15.0 & 66.3 & 82.1 & \U{0.9} &  & \U{15.8} & \B{75.5} & \B{87.8} & 2.1 &  & \U{51.7} & \B{86.0} & \B{93.0} & 1.4 &  & 51.2 & 89.2 & 94.6 & 1.0 &  & 53.8 & \U{86.8} & \U{93.4} & \U{0.8} \\
GGPT (MASt3R$^\dagger$)~\cite{chen2026ggpt} &  & 12.5 & 62.7 & 80.5 & \U{0.9} &  & 14.2 & 73.8 & 86.9 & 2.1 &  & 46.7 & 83.2 & 91.6 & 1.5 &  & \B{66.8} & \B{92.6} & \B{96.3} & 1.0 &  & 55.7 & 86.6 & 93.3 & \U{0.8} \\
\midrule
\addlinespace[0.15em]
\multicolumn{26}{l}{\emph{Geometry-conditioned predictors}} \\
\midrule
MVSAnywhere~\cite{izquierdo2025mvsanywhere} &  & 12.5 & 62.7 & 80.5 & \U{0.9} &  & 14.2 & 73.8 & 86.9 & 2.1 &  & 46.7 & 83.2 & 91.6 & 1.5 &  & \B{66.8} & \B{92.6} & \B{96.3} & 1.0 &  & 55.7 & 86.6 & 93.3 & \U{0.8} \\
MapAnything~\cite{keetha2025mapanything} &  & 15.0 & 65.0 & 81.6 & 1.1 &  & 10.0 & 71.8 & 85.9 & 2.3 &  & 45.0 & 83.7 & 91.8 & 1.6 &  & \U{58.0} & \U{90.8} & \U{95.4} & 1.4 &  & \B{58.3} & \B{87.1} & \B{93.6} & 0.9 \\
Pi3X~\cite{wang2025pi3} &  & 12.5 & 62.7 & 80.5 & \U{0.9} &  & 14.2 & 73.8 & 86.9 & 2.1 &  & 46.7 & 83.2 & 91.6 & 1.5 &  & \B{66.8} & \B{92.6} & \B{96.3} & 1.0 &  & 55.7 & 86.6 & 93.3 & \U{0.8} \\
Pow3R~\cite{pow3r_cvpr25} &  & 12.5 & 62.7 & 80.5 & \U{0.9} &  & 14.2 & 73.8 & 86.9 & 2.1 &  & 46.7 & 83.2 & 91.6 & 1.5 &  & \B{66.8} & \B{92.6} & \B{96.3} & 1.0 &  & 55.7 & 86.6 & 93.3 & \U{0.8} \\
\rowcolor{cyan!16} Scaffold3D (ours) &  & 12.5 & 62.7 & 80.5 & \U{0.9} &  & 14.2 & 73.8 & 86.9 & 2.1 &  & 46.7 & 83.2 & 91.6 & 1.5 &  & \B{66.8} & \B{92.6} & \B{96.3} & 1.0 &  & 55.7 & 86.6 & 93.3 & \U{0.8} \\
\bottomrule
\end{tabular}
}
\label{tab:supplemental_camera_diagnostics}
\end{table*}

%% file: tables/19_ood_gt_scaffold.tex
\begin{table}[!htbp]
\setlength{\tabcolsep}{3.0pt}
\centering
\caption{\textbf{What limits the scaffold on DexYCB and EgoBody.} AUC@10 of Scaffold3D when ground-truth points, ground-truth cameras, or both are substituted into the estimated scaffold. Components not replaced by ground truth remain estimated. Coverage (in parentheses) is the scaffold's pixel coverage. Substituting points does not change the support, so coverage is shown once per camera setting. Correct cameras tighten the epipolar and score filters and reduce coverage, most sharply on EgoBody-Exo. Feed-forward results from \cref{tab:supplemental_ood_results} are given for reference.}
\resizebox{\linewidth}{!}{%
\begin{tabular}{l ccc cccc}
\toprule
 & \multicolumn{3}{c}{Feed-forward} & \multicolumn{4}{c}{Scaffold3D} \\
\cmidrule(lr){2-4} \cmidrule(lr){5-8}
Dataset & Pi3 & Pi3X & MASt3R$^\dagger$ & Estimated (cov.) & GT points & GT cameras (cov.) & GT points and cameras \\
\midrule
DexYCB-4A & \B{78.9} & \U{78.8} & 59.3 & 65.7 (5.6\%) & 71.8 & 67.7 (4.1\%) & 75.0 \\
DexYCB-4B & \B{82.4} & 76.9 & 66.5 & 72.5 (5.5\%) & 81.0 & 74.7 (4.0\%) & \U{82.2} \\
DexYCB-4C & \B{84.5} & \U{82.9} & 68.0 & 76.3 (13.4\%) & 77.7 & 78.7 (13.3\%) & 79.2 \\
DexYCB-8 & \B{85.0} & \U{84.8} & 73.0 & 79.3 (7.7\%) & 82.1 & 78.8 (5.7\%) & 81.9 \\
EgoBody-Exo & 41.8 & 42.9 & 35.1 & 38.4 (31.0\%) & \U{44.1} & 44.0 (5.8\%) & \B{45.3} \\
\bottomrule
\end{tabular}
}
\label{tab:ood_gt_scaffold}
\end{table}

%% file: tables/17_scaffold_source_full.tex
\begin{table*}[!htbp]
\centering
\caption{\textbf{Scaffold-source study: per-dataset results and scaffold quality.} Extends \cref{tab:scaffold_quality} (left). For each dataset, @10 is the AUC@10 of Scaffold3D conditioned on that scaffold. The remaining columns describe the scaffold itself: camera pose AUC@5$^\circ$ (pose), mean field-of-view error in degrees (F$^\circ$), pixel coverage in \% of ground-truth-valid pixels (cov), and Chamfer accuracy and completeness at 5\,cm (5\,mm for MV-dVRK) after Sim(3) alignment (acc, comp). C and F denote matching on 512$\times$384 inputs and full-resolution 4:3 center crops; SP+LG denotes SuperPoint+LightGlue. The ``None'' rows use no scaffold, and their pose and F$^\circ$ columns describe the feed-forward cameras. Lower is better for F$^\circ$. ``Best FF'' is the per-dataset best feed-forward baseline ranked in \cref{tab:main_h16_results}: Pi3 on ScanNet++ and MASt3R-SfM on the other datasets.}
\resizebox{\linewidth}{!}{%
\setlength{\tabcolsep}{2.0pt}
\begin{tabular}{ll c rrrrrr c rrrrrr c rrrrrr}
\toprule
\multirow[b]{2}{*}{Scaffold source} & \multirow[b]{2}{*}{Res.} & & \multicolumn{6}{c}{\dsid{ScanNet++$_4$}} & & \multicolumn{6}{c}{\dscd{ETH3D$_4$}} & & \multicolumn{6}{c}{\dscd{Tanks\&Temples$_4$}} \\
\cmidrule(lr){4-9} \cmidrule(lr){11-16} \cmidrule(lr){18-23}
 & & & @10$\uparrow$ & pose$\uparrow$ & F$^\circ\downarrow$ & cov$\uparrow$ & acc$\uparrow$ & comp$\uparrow$ & & @10$\uparrow$ & pose$\uparrow$ & F$^\circ\downarrow$ & cov$\uparrow$ & acc$\uparrow$ & comp$\uparrow$ & & @10$\uparrow$ & pose$\uparrow$ & F$^\circ\downarrow$ & cov$\uparrow$ & acc$\uparrow$ & comp$\uparrow$ \\
\midrule
None (MASt3R$^\dagger$) & -- &  & 65.6 & 74.5 & 1.5 & -- & -- & -- &  & 40.6 & 43.1 & 3.0 & -- & -- & -- &  & 44.4 & 61.0 & 1.5 & -- & -- & -- \\
None (best FF) & -- &  & 68.0 & 84.0 & 0.7 & -- & -- & -- &  & 49.3 & 76.3 & 0.5 & -- & -- & -- &  & 52.9 & 90.0 & 0.7 & -- & -- & -- \\
\midrule
SP+LG~\cite{superpoint,lindenberger2023lightglue} & -- &  & 64.6 & 85.7 & 0.5 & 0.3 & 79.4 & 10.7 &  & 42.7 & 76.4 & 1.5 & 0.4 & 57.8 & 7.1 &  & 49.4 & 92.6 & \B{0.3} & 0.7 & 53.1 & 14.7 \\
UFM~\cite{zhang2025ufm} & C &  & 58.9 & 81.1 & \U{0.2} & 19.6 & 81.6 & 44.8 &  & 49.7 & 78.6 & 0.9 & 36.8 & 63.4 & 43.5 &  & 49.5 & 93.6 & 0.8 & 48.0 & 59.1 & 54.5 \\
UFM~\cite{zhang2025ufm} & F &  & 59.9 & 79.9 & 0.5 & 20.7 & 82.6 & 44.3 &  & 52.1 & 83.6 & 0.5 & 40.9 & 67.6 & 47.4 &  & 49.8 & 92.9 & 0.7 & 47.2 & 59.0 & 55.3 \\
LoMa~\cite{nordstrom2026loma} & C &  & 65.8 & 86.3 & 0.4 & 0.2 & 79.0 & 13.2 &  & 47.9 & 84.7 & 0.5 & 0.2 & 51.6 & 6.7 &  & 50.8 & 94.6 & \U{0.4} & 0.5 & 49.4 & 13.2 \\
RoMa~\cite{edstedt2024roma} & F &  & 58.1 & 69.2 & 1.2 & 4.9 & 89.5 & 13.5 &  & 55.7 & 85.1 & 0.5 & 21.3 & 77.3 & 30.3 &  & 55.9 & 91.3 & 0.5 & 37.7 & \B{69.7} & 49.7 \\
MASt3R~\cite{mast3r_eccv24} & F &  & 64.5 & 87.4 & 0.6 & 13.0 & 86.5 & \U{52.2} &  & 50.5 & 83.3 & 0.5 & 22.1 & 64.2 & 53.8 &  & 52.4 & 93.1 & 0.9 & 34.2 & 58.9 & 66.9 \\
RoMa~\cite{edstedt2024roma} & C &  & 60.8 & 74.8 & 0.6 & 6.8 & 85.4 & 17.1 &  & 54.7 & 83.7 & 0.3 & 18.5 & 74.0 & 28.0 &  & 56.9 & 91.0 & 0.5 & 44.2 & \U{68.2} & 54.5 \\
MASt3R~\cite{mast3r_eccv24} & C &  & 64.9 & 86.1 & 0.5 & 13.1 & 87.2 & \B{52.6} &  & 50.8 & 83.8 & 0.4 & 21.7 & 65.6 & 53.4 &  & 52.8 & 92.9 & 0.8 & 34.4 & 59.1 & 66.9 \\
LoMa~\cite{nordstrom2026loma} & F &  & 66.1 & 89.8 & \U{0.2} & 0.3 & 84.5 & 12.6 &  & 53.1 & 96.0 & \B{0.1} & 0.3 & 62.5 & 8.3 &  & 51.2 & \U{94.8} & \U{0.4} & 0.5 & 55.3 & 13.9 \\
\rowcolor{cyan!16} RoMa v2~\cite{edstedt2025romav2} & C &  & \B{69.3} & \U{92.8} & \B{0.1} & 30.8 & \U{90.3} & 50.0 &  & \U{66.1} & \U{96.3} & \U{0.2} & 52.0 & \U{77.5} & \U{56.9} &  & \B{60.5} & \B{97.0} & 0.5 & 71.0 & 65.7 & \B{70.3} \\
RoMa v2~\cite{edstedt2025romav2} & F &  & \U{69.2} & \B{93.6} & \B{0.1} & 32.4 & \B{90.8} & 50.8 &  & \B{71.0} & \B{98.3} & \B{0.1} & 61.8 & \B{81.6} & \B{65.1} &  & \U{60.0} & 94.3 & 0.7 & 71.6 & 65.2 & \U{69.0} \\
\bottomrule
\end{tabular}}

\vspace{0.6em}
\resizebox{0.7\linewidth}{!}{%
\setlength{\tabcolsep}{2.0pt}
\begin{tabular}{ll c rrrrrr c rrrrrr}
\toprule
\multirow[b]{2}{*}{Scaffold source} & \multirow[b]{2}{*}{Res.} & & \multicolumn{6}{c}{\dsod{4D-DRESS}} & & \multicolumn{6}{c}{\dsod{MV-dVRK}} \\
\cmidrule(lr){4-9} \cmidrule(lr){11-16}
 & & & @10$\uparrow$ & pose$\uparrow$ & F$^\circ\downarrow$ & cov$\uparrow$ & acc$\uparrow$ & comp$\uparrow$ & & @10$\uparrow$ & pose$\uparrow$ & F$^\circ\downarrow$ & cov$\uparrow$ & acc$\uparrow$ & comp$\uparrow$ \\
\midrule
None (MASt3R$^\dagger$) & -- &  & 26.5 & 0.3 & 29.2 & -- & -- & -- &  & 70.1 & 29.0 & 2.5 & -- & -- & -- \\
None (best FF) & -- &  & 76.9 & 67.5 & 1.0 & -- & -- & -- &  & 75.8 & 78.9 & 1.9 & -- & -- & -- \\
\midrule
SP+LG~\cite{superpoint,lindenberger2023lightglue} & -- &  & 68.2 & 79.7 & 3.4 & 2.6 & 89.0 & 76.9 &  & 66.5 & 59.8 & 1.4 & 0.4 & 92.3 & 56.9 \\
UFM~\cite{zhang2025ufm} & C &  & 77.9 & 51.5 & 4.7 & 83.5 & 77.4 & 94.2 &  & 70.0 & 56.9 & 1.3 & 72.7 & 92.1 & 89.4 \\
UFM~\cite{zhang2025ufm} & F &  & 77.5 & 58.0 & 3.8 & 83.1 & 75.2 & 93.0 &  & 71.4 & 57.9 & 1.4 & 72.1 & 93.6 & 89.8 \\
LoMa~\cite{nordstrom2026loma} & C &  & 81.1 & 89.1 & \U{0.4} & 2.7 & 97.8 & 93.9 &  & 75.8 & 82.7 & 1.6 & 0.6 & 91.2 & 83.7 \\
RoMa~\cite{edstedt2024roma} & F &  & 83.7 & 84.1 & 0.5 & 70.2 & 97.2 & 90.0 &  & 73.4 & 62.6 & 1.8 & 51.9 & 96.4 & 83.2 \\
MASt3R~\cite{mast3r_eccv24} & F &  & 83.7 & 75.2 & 2.0 & 36.7 & 94.5 & \U{94.3} &  & 81.4 & 82.6 & 1.8 & 31.2 & 96.7 & 93.0 \\
RoMa~\cite{edstedt2024roma} & C &  & 86.2 & 85.7 & 0.6 & 72.1 & \U{98.3} & 91.5 &  & 74.0 & 67.5 & 1.0 & 58.9 & 95.1 & 86.5 \\
MASt3R~\cite{mast3r_eccv24} & C &  & 82.0 & 70.7 & 1.2 & 36.9 & 83.5 & 93.5 &  & 83.4 & 84.6 & 0.9 & 31.3 & 97.9 & 94.4 \\
LoMa~\cite{nordstrom2026loma} & F &  & 86.2 & \U{95.6} & \B{0.3} & 3.5 & \B{99.0} & \U{94.3} &  & 78.3 & 92.8 & 1.0 & 0.6 & 94.4 & 79.1 \\
\rowcolor{cyan!16} RoMa v2~\cite{edstedt2025romav2} & C &  & \U{93.2} & \B{96.9} & 0.5 & 92.6 & \B{99.0} & \B{97.5} &  & \B{90.7} & \B{96.3} & \B{0.2} & 88.7 & \B{98.5} & \B{97.0} \\
RoMa v2~\cite{edstedt2025romav2} & F &  & \B{93.5} & \B{96.9} & \B{0.3} & 92.4 & 97.9 & \B{97.5} &  & \U{90.4} & \U{96.2} & \U{0.3} & 88.3 & \U{98.2} & \U{96.8} \\
\bottomrule
\end{tabular}}
\label{tab:scaffold_source_full}
\end{table*}

%% file: tables/5_camera_diagnostics.tex
\begin{table*}[!htbp]
\setlength{\tabcolsep}{2.1pt}
\centering
\caption{\textbf{Camera diagnostics.} We report camera-pose AUC at 1, 5, and 10 degrees and F$^\circ$, the mean field-of-view error in degrees. Most geometry-conditioned rows inherit the same SfM cameras, so the reconstruction gaps in \cref{tab:main_h16_results} cannot be explained by camera accuracy alone. MapAnything receives the same camera conditioning as Scaffold3D, but nevertheless predicts camera updates rather than keeping the input cameras fixed. Lower is better for F$^\circ$.}
\resizebox{\linewidth}{!}{%
\begin{tabular}{l crrrr crrrr crrrr crrrr crrrr}
\toprule
\multirow[b]{2}{*}{Method} & & \multicolumn{4}{c}{\dsid{ScanNet++$_4$}} & & \multicolumn{4}{c}{\dscd{ETH3D$_4$}} & & \multicolumn{4}{c}{\dscd{Tanks\&Temples$_4$}} & & \multicolumn{4}{c}{\dsod{4D-DRESS}} & & \multicolumn{4}{c}{\dsod{MV-dVRK}} \\
\cmidrule(lr){3-6} \cmidrule(lr){8-11} \cmidrule(lr){13-16} \cmidrule(lr){18-21} \cmidrule(lr){23-26}
& & \multicolumn{4}{c}{\textbf{AUC@1/5/10/F$^\circ$}} & & \multicolumn{4}{c}{\textbf{AUC@1/5/10/F$^\circ$}} & & \multicolumn{4}{c}{\textbf{AUC@1/5/10/F$^\circ$}} & & \multicolumn{4}{c}{\textbf{AUC@1/5/10/F$^\circ$}} & & \multicolumn{4}{c}{\textbf{AUC@1/5/10/F$^\circ$}} \\
\midrule
\addlinespace[0.15em]
\multicolumn{26}{l}{\emph{Feed-forward predictors}} \\
\midrule
DUSt3R~\cite{wang2024dust3r} &  & 5.8 & 48.6 & 67.6 & 2.8 &  & 7.7 & 39.4 & 58.3 & 3.4 &  & 15.6 & 51.8 & 69.3 & 2.6 &  & 0.0 & 1.1 & 8.1 & 27.2 &  & 0.7 & 16.7 & 36.9 & 4.1 \\
MASt3R~\cite{mast3r_eccv24} &  & 19.5 & 53.3 & 69.1 & 3.4 &  & 12.8 & 47.3 & 61.3 & 3.2 &  & 18.8 & 59.0 & 74.4 & 2.7 &  & 0.1 & 10.3 & 24.0 & 4.8 &  & 0.3 & 3.9 & 19.7 & 19.9 \\
MASt3R-SfM~\cite{duisterhof2025mastrsfm} &  & 40.2 & 77.6 & 86.1 & 1.2 &  & 52.6 & 76.3 & 81.3 & 0.5 &  & 68.8 & 90.0 & 93.4 & 0.7 &  & 39.1 & 67.5 & 73.2 & 1.0 &  & 58.8 & 78.9 & 85.2 & 1.9 \\
Pow3R~\cite{pow3r_cvpr25} &  & 10.9 & 55.4 & 69.4 & 2.1 &  & 9.0 & 40.8 & 60.5 & 3.4 &  & 18.8 & 56.8 & 71.4 & 1.8 &  & 0.0 & 2.5 & 9.7 & 17.8 &  & 0.2 & 4.3 & 19.6 & 18.5 \\
MapAnything~\cite{keetha2025mapanything} &  & 46.5 & 78.3 & 86.1 & 0.6 &  & 5.8 & 33.8 & 51.5 & 6.0 &  & 22.4 & 69.5 & 82.3 & 3.2 &  & 0.0 & 3.9 & 16.9 & 30.9 &  & 0.0 & 2.1 & 8.5 & 23.7 \\
VGGT~\cite{wang2025vggt} &  & 6.9 & 40.3 & 60.5 & 1.4 &  & 26.9 & 65.0 & 77.4 & 2.0 &  & 65.1 & 84.0 & 88.1 & 0.6 &  & 0.0 & 9.3 & 28.9 & 28.4 &  & 0.0 & 9.2 & 25.3 & 14.5 \\
Pi3~\cite{wang2025pi3} &  & 60.9 & 84.0 & 89.7 & 0.7 &  & 39.1 & 67.6 & 78.4 & 2.3 &  & 55.2 & 81.8 & 87.6 & 0.8 &  & 1.7 & 36.8 & 66.3 & 5.2 &  & 2.3 & 30.3 & 51.5 & 3.4 \\
Pi3X~\cite{wang2025pi3} &  & 55.7 & 81.4 & 88.2 & 0.7 &  & 47.4 & 78.8 & 87.1 & 0.9 &  & 66.7 & 86.4 & 90.5 & \B{0.4} &  & 0.1 & 10.6 & 33.2 & 23.9 &  & 1.2 & 3.9 & 18.5 & 19.2 \\
MASt3R$^\dagger$ &  & 37.9 & 74.5 & 83.6 & 1.5 &  & 8.3 & 43.1 & 62.2 & 3.0 &  & 29.2 & 61.0 & 74.8 & 1.5 &  & 0.0 & 0.3 & 1.3 & 29.2 &  & 2.2 & 29.0 & 52.5 & 2.5 \\
\midrule
\addlinespace[0.15em]
\multicolumn{26}{l}{\emph{Monocular depth completion}} \\
\midrule
OMNI-DC~\cite{zuo2025omni} &  & \B{85.6} & \B{92.8} & \B{94.1} & \B{0.1} &  & \B{91.7} & \B{96.3} & \U{97.8} & \B{0.2} &  & \B{92.7} & \B{97.0} & \B{97.7} & \B{0.4} &  & \B{93.5} & \B{96.9} & \B{97.3} & \B{0.5} &  & \B{91.5} & \B{96.3} & \U{97.1} & \B{0.2} \\
Marigold-DC~\cite{viola2024marigolddc} &  & \B{85.6} & \B{92.8} & \B{94.1} & \B{0.1} &  & \B{91.7} & \B{96.3} & \U{97.8} & \B{0.2} &  & \B{92.7} & \B{97.0} & \B{97.7} & \B{0.4} &  & \B{93.5} & \B{96.9} & \B{97.3} & \B{0.5} &  & \B{91.5} & \B{96.3} & \U{97.1} & \B{0.2} \\
\midrule
\addlinespace[0.15em]
\multicolumn{26}{l}{\emph{Point-token propagation}} \\
\midrule
GGPT (VGGT)~\cite{chen2026ggpt} &  & \U{79.9} & \U{91.4} & 93.4 & \U{0.3} &  & 87.8 & 94.5 & 97.0 & \B{0.2} &  & 89.1 & 91.8 & 92.8 & \B{0.4} &  & \U{92.2} & 93.3 & 94.4 & 1.0 &  & 76.8 & 92.0 & 94.9 & 0.6 \\
GGPT (Pi3)~\cite{chen2026ggpt} &  & \U{79.9} & \U{91.4} & 93.4 & \U{0.3} &  & 87.8 & 94.9 & 97.4 & \U{0.3} &  & 89.1 & 93.1 & 94.2 & \B{0.4} &  & \B{93.5} & \B{96.9} & \B{97.3} & \B{0.5} &  & \U{79.8} & \U{94.6} & \B{97.3} & \U{0.4} \\
GGPT (Pi3X)~\cite{chen2026ggpt} &  & 79.3 & 91.2 & \U{93.5} & 0.4 &  & \U{88.5} & \U{95.9} & \B{97.9} & \U{0.3} &  & \U{91.1} & 93.4 & 94.4 & \U{0.5} &  & 91.2 & \U{96.5} & \U{97.1} & \B{0.5} &  & 77.8 & 92.6 & 95.5 & \U{0.4} \\
GGPT (MASt3R$^\dagger$)~\cite{chen2026ggpt} &  & \B{85.6} & \B{92.8} & \B{94.1} & \B{0.1} &  & \B{91.7} & \B{96.3} & \U{97.8} & \B{0.2} &  & \B{92.7} & \B{97.0} & \B{97.7} & \B{0.4} &  & 55.2 & 56.9 & 57.2 & 2.8 &  & \B{91.5} & \B{96.3} & \U{97.1} & \B{0.2} \\
\midrule
\addlinespace[0.15em]
\multicolumn{26}{l}{\emph{Geometry-conditioned predictors}} \\
\midrule
MVSAnywhere~\cite{izquierdo2025mvsanywhere} &  & \B{85.6} & \B{92.8} & \B{94.1} & \B{0.1} &  & \B{91.7} & \B{96.3} & \U{97.8} & \B{0.2} &  & \B{92.7} & \B{97.0} & \B{97.7} & \B{0.4} &  & \B{93.5} & \B{96.9} & \B{97.3} & \B{0.5} &  & \B{91.5} & \B{96.3} & \U{97.1} & \B{0.2} \\
MapAnything~\cite{keetha2025mapanything} &  & 79.3 & 88.6 & 91.4 & \U{0.3} &  & 73.7 & 89.0 & 93.4 & 0.5 &  & 90.1 & \U{95.5} & \U{96.5} & 0.6 &  & 83.4 & 94.6 & 96.2 & \U{0.9} &  & 62.5 & 87.6 & 92.6 & 0.9 \\
Pi3X~\cite{wang2025pi3} &  & \B{85.6} & \B{92.8} & \B{94.1} & \B{0.1} &  & \B{91.7} & \B{96.3} & \U{97.8} & \B{0.2} &  & \B{92.7} & \B{97.0} & \B{97.7} & \B{0.4} &  & \B{93.5} & \B{96.9} & \B{97.3} & \B{0.5} &  & \B{91.5} & \B{96.3} & \U{97.1} & \B{0.2} \\
Pi3X$^\dagger$~\cite{wang2025pi3} &  & \B{85.6} & \B{92.8} & \B{94.1} & \B{0.1} &  & \B{91.7} & \B{96.3} & \U{97.8} & \B{0.2} &  & \B{92.7} & \B{97.0} & \B{97.7} & \B{0.4} &  & \B{93.5} & \B{96.9} & \B{97.3} & \B{0.5} &  & \B{91.5} & \B{96.3} & \U{97.1} & \B{0.2} \\
Pow3R~\cite{pow3r_cvpr25} &  & \B{85.6} & \B{92.8} & \B{94.1} & \B{0.1} &  & \B{91.7} & \B{96.3} & \U{97.8} & \B{0.2} &  & \B{92.7} & \B{97.0} & \B{97.7} & \B{0.4} &  & \B{93.5} & \B{96.9} & \B{97.3} & \B{0.5} &  & \B{91.5} & \B{96.3} & \U{97.1} & \B{0.2} \\
\rowcolor{cyan!16} Scaffold3D (ours) &  & \B{85.6} & \B{92.8} & \B{94.1} & \B{0.1} &  & \B{91.7} & \B{96.3} & \U{97.8} & \B{0.2} &  & \B{92.7} & \B{97.0} & \B{97.7} & \B{0.4} &  & \B{93.5} & \B{96.9} & \B{97.3} & \B{0.5} &  & \B{91.5} & \B{96.3} & \U{97.1} & \B{0.2} \\
\bottomrule
\end{tabular}
}
\label{tab:camera_diagnostics}
\end{table*}